\pdfoutput=1

\PassOptionsToPackage{table}{xcolor}

\documentclass[letterpaper]{article}

\usepackage{times}
\usepackage{latexsym}

\usepackage[T1]{fontenc}

\usepackage[utf8]{inputenc}

\usepackage{microtype}

\usepackage{inconsolata}

\usepackage{graphicx}

\usepackage{subfigure}
\usepackage{booktabs}

\usepackage{hyperref}
\usepackage{url}

\usepackage{amssymb}
\usepackage{mathtools}
\usepackage{amsthm}

\usepackage[preprint]{acl}

\usepackage[table]{xcolor}
\usepackage{wrapfig}

\usepackage
[
capitalize,
]
{cleveref}

\usepackage[disable,textsize=small]{todonotes}

\usepackage{xurl}

\usetikzlibrary {shapes.geometric}

\usepackage{amsmath,amsfonts,bm}

\def\eqref#1{equation~\ref{#1}}
\def\1{\bm{1}}

\def\vw{{\bm{w}}}
\def\vx{{\bm{x}}}

\def\mW{{\bm{W}}}

\DeclareMathAlphabet{\mathsfit}{\encodingdefault}{\sfdefault}{m}{sl}
\SetMathAlphabet{\mathsfit}{bold}{\encodingdefault}{\sfdefault}{bx}{n}

\DeclareMathOperator*{\argmax}{arg\,max}

\theoremstyle{plain}

\theoremstyle{definition}

\theoremstyle{remark}

\def\win{\vw_{\text{in}}}
\def\wgate{\vw_{\text{gate}}}
\def\wout{\vw_{\text{out}}}
\def\xin{x_{\text{in}}}
\def\xgate{x_{\text{gate}}}
\def\xpost{x_{\text{post}}}
\def\xln{\vx}

\def\wp{\texttt{RefactorGLU}}%name of our weight preprocessing step

\def\blank{\hspace*{.5em}}

\def\shortpar#1{\textbf{#1.}}

\long\def\devour#1{}

\newcounter{notecounter}
\newcommand{\enotesoff}{\long\gdef\enote##1##2{}}
\newcommand{\enoteson}{\long\gdef\enote##1##2{{
\stepcounter{notecounter}
{\large\bf
\hspace{0cm}\arabic{notecounter} $<<<$ ##1: ##2
$>>>$\hspace{1cm}}}}}

\enoteson
\enotesoff

\newcounter{oldnotecounter}
\newcommand{\eoldnotesoff}{\long\gdef\eoldnote##1##2{}}
\newcommand{\eoldnoteson}{\long\gdef\eoldnote##1##2{{
\stepcounter{oldnotecounter}
{\large\bf
\hspace{0cm}\arabic{oldnotecounter} $<<<$ ##1: ##2
$>>>$\hspace{1cm}}}}}

\eoldnoteson
\eoldnotesoff

	\title{Weakening Neurons:\\
	An Input-Output Functionality in Transformers with Outsize Influence}
	
	\author{Sebastian Gerstner$^1$, Hilal AlQuabeh$^2$, Kentaro Inui$^{2,3,4}$, Hinrich Schütze$^1$ \\
	  $^1$ LMU Munich and Munich Center for Machine Learning \\
	  $^2$ MBZUAI \ $^3$ Tohoku University \ $^4$ RIKEN\\
	  \texttt{sgerstner at cis dot lmu dot de}
	  }

\begin{document}
	
	\maketitle

\begin{abstract}
We analyze
the learned input-output behavior of GLU-based neurons in large language models (LLMs).
We propose a simple analysis method:
For each neuron,
we compute
the cosine similarities between its input (reading) and output (writing) weight vectors.
In this scheme, a strong negative cosine similarity indicates the neuron \textit{weakens} the direction it detects in the residual stream, so we call this a \textit{weakening neuron}.
This allows us to gain a number of novel insights.
First,
we show that
nine different LLMs have similar patterns:
weakening neurons appear mostly in late layers
whereas their counterparts,
\textit{(conditional) strengthening} neurons, are
frequent in early-middle layers.
Second, we find that
weakening neurons display surprising behavior:
even
though there are few,
they activate
often and have
a large influence on model behavior.
Third, 
weakening neurons have a strong effect on model output when
gate values are negative -- which is surprising since
negative gate values are not expected to encode
functionality.
\end{abstract}

\section{Introduction}

Mechanistic interpretability research \cite{Elhage2021mathematicalframeworktransformer,saphra-wiegreffe-2024-mechanistic}
attempts to reverse-engineer the \textit{mechanisms} inside neural networks,
such as transformer-based \cite{2017_Vaswani} large language models (LLMs).
Some of this work has
addressed the interpretation of neurons within MLP sublayers;
we follow this line of research.

\eoldnote{hs}{cut for space
footnote{
We use "neuron" to refer to a hidden dimension inside the MLP layer,
not other hidden states of the model.
}
}

Much previous work characterizes neurons
through observational signals; either by examining 
the contexts in which they activate
(e.g. \citealp{voita-etal-2024-neurons})
or by analyzing the output directions associated with their weights\footnote{We use ``weight'' to refer to a
weight vector, not a scalar.}
(e.g. \citealp{2024_Gurnee}).
However, neither of these fully captures the \textit{mechanisms} that neurons implement:
a neuron is defined not only by when it activates or what it writes, but also by the relation between the direction it reads from the residual and the direction it writes back. 
Related input-output (IO) ideas have appeared in prior work. \citet{Elhage2021mathematicalframeworktransformer} hypothesize that negative IO weight cosines may implement memory management, and~\citet{2024_Gurnee} compute such cosines for \mbox{GPT-2} neurons.
However, prior work has not developed an IO taxonomy for gated activation functions
\citep{2020_Shazeer}, which are used in recent LLMs (e.g. \citealp{Llama3.2,Gemma3,Olmo3,Yang2025Qwen3});
nor has it systematically analyzed how IO-defined neuron classes are distributed across modern LLMs and affect model behavior.

\begin{figure}
    \centering
    \includegraphics[width=0.75\linewidth]{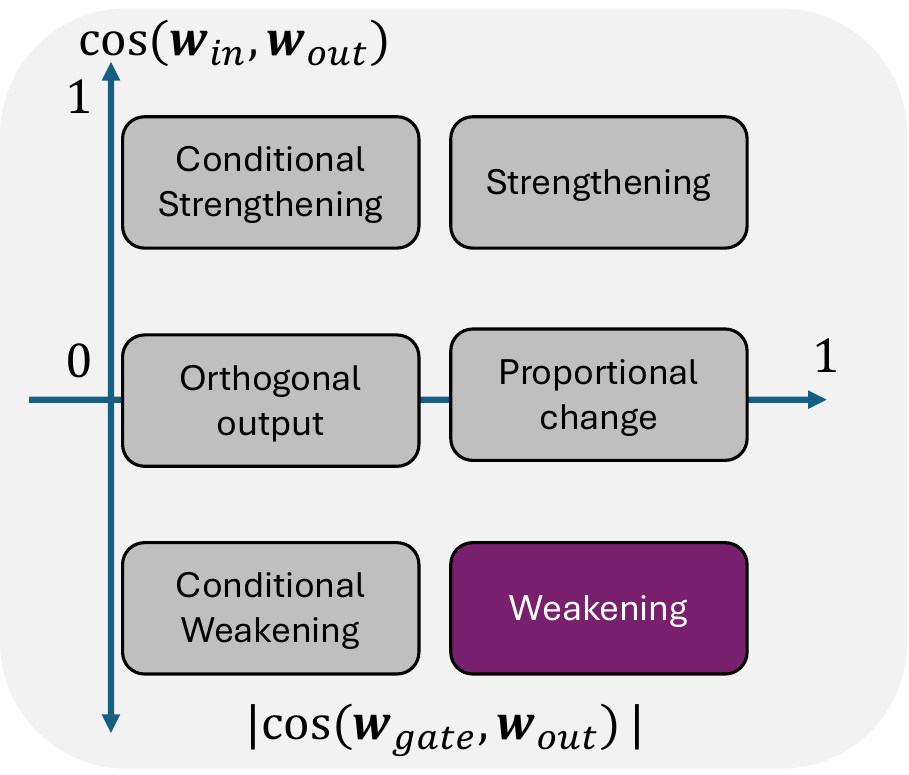}
    \caption{Our IO taxonomy for gated neurons. We classify neurons by how their input, gate, and output weight directions align. See \cref{sec:taxonomy-glu} for details.}\label{fig:table_1_diagram}
\end{figure}

We therefore make this IO relationship the focus of our analysis, starting with a simple yet effective measure: the cosine similarity between the neuron's input (read) weights and its output (write) weights (see \cref{fig:table_1_diagram}).
The input weight represents the direction in the residual stream that causes the neuron to activate, and the output weight represents what the neuron adds back to the residual stream. So, for example, if the output weight vector points in the opposite of the input weight direction, the neuron \textit{weakens} this direction.

Classifying neurons by their read-write relationship reveals systematic structure across models, and the resulting IO classes have distinct effects on model behavior under ablation. In particular, we identify a small class of neurons with strongly negative read-write alignment, which we call \textit{weakening neurons}: although they are relatively few, they activate frequently and have a disproportionately large effect on model behavior.

Our contributions are as follows:
(i)
We introduce an input-output analysis for gated neurons,
using cosine similarities of weight vectors to define a taxonomy of IO functionalities (\cref{fig:table_1_diagram,fig:wcos_selected}).
(ii)
Applying this method to nine LLMs,
we observe consistent patterns:
Early-middle layers contain many \textit{conditional strengthening} neurons,
and late layers tend more towards \textit{weakening} (\cref{fig:medians,fig:bar}).
(iii)
We observe a strong correlation between IO functionalities and activation frequencies (\cref{fig:freq}).
(iv)
We discover
that one small IO class (weakening neurons)
is highly influential in surprising ways:
they activate often
(in the sense of having a gate value above zero),
and they influence various metrics,
even when their gate value is negative (\cref{fig:entropy}).

\section{Related work}
\label{sec:related}

There is a large body of work on interpretability of
transformer-based LLMs.
\citet{Elhage2021mathematicalframeworktransformer} introduce the notion of residual stream.
\citet{2023_Belrose},
\citet{2020_LogitLens}
propose to interpret residual stream states as intermediate guesses about the next token.
\citet{pmlr-v235-rushing24a} discuss this as the \textit{iterative inference hypothesis}.
On a similar note, many works hypothesize that directions in model space can correspond to concepts.
This is the \textit{linear representation hypothesis},
some aspects of which are discussed by \citet{pmlr-v235-park24c}.
\citet{lad2024remarkablerobustnessllmsstages} define \textit{stages of inference}.

\textbf{Neuron analysis.} Much research has attempted to understand individual
neurons;
examples include
\citet{2023_Miller,dai-etal-2022-knowledge,2024_Niu}.

The focus on individual neurons has been criticized.
\citet{2022_Millidge}
find interpretable directions that are not based on individual neurons.
\citet{Elhage2022Toymodelssuperposition} argue that interpretable features 
are non-orthogonal directions in model space
and can be superposed.
This corresponds to sparse linear combinations of neurons in MLP space.
This has inspired a series of work on sparse autoencoders (SAEs), starting with \citet{Sharkey2022}.

The focus on SAEs has  been criticized:
recent studies indicate that they do not always outperform baselines \cite{Kantamneni2025,Mueller2025,Wu2025,Arora2026Languagemodelcircuits},
or that their features are not "canonical" \cite{Leask2025}.
A middle ground is possible:
\citet{2023_Gurnee}
argue that interpretable features correspond to sparse combinations of neurons;
this includes 1-sparse combinations, i.e., individual neurons.
There is recent work that still finds new meaningful classes of neurons
(e.g. "culture-sensitive neurons" in \citealp{zhao-etal-2026-finding},
or "prominent but detrimental neurons" in \citealp{Ali2025Detectingpruningprominent}).

\citet{voita-etal-2024-neurons, 2024_Gurnee}
classify neurons based on the
\textbf{contexts} in which they activate.
E.g., \citet{voita-etal-2024-neurons} find \textit{token
detectors}. \citet{2024_Gurnee}
define \emph{functional roles} of neurons based on
their \textbf{output} weight vector, such
as \textit{suppression neurons} that suppress a specific set
of tokens.

There has been less focus on the \textbf{input-output}
perspective.
A few works apply this perspective at a semantic level:
\citet{geva-etal-2021-transformer} interpret neurons as a key-value memory, where the key corresponds to activation patterns and the value to the tokens promoted by the output weights.
Following this approach,
\citet{gur-arieh-etal-2025-enhancing} show that analyzing the output of a feature helps guess sequences on which it activates.
\citet{mcdougall-etal-2024-copy,elhelo-geva-2025-inferring}
provide a weight-based analysis of input and output tokens of the OV circuit \citep{Elhage2021mathematicalframeworktransformer} of attention heads.

In contrast, we adopt the input-output perspective at the basic mathematical level,
by computing cosine similarities between input and output weights.
\citet{2024_Gurnee} do this for a range of non-GLU models, but do not interpret their results;
and \citet{Elhage2021mathematicalframeworktransformer} 
mention the idea
(footnote 7), but do not follow up. Note 
that input-output analysis for gated activation functions
is complex because, in addition to input and output
weight vectors, the gating mechanism is crucial for IO functionality.

\textbf{Suppression and sharpening mechanisms.}
Several works have addressed various suppression and sharpening mechanisms in transformers.
\citet{lad2024remarkablerobustnessllmsstages} seem to conflate some of them as "residual sharpening",
but there are several concepts to be distinguished from each other and from our IO based neuron analysis:

\citet{Elhage2021mathematicalframeworktransformer}
(mentioned above)
hypothesize that
negative cosines between input and output weights are
mechanisms for \textit{memory management} (MM):
getting rid of intermediate representations that are not needed in later layers.
(Explicit erasure is not the only possible mechanism for MM:
\citet{2023_Heimersheim} also find evidence of overwriting.)
Others interpret erasure mechanisms as
MM:
\citet{janiak-etal-2024-adversarial} observe an attention head erasing the output of a previous one.
\citet{McGrath2023hydraeffect} observe MLP erasure:
MLPs (partially) erasing what previous units have written.
Investigating MLP erasure in more detail,
\citet{pmlr-v235-rushing24a} find
sparse sets of erasure neurons.
They hypothesize a relationship to \citet{2024_Gurnee}'s
"suppression neurons",
an output-based class.

Another class of phenomena is \textbf{confidence management and calibration}.
\citet{mcdougall-etal-2024-copy} describe \textit{copy suppression heads}
that suppress the behavior of other components that copy tokens across positions.
This is distinct from MM in that the suppressed representation was not needed at a previous stage.
They argue that this mechanism improves model calibration.
\citet{2024_Stolfo} (building on \citealp{2024_Gurnee})
find classes of neurons
(entropy neurons, token frequency neurons)
that regulate confidence;
these are output-based classes and therefore \textit{a priori} distinct from our IO classes.
Similar to token frequency neurons,
\citet{2024_Lv} find an anti-overconfidence mechanism at the final layer,
in which attention heads output frequent tokens,
and the MLP directs the residual stream towards a frequency-weighted average token embedding.
\citet{joshi-etal-2025-calibration}
confirm the existence of a confidence correction phase in final layers
and discover a
"calibration direction" mechanism
in early layers.
Other works probe for uncertainty signals in hidden layers; \citet{Stacey2026Hiddenfailuresrobustness} analyze the robustness of these probes.
None of these works relate confidence to IO cosine similarities.
We hypothesize that there is a link between the two, see \cref{sec:discussion}.

\section{Gated activation functions}\label{sec:swiglu}
\label{sec:swiglu-def}
Our work focuses on
\textit{gated activation functions} like SwiGLU or GEGLU \cite{2020_Shazeer}.
Gated activation functions are used widely, e.g.,
Llama \cite{llama,Touvron2023Llama2,Llama3.1,Llama3.2},
OLMo \cite{groeneveld-etal-2024-olmo,Olmo3},
and Qwen \cite{qwen2,Yang2025Qwen3}
use SwiGLU,
and Gemma \cite{gemma_2024,Gemma3}
uses GEGLU.
Here we briefly describe SwiGLU.
GEGLU replaces Swish with GELU, but is otherwise identical.

Traditional activation functions like ReLU
require one weight matrix on the input side and one on the output side:
The MLP outputs
\[\mW_\text{out} \text{ReLU}(\mW_\text{in} \xln),\]
where ReLU is applied element-wise to each
neuron.

Other traditional activation functions are
$\text{Swish}(x):= x / (1+\exp(-x))$ \cite{Ramachandran2017}
and $\text{GELU}(x) := x \Phi(x)$
%\footnote{
%	$\Phi$ is the cumulative distibution function (cdf) of a standard normal distribution.
%}
\cite{Hendrycks2016}.
Both of these can be seen as smooth approximations of ReLU.
They are believed to work better than ReLU
because of better differentiability
(e.g., \citealp{Lee2023}),
i.e., better training dynamics.

In contrast to these traditional functions, a \emph{gated activation function}
like SwiGLU
requires two weight matrices on the input side:
The MLP outputs
\begin{equation}
	\mW_\text{out} \left(\text{Swish}(\mW_{\text{gate}} \xln) \odot (\mW_{\text{in}} \xln)\right),
	\label{eq:mlp}
\end{equation}
where $\odot$ denotes element-wise multiplication
(a.k.a. Hadamard product).
%\footnote{
%In other works,
%$\mW_\text{in}$ and $\mW_\text{out}$
%are also called
%$\mW_\text{up}$ and $\mW_\text{down}$, respectively.
%}

We find it more intuitive to separately consider each neuron:
The neuron adds the vector
\begin{equation}
	\mbox{Swish}(\langle\wgate,\xln\rangle) \cdot \langle\win,\xln\rangle \cdot \wout
	\label{eq:neuron}
\end{equation}
to the residual stream.
Here $\wgate$ and $\win$ are one of the $d_\text{MLP}$ \textit{rows} of $\mW_\text{gate}$ and $\mW_\text{in}$, respectively.
$\wout$ is one of the $d_\text{MLP}$ \textit{columns} of
$\mW_\text{out}$.
%\footnote{
%This is assuming the right-to-left notation of our \cref{eq:mlp}.
%With left-to-right notation rows and columns are switched.
%}
These weight vectors, as well as $\xln$, 
are $\in \mathbb{R}^{d_\text{model}}$, the space of
the residual stream.

In this framework, SwiGLU can be described as a function of two scalars:
\[\mbox{SwiGLU}(\xgate, \xin) := \mbox{Swish}(\xgate) \cdot \xin,\]
where $\xgate := \langle\wgate,\xln\rangle$ and $\xin := \langle\win,\xln\rangle$.
We use $\xpost$ for the neuron activation, i.e.,
$\xpost = \text{SwiGLU}(\xgate, \xin)$.

Unlike ReLU, gated activation functions can output arbitrary positive or negative values.
For example, if $\xgate>0$ and $\xin\ll0$, then $\text{SwiGLU}(\xgate,\xin)\ll0$.

\section{Method}
\label{sec:theory}

\subsection{Approach}\label{sec:approach}
The transformer architecture \citep{2017_Vaswani} includes residual connections \citep{He2016Deepresiduallearning}.
This enables us to adopt the \textit{residual stream} perspective  \citep{Elhage2021mathematicalframeworktransformer}:
Each model unit \textit{reads} from the residual stream
and then updates it by \textit{writing}
to it.
In the case of a SwiGLU neuron,
the scalar products 
$\langle\wgate,\xln\rangle$ and $\langle\win,\xln\rangle$
can be thought of as \textit{reading} how much the residual stream $\xln$ conforms to the \textit{directions} $\wgate$ and $\win$.
The neuron then \textit{writes} a multiple of the direction $\wout$ to the residual stream.
In other words,
$\win$ and $\wgate$ represent the directions in the residual stream that cause the neuron to activate, and $\wout$ represents what the neuron adds back to the residual stream.

A semantic interpretation is that a neuron
detects a \textit{concept} in the residual stream,
and in turn also writes a concept.
This semantic interpretation is not a necessary assumption for our neuron classification,
but is helpful for building intuition and interpreting results.

This framework leads to
our main research question:
\textbf{What is the relation between what a neuron reads and what it writes?}
Our approach is based on \textbf{weights} (as opposed to activations) of \textbf{neurons} (as opposed to, e.g., transcoder features \citep{Dunefsky2024Transcodersfindinterpretable}).

\shortpar{Weight-based}
There are many ways to address the question;
we choose a purely weight-based approach:
computing the
\textbf{cosine similarity of input and output weights}.
This lets us
understand the mathematical function that a neuron implements
in terms of updates to the residual stream.
For example, if the output weight vector points in the opposite of the input weight direction, the neuron \textit{weakens} this direction.

\shortpar{Neuron-based}
This cosine similarity method could
also be applied to transcoder \citep{Dunefsky2024Transcodersfindinterpretable} features instead of neurons.
However,
for this paper we decided to investigate neurons,
and defer a possible investigation of transcoders to future work.
\Cref{sec:stat} shows that,
despite being "only" weight-based and neuron-based,
our method
yields striking results.

\subsection{Taxonomy of IO functionalities}\label{sec:taxonomy}
We now think through what different combinations of weight cosine similarities would mean for neuron IO functionality,
and introduce our terminology.
For the moment we focus on the prototypical cases,
in which cosine similarities are approximately $\pm 1$ or $0$.

Generally, when the output weight is similar enough to (one of) the detected directions,
we speak of \textbf{input manipulation},
as opposed to \textbf{orthogonal output} neurons
which write to
directions not detected in the input.
Intuitively, input manipulator neurons \textit{manipulate} the concept that they detect.

\subsubsection{Taxonomy for non-GLU models}
In non-GLU (e.g. ReLU) models, there are two possible cases of \textbf{input manipulation}.
Either $\cos(\win,\wout)$ is very negative, so the neuron detects a direction and then writes its negation to the residual stream:
we call this a \textbf{weakening} neuron.
Or the cosine is very positive, so the neuron detects a direction and then writes the same direction to the residual stream: we call this a \textbf{strengthening neuron}.

\subsubsection{Taxonomy for GLU models}\label{sec:taxonomy-glu}
GLU models are more complex, because each neuron has a third weight vector, $\wgate$.
In
\cref{fig:table_1_diagram}
we
present a taxonomy of IO functionalities for GLU models.

As special cases of \textbf{input manipulation},
we define:
(i)
\textbf{Strengthening}
and \textbf{weakening} neurons:
all three weight vectors are roughly collinear,
and specifically
$\cos(\win,\wout) \approx \pm 1$.
The neuron detects a direction
and then
adds it to / removes it from the residual stream.
(ii)
\textbf{Conditional strengthening / weakening} neurons:
$\win$ and $\wout$ are roughly collinear and
$\wgate$ is orthogonal to them.
The neuron also strengthens / weakens the direction detected by its $\win$ vector,
but will only activate \emph{conditional on
	$\wgate$ being present in the residual stream}.
(iii)
\textbf{Proportional change} neurons:
$\wout$ is collinear to $\wgate$,
but is orthogonal to $\win$.
If $\wgate$ is present in the residual stream,
then the neuron writes a \emph{positive or negative} multiple of this direction to the residual stream.
This multiple is proportional to the presence of $\win$ in the residual stream.

These prototypical classes are limited in scope:
Many cosines will not be close to 0 or $\pm 1$.
For this general case, this paper explores three options to understand
neuron IO functionalities
at different levels of granularity:
(1)
Classify neurons according to the closest prototypical case (we choose a threshold $\tau=\pm0.5$).
(2)
Plot the marginal distributions of the three cosine similarities.
(3)
Place neurons in a scatter plot,
based on their three weight cosines.

In option 1 (threshold-based classification),
$\cos(\win,\wgate)$ may not always ``match'' the other two cosine similarities.
Consider the case of strengthening:
In the prototypical case with exact equalities
($\cos(\win,\wout)=\cos(\wgate,\wout)$=1),
all three weight vectors are collinear,
so we also have $\cos(\win,\wgate)$=1.
But without exact equalities
(if we just know $\cos(\win,\wout)$ and $\cos(\wgate,\wout)$ are both above $0.5$),
it does not follow that $\cos(\win,\wgate)$ is also above $0.5$.\footnote{
For example, the two reading weights
may be orthogonal ($\cos(\win,\wgate)=0<0.5$), but
$\wout= \wgate + \win$;
then
$\cos(\win,\wout) = \cos(\wgate,\wout) \approx 0.71 > 0.5$.}
When such a ``mismatch''
occurs,
we prepend \textit{atypical}
to the category's name:
In this example, we will speak of
an atypical strengthening neuron.
In \cref{fig:bar} we will see that such neurons exist,
but are quite rare overall.

See \cref{tab:alldefs} in \cref{ap:alldefs} for the complete definitions of all classes, in the form of an explicit decision table.

\subsection{Random baselines}
\label{sec:baseline}
To test whether 
a cosine similarity between weight vectors
is significantly different from random,
we consider two baselines:
(i)
i.i.d.\ Gaussian vectors
(i.e., a randomly initialized model);
(ii)
a layer-specific baseline based on "mismatched cosines".
See \cref{ap:baseline} for details.

\subsection{Implementation}
\label{sec:processing}

We publish our code at \url{https://github.com/sjgerstner/RW_functionalities}.

Our code uses TransformerLens v2 \cite{nanda2022transformerlens}.
When a model is loaded, preprocessing steps are applied 
to make the weights more interpretable without changing model behavior.\footnote{
See the TransformerLens documentation at \url{https://github.com/TransformerLensOrg/TransformerLens/blob/main/further_comments.md}.
}
We use a custom version of TransformerLens that
applies an additional preprocessing step.
See \cref{ap:preprocessing} for details.

\begin{figure}
	\includegraphics
	[width=\linewidth]
	{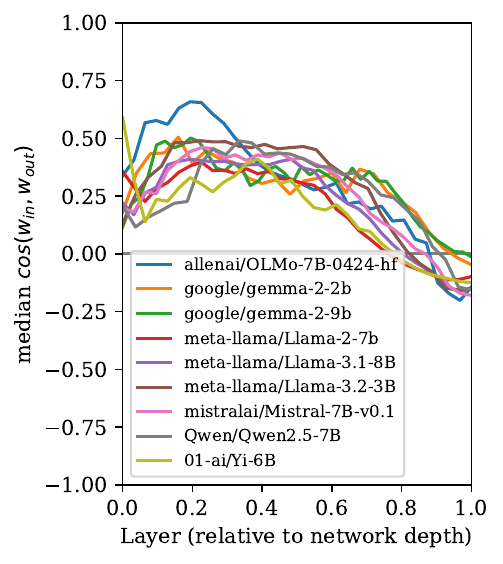}
	\caption{
		Median of $\cos(\win,\wout)$ (y-axis)
		by layer (x-axis)
		for 9 GLU models of 2B to 9B parameters.
		(See
		\cref{fig:all_medians}
		in
		\cref{ap:hr-glu}
		for a plot with all 12 GLU models investigated, including those smaller than 2B.)
		For all models, the value is positive in
		the beginning
		and negative in the end,
		indicating that early-middle layers ``strengthen'' directions they find in the
		residual stream whereas
		later layers tend
		more towards ``weakening'' them.
	}
	\label{fig:medians}
\end{figure}

\section{Where to find weakening neurons}
\label{sec:stat}

In this section we compute cosine similarities of neuron weights
as described in \cref{sec:theory},
to investigate
which IO functionalities actually appear in LLMs,
and in which layers.
We first briefly investigate non-GLU models,
and then focus on GLU models (which are more recent).
Within each group, our results are strikingly \textbf{consistent across models}:
In particular,
there is always a
substantial
number of \textbf{weakening neurons},
and many other neurons also have IO cosine similarities far from zero.

\subsection{Non-GLU models}

We apply our method to 24 transformer models, covering a wide range of
architectures (encoder-decoder, encoder-only, decoder-only),
activation functions (ReLU and GELU),
and training data (language and non-language).
See \cref{ap:models-non-GLU} for the full list.

As this paper's focus is on the more recent GLU models,
we defer the results to
\cref{ap:hr-non-glu}.

\subsection{GLU models}
\label{sec:stat-glu}
\begin{figure}
	\centering
	\includegraphics
	[height=.4\textheight]
	{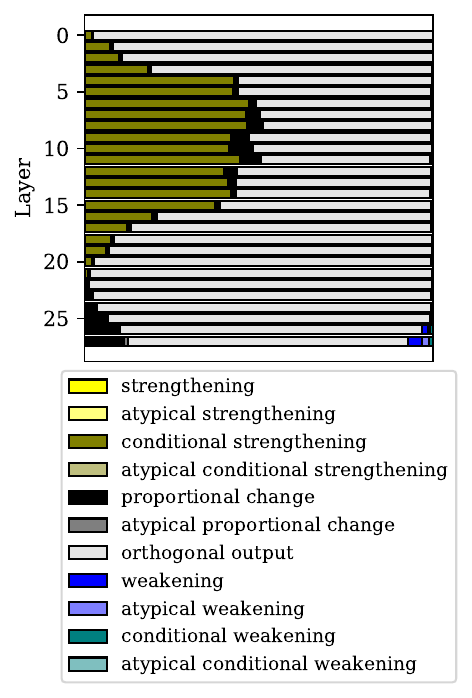}
	\caption{
		Distribution of neuron IO classes by layer and category in a GLU-based model (Llama-3.2-3B).
		Length of bars represents number of neurons.
		See \cref{tab:meta-llama/Llama-3.2-3B} for the exact numbers.
	}
	\label{fig:bar}
\end{figure}

\begin{figure*}
	\centering
	\includegraphics
	[width=\textwidth]
	{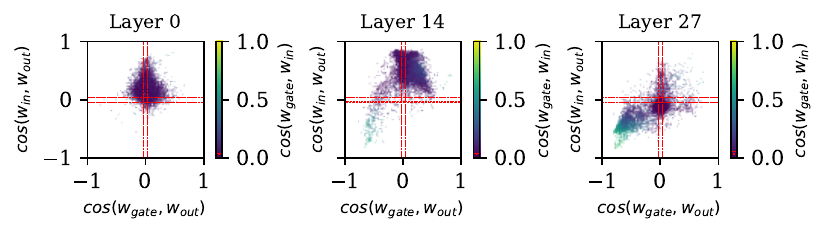}
	\caption{
		Fine-grained analysis of neuron IO
		behavior in three layers
		of Llama-3.2-3B,
		based on the configuration of
		their three weight vectors in parameter space.
		Each subplot represents a layer, each dot
		a neuron.
		The red lines mark the 95\% randomness regions for each of the three cosine values.
		(There is a dotted line for variant (i) and a dashed line for variant (ii) in
		\cref{sec:baseline},
		but they are almost the same.)\\
		We see that many neurons are outside the randomness regions,
		indicating that they manipulate their input in some way.
		Purple dots at the top of the plots are conditional strengthening neurons.
		Lighter dots in the bottom left corner are weakening neurons.
	}
	\label{fig:wcos_selected}
\end{figure*}

We apply our method to 12 GLU-based LLMs,
covering both the SwiGLU and GEGLU activation functions.
See \cref{ap:models-GLU} for a full list.

To demonstrate our finding,
we present three representative plots.
(See \cref{ap:hr-glu} for more.)
\Cref{fig:medians}
shows the median value of $\cos(\win,\wout)$
across all layers of the nine larger models.
The common pattern is clearly visible:
In early-middle layers of all models,
a majority of neurons has a $\cos(\win,\wout)$ high above zero,
indicating strengthening;
in late layers,
this median cosine similarity goes slightly below zero,
indicating a relative majority of weakening neurons.

The other two plots focus on 
\textbf{Llama-3.2-3B} \citep{Llama3.2},
but
the patterns we describe are general:
see \cref{ap:hr-glu} for other models.
\Cref{fig:bar}
(equivalently, \cref{tab:meta-llama/Llama-3.2-3B})
shows IO class distribution across layers.
In \cref{fig:wcos_selected},
we plot
the distribution of neurons
in a few selected layers,
by displaying each neuron as a point with
$\cos(\wgate,\wout)$ indicated on the x-axis,
$\cos(\win,\wout)$ on the y-axis
and
$\cos(\wgate,\win)$ as its color.

\shortpar{Input manipulation}
First,
we see that a large proportion of neurons
are input manipulators
(i.e., they are not orthogonal output neurons):
In \cref{fig:bar},
these are 25\% of all neurons,
and as much as 50\% in early-middle layers
(layers 7--11 -- we use
        zero-based indexing).
What is more,
\cref{fig:wcos_selected}
shows that even
neurons classified as orthogonal output often
belong to clusters centered above/below
the horizontal line.
Their weight cosine similarities often exceed the
significance threshold
(indicated in the figure as red lines).
E.g., in layer 14, there are many neurons whose $\cos(\win,\wout)$ (y-axis) is below 0.5 but above the significance threshold.
This suggests that even orthogonal output neurons perform input manipulation to some extent.

\shortpar{Different IO functionalities}
Weakening neurons represent a large share of
the (relatively few) input manipulators in late layers.
They form a somewhat separate cluster in
\cref{fig:wcos_selected}
(in the bottom-left corner of the rightmost subplot).
Another important input manipulator class in late layers is proportional change.
In contrast,
across all models,
early middle layers are dominated by conditional strengthening.
In fact,
the majority of input manipulators
(more than 80\% in Llama)
belong to just this one class.

This general pattern of strengthening-then-weakening holds across models,
as \cref{fig:medians}
shows at one glance.
In \cref{fig:wcos_selected} (and \cref{fig:wcos} in the appendix),
the pattern manifests as a large cluster of neurons,
centered clearly above the x-axis in most layers,
but moving below it in the last few layers.

In summary,
we find across models that
conditional strengthening dominates in early-middle layers,
but in late layers we find more weakening neurons.

\begin{table}
	\caption{Pearson correlations between $\cos(\win,\wout)$ and frequency of $\xgate>0$. Sample size is number of MLP neurons in the model. All correlations are significant with $p<0.01$.}
	\label{tab:freq-all}
	\centering
	\begin{tabular}{c|r}
		Model & correlation \\
		\hline
		OLMo-7B-0424 & -0.91 \\
		Gemma-2-2B & -0.76 \\
		Llama-3.1-8B & -0.59 \\
		Llama-3.2-3B & -0.54 \\
	\end{tabular}
\end{table}

\begin{figure}
	\centering
	\includegraphics
	[width=.45\textwidth]
	{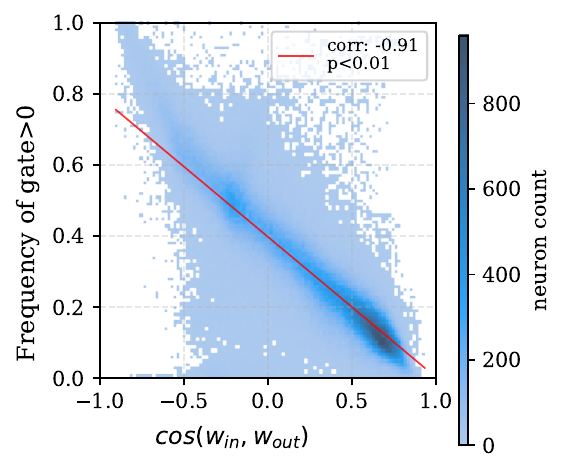}
	\caption{
		Heatmap of $\cos(\win,\wout)$ (x-axis) vs. activation frequency (y-axis), when running OLMo-7B on a subset of Dolma (\cref{sec:freq}).
		The darker the color, the more neurons in the given area.
		This result is
		evidence that weakening neurons are a disproportionately important IO functionality, despite being only a small class.
	}
	\label{fig:freq}
\end{figure}

\section{Weakening neurons activate often}
\label{sec:freq}

In \cref{sec:stat-glu}, we found striking differences in the number of neurons of different IO classes.
This
raises the question of
how often neurons of a given class activate,
i.e., how often their gate value is positive.
For example, there is a large number of conditional strengthening neurons, but how often does each of them actually activate?

In fact, \citet{2024_Gurnee} found a negative correlation between activation frequency and $\cos(\win,\wout)$ -- but in GELU models.
We now investigate whether a similar phenomenon occurs with gated activation functions.

\paragraph{Models and corpus.}
We analyze four GLU-based models:
OLMo-7B-0424 (\citealp{groeneveld-etal-2024-olmo}, henceforth "OLMo-7B"),
Gemma-2-2B \citep{gemma_2024},
Llama-3.1-8B \citep{Llama3.1},
and Llama-3.2-3B \citep{Llama3.2}.
As a dataset, we use a random subset of 20M tokens
(following \citet{voita-etal-2024-neurons})
from Dolma \citep{soldaini-etal-2024-dolma},
the training dataset of OLMo.
For the Gemma and Llama models, there is no publicly available training dataset, so Dolma may only represent a sub-distribution of their training distribution (for example, Dolma is only English and code, whereas the Llama models were trained on a multilingual dataset). We believe, however, that this sub-distribution is enough for our purposes here.

\paragraph{Results.}
We show an overview of results in \cref{tab:freq-all},
and a plot for OLMo in \cref{fig:freq}.
A layer-wise analysis
can be found
in \cref{ap:freq}.
The results for the other three models can be found in the \href{https://github.com/sjgerstner/RW_functionalities_results}{supplementary material}.

Consistent with \citet{2024_Gurnee},
we find that
the many (conditional) strengthening neurons activate very rarely,
and (conditional) \textbf{weakening neurons activate very often}.
In fact,
there is an almost linear negative relationship between $\cos(\win,\wout)$ and activation
frequency:
the correlation is $-0.91$ for OLMo,
which is even stronger than the correlations reported by \citet{2024_Gurnee}
(up to $-0.69$ depending on the model).
When considering layers separately (\cref{fig:freq-all}),
we see that the effect is mostly due to middle layers.

This result is
a first indication that weakening neurons are a disproportionately important IO functionality, despite being only a small class.

\section{Ablation experiments}\label{sec:ablation}

Since model training produced so many input manipulator neurons (\cref{sec:stat}),
we hypothesize that they must contribute to model performance in an important way.
We now test this by ablating neurons
based on their IO functionality.

We find that
weakening neurons have the highest effect
on the metrics that we tested -- this is
unexpected since weakening neurons are a small class of
a few hundred neurons.
We also find that this phenomenon is not fully explained by their activation frequency:
even the (rare and small) negative gate values of weakening neurons are influential.

We use the same corpus as in \cref{sec:freq} (20M tokens from Dolma).
To save resources, we further narrow down the range of models,
and focus on
OLMo-7B and Llama-3.2-3B.
This section mostly shows the results from OLMo; those from Llama are similar, see \href{https://github.com/sjgerstner/RW_functionalities_results}{supplementary material}.

\subsection{Effect size of ablating different IO classes}
We run the model on our corpus and record various metrics,
such as the loss
and the entropy of the output distribution.
In each run we ablate a number of neurons from a different IO class,
or (as a baseline) the same number of \textit{random} neurons from the same layers.
This enables us to observe the effect of various IO classes on these metrics.
The baseline checks whether effects are due to the layers rather than IO classes.

In each run, we ablate as many neurons from the given class as there are weakening neurons. For example, OLMo-7B has 526 weakening neurons, so in each run we ablate 526 neurons of a given class.

We try two types of ablation:
zero ablation (setting activations to zero),
and mean ablation (setting them to the mean activation of
the given neuron, see \cref{ap:mean-ablation} for details).
In this section we focus on mean ablation;
for zero-ablation results
see \cref{ap:ablations,ap:hr-ablations}.

We find that \textbf{ablating weakening neurons has the highest effect} on several metrics,
compared with other classes or with other neurons from the same layers.

\begin{figure}
	\centering
	\includegraphics
	[width=.8\linewidth]
	{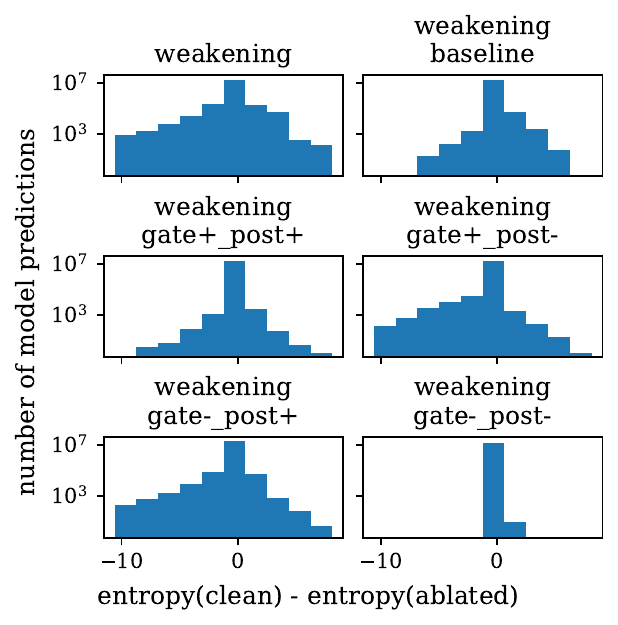}
	\caption{
		Effect of mean-ablating weakening neurons on entropy of the model's output distribution.
		%NOTE Whoever wrote "entropy of neuron activations": that's wrong!!
		For example (top left subplot, leftmost bar), in $\approx 10^3$ next-token
	predictions, weakening neurons decrease the entropy
	by about 10 nats, whereas they increase it more rarely.
		"Weakening baseline" denotes random neurons from the same layers as weakening neurons.
		The bottom four plots describe the results of conditional ablations
		(\cref{sec:conditional}).
		E.g., "gate+\_post+" describes the effect of those activations in which $\xgate$$>$$0$ and $\xpost$$>$$0$.
		\label{fig:entropy}\label{fig:conditional}
		}
\end{figure}

For the effect on entropy (\cref{fig:entropy,ap:hr-ablations}), we have an expected finding and a surprising one:
The expected one is that strengthening neurons often make the output distribution sharper. (They reduce entropy in \cref{fig:entropy-nstrengthening-mean}.)
Weakening neurons (top left panel of \cref{fig:entropy})
often flatten the output distribution, as expected;
but surprisingly, they even more often sharpen it.
Other classes do not have such a big effect (\cref{fig:entropy-nweakening-mean}).
We would expect the opposite: removing information from the
residual stream should make it less informative and
therefore flatten the output distribution.

\subsection{Conditional ablations}
\label{sec:conditional}
We now try to
explain why weakening neurons sharpen (instead of flatten)
the distribution.
We use \textbf{conditional ablations}:
We ablate only some activations of each neuron,
based on the signs of the corresponding $\xgate$ and $\xin$.
Specifically, we consider the following four conditions,
which we call \textit{activation quadrants}
(the definitions are simplified here, see \cref{ap:preprocessing2} for a more precise statement):
(i)
$\xgate>0, \xin>0$, leading to $\xpost>0$;
(ii)
$\xgate>0, \xin<0$, leading to $\xpost<0$;
(iii)
$\xgate<0, \xin<0$, leading to $\xpost>0$;
(iv)
$\xgate<0, \xin>0$, leading to $\xpost<0$.

We find that a large part of
the sharpening effect
of weakening neurons
(and to a lesser extent also their flattening effect in other contexts)
is due to quadrants (ii) and (iii):
In \cref{fig:conditional},
these quadrants (middle right and bottom left subplot)
show entropy effects similar to those of weakening neurons as a whole,
whereas this is much less the case for the other subplots.
In the following we argue that
this is surprising,
but also partially
solves the mystery we encountered earlier.

The large effect of
quadrant
(iii) activations ($\xgate<0$,
$\xin<0$) is especially
surprising
for two reasons:
First, these negative $\xgate$ activations are relatively rare in weakening neurons (\cref{sec:freq}).
Second, because of the Swish function,
\textbf{negative gate values} are relatively small (whereas positive values can be arbitrarily large),
and it was often assumed they were only useful for training dynamics (see \cref{sec:swiglu}).
Our results show
for the first time
(concurrently with \citealp{Kong2026Negativepreactivations} who focus on a different
phenomenon)
that negative gate values
have a strong effect on model behavior (not just
training).
This shows that, for mechanistic interpretability research,
\textbf{Swish is not reducible to ReLU.}

The large effect of
quadrant
(ii) activations ($\xgate>0$, but $\xin<0$) is surprising in a different way:
Weakening neurons have a strong $\cos(\wgate,\win)$
(and this similarity is always positive
thanks to the weight preprocessing described in \cref{ap:preprocessing2}).
Therefore the region of activation space where $\xgate>0$, but $\xin<0$, is small.
Our finding shows that the residual stream still ends up in this region often enough
that it has a tangible effect.

Our finding could partially
explain
the sharpening effects of weakening neurons:
For quadrants (ii) and (iii) activations, the usual neuron behavior gets a minus sign in front,
so that weakening neurons take on a strengthening behavior.
Consider
OLMo neuron 31.9634,
further investigated in \cref{sec:cs}.
It
usually
detects "minus \textit{again}" ($\wgate$)
and writes "\textit{again}" ($\wout$); then
in quadrant (iii) it
detects "\textit{again}" ($-\wgate$)
and writes "\textit{again}" ($\wout$),
which indeed makes the output distribution sharper.
Similarly, in quadrant (ii),
it both detects and writes "minus \textit{again}" ($\wgate, -\wout$).

However, the case study below suggests that this is not the full explanation, at least not in all cases.

\subsection{Case study of entropy reduction}
To understand this phenomenon further,
we study the text example
where the entropy reduction by quadrant (iii) activations of weakening neurons is most extreme
(with zero ablation).

The input text is:
\textit{Yesterday (21 December) the Government announced a package of support for hospitality and leisure businesses that are losing trade because of the O}
and the correct next token is
\textit{mic} (as in \textit{Omicron}).
The model predicts this next token correctly.

Which tokens have the largest score difference between clean and ablated runs?
We find that, in the clean run, \textit{mic} and similar tokens get a massive boost (of up to 12 points) compared to the ablated run,
whereas no token gets its score reduced by nearly as much.
Thus, at least in this case, the quadrant (iii) activations of weakening neurons sharpen the output distribution by boosting the correct next token.

Ablating various subsets of weakening neurons, we find
that in this case no single weakening neuron achieves the observed effect on its own
-- but there is a small subset of 8 weakening neurons that does.
Moreover, the direct effect of weakening neurons does not correspond to the observed effect,
so the relevant effect is indirect.

Thus, in this particular case, the entropy-reducing effect of weakening neurons is due to an indirect and cumulative effect.
More fully understanding the mechanisms behind this is left to future work.

\section{Case study of a weakening neuron}
\label{sec:cs}
To complement the quantitative results of the previous sections,
we qualitatively examine a few neurons based on their IO class.
In this section we focus on two OLMo-7B neurons:
a strengthening and a weakening neuron.
See \cref{ap:cs} for more details and analysis of additional
neurons.
These case studies show some things that IO-based classes of neurons (such as weakening neurons) \textit{can} do; we do not use the case studies to back up any statistical claims.

To analyze the neurons,
we combine the IO perspective with two well-established neuron analysis methods:
projecting weights to vocabulary space \cite{geva-etal-2022-transformer, dar-etal-2023-analyzing, 2024_Gurnee, voita-etal-2024-neurons},
and finding text examples which strongly activate the neuron \cite{
	geva-etal-2021-transformer, Nanda2022, voita-etal-2024-neurons, 2024_Gurnee}.

For the activation-based analyses,
we publish code at
\url{https://github.com/sjgerstner/gluscope}
and visualize results at
\url{https://gluscope.github.io}.

We choose two OLMo-7B
neurons: \textbf{\href{https://gluscope.github.io/OLMo-7B-0424/L28/N4737/vis.html}{28.4737}} for strengthening
and \textbf{\href{https://gluscope.github.io/OLMo-7B-0424/L31/N9634/vis.html}{31.9634}} for weakening.\footnote{
	The notation is "layer.neuron", with zero-based indexing. The model has 32 layers, so our weakening neuron is in the final layer.
	%NOTE Don't remove this footnote! It was explicitly asked for by an ARR reviewer!
}

From the data in \cref{ap:cs},
we can see that
\textbf{strengthening neuron 28.4737}
has a straightforward input-output behavior:
It further promotes \textit{review} when
the residual stream already indicates that
this should be the next token.

In contrast,
\textbf{weakening neuron 31.9634}
is harder to interpret.
The weights indicate that
this neuron produces "\textit{again}" when the residual stream contains "minus \textit{again}";
but the examples strongly activating the neuron do not have an obvious semantic relationship to \textit{again}.

31.9634 activates
weakly positively
when \textit{again} is a plausible continuation,
e.g., on the token \textit{\textbf{once}} (as in \textit{once again}).
These are cases with negative $\xgate$ values (and also $\xin<0$, hence positive activations) --
a case that we found to be important in \cref{sec:conditional}.
In
these cases, \textit{again} is already weakly present in the
residual stream before the last MLP,
and the neuron
reinforces \textit{again}.
Thus the behavior of this particular weakening neuron is interpretable in the $\xgate<0$ case,
echoing our finding from \cref{sec:conditional} that this case is surprisingly relevant to model behavior.

These two case studies show that
even when the output weights are highly interpretable,
strengthening and weakening have a very different overall behavior,
and the weakening behavior is more complex.
We think that this is due to the nature of weakening:
at least in quadrant (i) ($\xgate,\xin>0$),
it inherently involves
(an apparent) conflict
between the intermediate model prediction and what the neuron promotes.

\section{Discussion}
\label{sec:discussion}

In summary, we have found:
(i)
In all investigated LLMs, early-middle layers contain many conditional strengthening neurons, and the last few layers contain a small but substantial number of weakening neurons (\cref{sec:stat}).
(ii)
In most layers except the last two, IO functionalities are strongly correlated to activation frequencies: a (conditional) strengthening neuron rarely has positive gate values, a (conditional) weakening neuron more often (\cref{sec:freq}).
(iii)
Even partially ablating weakening neurons has a big influence on model output; ablating neurons from other IO classes has a smaller impact (\cref{sec:ablation}).

In this section, we discuss possible explanations for these findings.
These are only speculative hypotheses, not conclusive
interpretations.
We will work on confirming them in future work.

Recall from \cref{sec:related} that negative input-output cosine similarities were previously hypothesized (but not shown) to implement a \textit{memory management} mechanism.
\textit{Copy suppression heads} are another known suppression mechanism, distinct from memory management.
We think that such explanations are not sufficient: they can only explain \textit{negative} cosine similarities, but we also find many neurons with strong positive similarities.

In the rest of this section,
we focus first on middle layers with conditional strengthening neurons,
then on late layers with weakening neurons.

\subsection{Middle layers, conditional strengthening neurons}
In middle layers we found a large number of conditional strengthening neurons, and a strong negative correlation between $\cos(\win,\wout)$ and frequency of $\xgate>0$.

We think these findings are consistent with interpreting middle-layer IO functionalities as implementing a \textbf{confidence management} mechanism.
Generally speaking, pretrained models are
calibrated \citep{OpenAI2023GPT4,Kadavath2022Languagemodelsmostly}.
This entails
avoiding overconfidence,
so it makes sense to have (conditional) weakening mechanisms that activate often.
On the other hand, the model also needs to increase the confidence of certain hidden representations,
but only in very specific circumstances in which the confidence is actually warranted;
this could explain why we have many \textit{conditional} strengthening neurons
(each of them checks a condition that is distinct from the concept to be strengthened),
and also why each of them activates rarely.

This is not about
confidence in next-token predictions.
Otherwise conditional strengthening neurons would decrease
the output entropy (sharpen the distribution)
more than other neurons from the same
layer.
This is not the case
(\cref{sec:ablation,fig:entropy-nweakening-mean}).

Rather, middle layers tend to represent more abstract, high-level concepts \citep{lad2024remarkablerobustnessllmsstages}:
For example, \citet{wendler-etal-2024-llamas} find language-independent representations in middle layers of multilingual LMs.
We hypothesize that IO functionalities in middle layers
modulate confidence in
this kind of concepts.

\citet{Stacey2026Hiddenfailuresrobustness} found that probes trained on middle-layer representations predict uncertainty better than final-layer probes.
This is another indication that middle layers are relevant for confidence and uncertainty.

\subsection{Late layers, weakening neurons}

Weakening neurons are a relatively small class, but clearly very important:
They consistently appear in late layers of all models, and ablating them has an outsize influence on model output.

Their behavior is unintuitive in several ways.
In particular, negative gate activations, though small and rare, seem to play an important role in these neurons, and switch their behavior from the theoretically expected weakening to a strengthening-like functionality (\cref{sec:conditional}).
We therefore think it is important to consider their different activation quadrants separately, as we do in \cref{sec:cs,ap:cs-weakening}.

Quadrant (iii) ($\xgate<0, \xin<0$):
As found in \cref{sec:conditional}, this quadrant  is surprisingly relevant and corresponds to a strengthening-like behavior.

For quadrants (i) and (ii) ($\xgate>0$), we propose some hypotheses based on our case study in \cref{ap:cs-weakening}.

Quadrant (i) ($\xgate>0, \xin>0$):
We hypothesize that these activations of weakening neurons implement an \textbf{error correction} mechanism.
Specifically,
superposition \citep{Elhage2022Toymodelssuperposition} can have unwanted side effects:
the residual stream can sometimes end up near a meaningful direction (e.g., near "minus \textit{again}")
even when this is not semantically justified.
%In fact, directions of the type "minus <token>" are unlikely to be semantically justified in any situation:
%this would mean that the given token is a particularly bad prediction, while all other tokens (even \textit{SuperGoldMagicCarp}) are good enough.
A weakening neuron could detect such situations (e.g., detect the "minus again" direction) and then weaken the unwanted direction.

Quadrant (ii) ($\xgate>0, \xin<0$):
Weakening neurons have a strong similarity between $\wgate$ and $\win$,
so if $\xgate$ and $\xin$ have different signs this could indicate conflicting information in the residual stream.
The neuron resolves this conflict by boosting $-\wout$, corresponding to the information detected by $\wgate$.
In our case study of the \textit{again} neuron,
the conflicting information manifested as follows:
the correct next token was often an adverb like \textit{meanwhile} or \textit{instead} --
syntactically somewhat similar to \textit{again}, but clearly not the same.
The neuron may ensure only these tokens are predicted,
and not the relatively similar \textit{again}.
Such a mechanism could be described as \textbf{conflict resolution} or \textbf{prediction sharpening}.

Finally, the activations in quadrant (iv) are small and rare and seem to have little influence when ablated (see e.g. \cref{fig:conditional}).

\section{Conclusion}

We propose a new interpretability methodology, analyzing
the input-output behavior of neurons. Using this
methodology, we contribute several novel findings, including:
(i) IO functionality is broadly similar across current models.
(ii) IO functionality is predictive of
activation frequency.
(iii) One specific IO class,
the weakening neurons, greatly
influences model behavior and does so for negative gate
values, which were previously thought not to impact
transformer functionality.

We hope that IO functionality analysis will become an
important part of the toolkit of interpretability.

\section*{Limitations}
\subsection*{Limitations of weight cosine approach}
We focus on a \textit{parameter-based} interpretation of \textit{single neurons}.
This has the advantage of being simple and efficient,
but is also inherently limited in scope.
Accordingly, our method is not designed to replace other neuron analysis methods,
but to complement them.

The mathematical similarities of weights are insightful,
but they should not be taken as one-to-one representations of semantic similarity.
We find cases in which close-to-orthogonal vectors represent very similar concepts (double checking, \cref{sec:doublecheck}).

The IO approach inherently assumes that the representations at different layers are (to some extent) comparable to each other, which is guaranteed by residual connections.
Residual connections are used in essentially all transformer LLMs that we know of, and are also a prerequisite for well-known methods such as the logit lens \citep{2020_LogitLens}.
However, recent work proposed a method to train transformers without residual connections \citep{Ji2026Cuttingskip}.
If this becomes standard, our method will not be applicable any more.
Similarly, our method is not directly applicable to models like DeepSeek-V4 \citep{2026DeepSeekV4} that use hyper-connections \citep{Zhu2025Hyperconnections}.

\subsection*{Limitations of ablation experiments}
Our ablation experiments indicate what effect different IO classes have at the level of next-token predictions.
However, they do not show the intermediate steps that lead from the neuron activations to the model output.

A limitation specific to (our implementation of) conditional ablation is that it is brittle
when applied to several layers at once (as is the case in our work):
ablating an upstream neuron can change the sign of a downstream neuron activation, and thus influence whether this downstream neuron is ablated.

Conditional ablations have another limitation:
getting a similar effect to the original ablation does not imply that the effect happens for the same reason in both runs (although it is a strong indication).

\subsection*{Generalizability across models}
We applied our weight-based analysis (\cref{sec:stat}) to a wide range of models, but did not test mixture-of-experts (MoE) models, or models larger than 9B.

In the more compute-intensive corpus-based experiments (\cref{sec:ablation,sec:freq,sec:cs}),
we focused on a smaller set of models.

\subsection*{Limited understanding of results}
Our findings are striking,
but we do not yet fully understand the reasons behind them, and some of our interpretations are speculative.

In particular, we propose to interpret many of our findings as a confidence management system (see \cref{sec:discussion}).
If correct, this interpretation could explain both the general pattern of which IO functionalities are found in which layers, and specifically the correlation with activation frequencies.
However: (1) we do not currently have conclusive evidence for such an interpretation; (2) it would not explain everything, and in particular does not apply to later layers (and hence the majority of weakening neurons).

This means in particular that our understanding of weakening neurons is still limited.

\section*{Acknowledgements}
We would like to thank our colleagues
Florian Eichin, Dawar Hakimi, Lea Hirlimann, Yihong Liu, Ali Modarressi, Philipp Mondorf, Leonor Veloso and Mingyang Wang
for fruitful discussions and encouragements.
A special thank goes to Dawar Hakimi,
who programmed a TransformerLens version that supports OLMo,
thus sparing us some trouble in the early days of the project.

This work was funded by Deutsche Forschungsgemeinschaft (project SCHU 2246/14-1).

The authors gratefully acknowledge the scientific support and HPC resources provided by the Erlangen National High Performance Computing Center (NHR@FAU) of the Friedrich-Alexander-Universität Erlangen-Nürnberg (FAU) under the NHR project b309dd / JA-27440 / InterpGLU. NHR funding is provided by federal and Bavarian state authorities.

\bibliography{custom,anthology}

\begin{thebibliography}{81}
\providecommand{\natexlab}[1]{#1}

\bibitem[{01.AI et~al.(2025)01.AI, :, Young, Chen, Li, Huang, Zhang, Zhang,
  Wang, Li, Zhu, Chen, Chang, Yu, Liu, Liu, Yue, Yang, Yang, Xie, Huang, Hu,
  Ren, Niu, Nie, Li, Xu, Liu, Wang, Cai, Gu, Liu, and
  Dai}]{ai2025yiopenfoundationmodels}
01.AI, :, Alex Young, Bei Chen, Chao Li, Chengen Huang, Ge~Zhang, Guanwei
  Zhang, Guoyin Wang, Heng Li, Jiangcheng Zhu, Jianqun Chen, Jing Chang,
  Kaidong Yu, Peng Liu, Qiang Liu, Shawn Yue, Senbin Yang, Shiming Yang, Wen
  Xie, Wenhao Huang, Xiaohui Hu, Xiaoyi Ren, Xinyao Niu, Pengcheng Nie, Yanpeng
  Li, Yuchi Xu, Yudong Liu, Yue Wang, Yuxuan Cai, Zhenyu Gu, Zhiyuan Liu, and
  Zonghong Dai. 2025.
\newblock \href {https://arxiv.org/abs/2403.04652} {Yi: Open foundation models
  by 01.ai}.
\newblock \emph{Preprint}, arXiv:2403.04652.

\bibitem[{Ali et~al.(2025)Ali, Katz, Wolf, and
  Titov}]{Ali2025Detectingpruningprominent}
Ameen Ali, Shahar Katz, Lior Wolf, and Ivan Titov. 2025.
\newblock \href {https://openreview.net/forum?id=cRE1XrHf1h} {Detecting and
  pruning prominent but detrimental neurons in large language models}.
\newblock \emph{COLM}.

\bibitem[{Arora et~al.(2026)Arora, Wu, Steinhardt, and
  Schwettmann}]{Arora2026Languagemodelcircuits}
Aryaman Arora, Zhengxuan Wu, Jacob Steinhardt, and Sarah Schwettmann. 2026.
\newblock \href {https://arxiv.org/abs/https://arxiv.org/pdf/2601.22594.pdf}
  {Language model circuits are sparse in the neuron basis}.
\newblock \emph{Proceedings of the 43rd International Conference on Machine
  Learning}.

\bibitem[{Belrose et~al.(2025)Belrose, Ostrovsky, McKinney, Furman, Smith,
  Halawi, Biderman, and Steinhardt}]{2023_Belrose}
Nora Belrose, Igor Ostrovsky, Lev McKinney, Zach Furman, Logan Smith, Danny
  Halawi, Stella Biderman, and Jacob Steinhardt. 2025.
\newblock \href {https://arxiv.org/abs/2303.08112} {Eliciting latent
  predictions from transformers with the tuned lens}.
\newblock \emph{Preprint}, arXiv:2303.08112.

\bibitem[{Biderman et~al.(2023)Biderman, Schoelkopf, Anthony, Bradley,
  O’Brien, Hallahan, Khan, Purohit, Prashanth, Raff, Skowron, Sutawika, and
  Wal}]{Biderman2023Pythia}
Stella Biderman, Hailey Schoelkopf, Quentin~Gregory Anthony, Herbie Bradley,
  Kyle O’Brien, Eric Hallahan, Mohammad~Aflah Khan, Shivanshu Purohit,
  Usvsn~Sai Prashanth, Edward Raff, Aviya Skowron, Lintang Sutawika, and Oskar
  Van~Der Wal. 2023.
\newblock \href {https://proceedings.mlr.press/v202/biderman23a.html} {Pythia:
  a suite for analyzing large language models across training and scaling}.
\newblock In \emph{Proceedings of the 40th International Conference on Machine
  Learning}.

\bibitem[{BigScience et~al.(2023)BigScience, :, Scao, Fan, Akiki, Pavlick,
  Ilić, Hesslow, Castagné, Luccioni, Yvon, Gallé, Tow, Rush, Biderman,
  Webson, Ammanamanchi, Wang, Sagot, Muennighoff, del Moral, Ruwase, Bawden,
  Bekman, McMillan-Major, Beltagy, Nguyen, Saulnier, Tan, Suarez, Sanh,
  Laurençon, Jernite, Launay, Mitchell, Raffel, Gokaslan, Simhi, Soroa, Aji,
  Alfassy, Rogers, Nitzav, Xu, Mou, Emezue, Klamm, Leong, van Strien, Adelani,
  Radev, Ponferrada, Levkovizh, Kim, Natan, Toni, Dupont, Kruszewski, Pistilli,
  Elsahar, Benyamina, Tran, Yu, Abdulmumin, Johnson, Gonzalez-Dios, de~la Rosa,
  Chim, Dodge, Zhu, Chang, Frohberg, Tobing, Bhattacharjee, Almubarak, Chen,
  Lo, Werra, Weber, Phan, allal, Tanguy, Dey, Muñoz, Masoud, Grandury,
  Šaško, Huang, Coavoux, Singh, Jiang, Vu, Jauhar, Ghaleb, Subramani,
  Kassner, Khamis, Nguyen, Espejel, de~Gibert, Villegas, Henderson, Colombo,
  Amuok, Lhoest, Harliman, Bommasani, López, Ribeiro, Osei, Pyysalo, Nagel,
  Bose, Muhammad, Sharma, Longpre, Nikpoor, Silberberg, Pai, Zink, Torrent,
  Schick, Thrush, Danchev, Nikoulina, Laippala, Lepercq, Prabhu, Alyafeai,
  Talat, Raja, Heinzerling, Si, Taşar, Salesky, Mielke, Lee, Sharma, Santilli,
  Chaffin, Stiegler, Datta, Szczechla, Chhablani, Wang, Pandey, Strobelt,
  Fries, Rozen, Gao, Sutawika, Bari, Al-shaibani, Manica, Nayak, Teehan,
  Albanie, Shen, Ben-David, Bach, Kim, Bers, Fevry, Neeraj, Thakker, Raunak,
  Tang, Yong, Sun, Brody, Uri, Tojarieh, Roberts, Chung, Tae, Phang, Press, Li,
  Narayanan, Bourfoune, Casper, Rasley, Ryabinin, Mishra, Zhang, Shoeybi,
  Peyrounette, Patry, Tazi, Sanseviero, von Platen, Cornette, Lavallée,
  Lacroix, Rajbhandari, Gandhi, Smith, Requena, Patil, Dettmers, Baruwa, Singh,
  Cheveleva, Ligozat, Subramonian, Névéol, Lovering, Garrette, Tunuguntla,
  Reiter, Taktasheva, Voloshina, Bogdanov, Winata, Schoelkopf, Kalo, Novikova,
  Forde, Clive, Kasai, Kawamura, Hazan, Carpuat, Clinciu, Kim, Cheng, Serikov,
  Antverg, van~der Wal, Zhang, Zhang, Gehrmann, Mirkin, Pais, Shavrina,
  Scialom, Yun, Limisiewicz, Rieser, Protasov, Mikhailov, Pruksachatkun,
  Belinkov, Bamberger, Kasner, Rueda, Pestana, Feizpour, Khan, Faranak, Santos,
  Hevia, Unldreaj, Aghagol, Abdollahi, Tammour, HajiHosseini, Behroozi,
  Ajibade, Saxena, Ferrandis, McDuff, Contractor, Lansky, David, Kiela, Nguyen,
  Tan, Baylor, Ozoani, Mirza, Ononiwu, Rezanejad, Jones, Bhattacharya,
  Solaiman, Sedenko, Nejadgholi, Passmore, Seltzer, Sanz, Dutra, Samagaio,
  Elbadri, Mieskes, Gerchick, Akinlolu, McKenna, Qiu, Ghauri, Burynok, Abrar,
  Rajani, Elkott, Fahmy, Samuel, An, Kromann, Hao, Alizadeh, Shubber, Wang,
  Roy, Viguier, Le, Oyebade, Le, Yang, Nguyen, Kashyap, Palasciano, Callahan,
  Shukla, Miranda-Escalada, Singh, Beilharz, Wang, Brito, Zhou, Jain, Xu,
  Fourrier, Periñán, Molano, Yu, Manjavacas, Barth, Fuhrimann, Altay, Bayrak,
  Burns, Vrabec, Bello, Dash, Kang, Giorgi, Golde, Posada, Sivaraman,
  Bulchandani, Liu, Shinzato, de~Bykhovetz, Takeuchi, Pàmies, Castillo,
  Nezhurina, Sänger, Samwald, Cullan, Weinberg, Wolf, Mihaljcic, Liu,
  Freidank, Kang, Seelam, Dahlberg, Broad, Muellner, Fung, Haller,
  Chandrasekhar, Eisenberg, Martin, Canalli, Su, Su, Cahyawijaya, Garda,
  Deshmukh, Mishra, Kiblawi, Ott, Sang-aroonsiri, Kumar, Schweter, Bharati,
  Laud, Gigant, Kainuma, Kusa, Labrak, Bajaj, Venkatraman, Xu, Xu, Xu, Tan,
  Xie, Ye, Bras, Belkada, and Wolf}]{BigScience2022BLOOM}
Workshop BigScience, :, Teven~Le Scao, Angela Fan, Christopher Akiki, Ellie
  Pavlick, Suzana Ilić, Daniel Hesslow, Roman Castagné, Alexandra~Sasha
  Luccioni, François Yvon, Matthias Gallé, Jonathan Tow, Alexander~M. Rush,
  Stella Biderman, Albert Webson, Pawan~Sasanka Ammanamanchi, Thomas Wang,
  Benoît Sagot, Niklas Muennighoff, Albert~Villanova del Moral, Olatunji
  Ruwase, Rachel Bawden, Stas Bekman, Angelina McMillan-Major, Iz~Beltagy, Huu
  Nguyen, Lucile Saulnier, Samson Tan, Pedro~Ortiz Suarez, Victor Sanh, Hugo
  Laurençon, Yacine Jernite, Julien Launay, Margaret Mitchell, Colin Raffel,
  Aaron Gokaslan, Adi Simhi, Aitor Soroa, Alham~Fikri Aji, Amit Alfassy, Anna
  Rogers, Ariel~Kreisberg Nitzav, Canwen Xu, Chenghao Mou, Chris Emezue,
  Christopher Klamm, Colin Leong, Daniel van Strien, David~Ifeoluwa Adelani,
  Dragomir Radev, Eduardo~González Ponferrada, Efrat Levkovizh, Ethan Kim,
  Eyal~Bar Natan, Francesco~De Toni, Gérard Dupont, Germán Kruszewski, Giada
  Pistilli, Hady Elsahar, Hamza Benyamina, Hieu Tran, Ian Yu, Idris Abdulmumin,
  Isaac Johnson, Itziar Gonzalez-Dios, Javier de~la Rosa, Jenny Chim, Jesse
  Dodge, Jian Zhu, Jonathan Chang, Jörg Frohberg, Joseph Tobing, Joydeep
  Bhattacharjee, Khalid Almubarak, Kimbo Chen, Kyle Lo, Leandro~Von Werra, Leon
  Weber, Long Phan, Loubna~Ben allal, Ludovic Tanguy, Manan Dey, Manuel~Romero
  Muñoz, Maraim Masoud, María Grandury, Mario Šaško, Max Huang, Maximin
  Coavoux, Mayank Singh, Mike Tian-Jian Jiang, Minh~Chien Vu, Mohammad~A.
  Jauhar, Mustafa Ghaleb, Nishant Subramani, Nora Kassner, Nurulaqilla Khamis,
  Olivier Nguyen, Omar Espejel, Ona de~Gibert, Paulo Villegas, Peter Henderson,
  Pierre Colombo, Priscilla Amuok, Quentin Lhoest, Rheza Harliman, Rishi
  Bommasani, Roberto~Luis López, Rui Ribeiro, Salomey Osei, Sampo Pyysalo,
  Sebastian Nagel, Shamik Bose, Shamsuddeen~Hassan Muhammad, Shanya Sharma,
  Shayne Longpre, Somaieh Nikpoor, Stanislav Silberberg, Suhas Pai, Sydney
  Zink, Tiago~Timponi Torrent, Timo Schick, Tristan Thrush, Valentin Danchev,
  Vassilina Nikoulina, Veronika Laippala, Violette Lepercq, Vrinda Prabhu, Zaid
  Alyafeai, Zeerak Talat, Arun Raja, Benjamin Heinzerling, Chenglei Si,
  Davut~Emre Taşar, Elizabeth Salesky, Sabrina~J. Mielke, Wilson~Y. Lee,
  Abheesht Sharma, Andrea Santilli, Antoine Chaffin, Arnaud Stiegler, Debajyoti
  Datta, Eliza Szczechla, Gunjan Chhablani, Han Wang, Harshit Pandey, Hendrik
  Strobelt, Jason~Alan Fries, Jos Rozen, Leo Gao, Lintang Sutawika, M.~Saiful
  Bari, Maged~S. Al-shaibani, Matteo Manica, Nihal Nayak, Ryan Teehan, Samuel
  Albanie, Sheng Shen, Srulik Ben-David, Stephen~H. Bach, Taewoon Kim, Tali
  Bers, Thibault Fevry, Trishala Neeraj, Urmish Thakker, Vikas Raunak, Xiangru
  Tang, Zheng-Xin Yong, Zhiqing Sun, Shaked Brody, Yallow Uri, Hadar Tojarieh,
  Adam Roberts, Hyung~Won Chung, Jaesung Tae, Jason Phang, Ofir Press, Conglong
  Li, Deepak Narayanan, Hatim Bourfoune, Jared Casper, Jeff Rasley, Max
  Ryabinin, Mayank Mishra, Minjia Zhang, Mohammad Shoeybi, Myriam Peyrounette,
  Nicolas Patry, Nouamane Tazi, Omar Sanseviero, Patrick von Platen, Pierre
  Cornette, Pierre~François Lavallée, Rémi Lacroix, Samyam Rajbhandari,
  Sanchit Gandhi, Shaden Smith, Stéphane Requena, Suraj Patil, Tim Dettmers,
  Ahmed Baruwa, Amanpreet Singh, Anastasia Cheveleva, Anne-Laure Ligozat, Arjun
  Subramonian, Aurélie Névéol, Charles Lovering, Dan Garrette, Deepak
  Tunuguntla, Ehud Reiter, Ekaterina Taktasheva, Ekaterina Voloshina, Eli
  Bogdanov, Genta~Indra Winata, Hailey Schoelkopf, Jan-Christoph Kalo,
  Jekaterina Novikova, Jessica~Zosa Forde, Jordan Clive, Jungo Kasai, Ken
  Kawamura, Liam Hazan, Marine Carpuat, Miruna Clinciu, Najoung Kim, Newton
  Cheng, Oleg Serikov, Omer Antverg, Oskar van~der Wal, Rui Zhang, Ruochen
  Zhang, Sebastian Gehrmann, Shachar Mirkin, Shani Pais, Tatiana Shavrina,
  Thomas Scialom, Tian Yun, Tomasz Limisiewicz, Verena Rieser, Vitaly Protasov,
  Vladislav Mikhailov, Yada Pruksachatkun, Yonatan Belinkov, Zachary Bamberger,
  Zdeněk Kasner, Alice Rueda, Amanda Pestana, Amir Feizpour, Ammar Khan, Amy
  Faranak, Ana Santos, Anthony Hevia, Antigona Unldreaj, Arash Aghagol, Arezoo
  Abdollahi, Aycha Tammour, Azadeh HajiHosseini, Bahareh Behroozi, Benjamin
  Ajibade, Bharat Saxena, Carlos~Muñoz Ferrandis, Daniel McDuff, Danish
  Contractor, David Lansky, Davis David, Douwe Kiela, Duong~A. Nguyen, Edward
  Tan, Emi Baylor, Ezinwanne Ozoani, Fatima Mirza, Frankline Ononiwu, Habib
  Rezanejad, Hessie Jones, Indrani Bhattacharya, Irene Solaiman, Irina Sedenko,
  Isar Nejadgholi, Jesse Passmore, Josh Seltzer, Julio~Bonis Sanz, Livia Dutra,
  Mairon Samagaio, Maraim Elbadri, Margot Mieskes, Marissa Gerchick, Martha
  Akinlolu, Michael McKenna, Mike Qiu, Muhammed Ghauri, Mykola Burynok, Nafis
  Abrar, Nazneen Rajani, Nour Elkott, Nour Fahmy, Olanrewaju Samuel, Ran An,
  Rasmus Kromann, Ryan Hao, Samira Alizadeh, Sarmad Shubber, Silas Wang, Sourav
  Roy, Sylvain Viguier, Thanh Le, Tobi Oyebade, Trieu Le, Yoyo Yang, Zach
  Nguyen, Abhinav~Ramesh Kashyap, Alfredo Palasciano, Alison Callahan, Anima
  Shukla, Antonio Miranda-Escalada, Ayush Singh, Benjamin Beilharz, Bo~Wang,
  Caio Brito, Chenxi Zhou, Chirag Jain, Chuxin Xu, Clémentine Fourrier,
  Daniel~León Periñán, Daniel Molano, Dian Yu, Enrique Manjavacas, Fabio
  Barth, Florian Fuhrimann, Gabriel Altay, Giyaseddin Bayrak, Gully Burns,
  Helena~U. Vrabec, Imane Bello, Ishani Dash, Jihyun Kang, John Giorgi, Jonas
  Golde, Jose~David Posada, Karthik~Rangasai Sivaraman, Lokesh Bulchandani,
  Lu~Liu, Luisa Shinzato, Madeleine~Hahn de~Bykhovetz, Maiko Takeuchi, Marc
  Pàmies, Maria~A. Castillo, Marianna Nezhurina, Mario Sänger, Matthias
  Samwald, Michael Cullan, Michael Weinberg, Michiel~De Wolf, Mina Mihaljcic,
  Minna Liu, Moritz Freidank, Myungsun Kang, Natasha Seelam, Nathan Dahlberg,
  Nicholas~Michio Broad, Nikolaus Muellner, Pascale Fung, Patrick Haller, Ramya
  Chandrasekhar, Renata Eisenberg, Robert Martin, Rodrigo Canalli, Rosaline Su,
  Ruisi Su, Samuel Cahyawijaya, Samuele Garda, Shlok~S. Deshmukh, Shubhanshu
  Mishra, Sid Kiblawi, Simon Ott, Sinee Sang-aroonsiri, Srishti Kumar, Stefan
  Schweter, Sushil Bharati, Tanmay Laud, Théo Gigant, Tomoya Kainuma, Wojciech
  Kusa, Yanis Labrak, Yash~Shailesh Bajaj, Yash Venkatraman, Yifan Xu, Yingxin
  Xu, Yu~Xu, Zhe Tan, Zhongli Xie, Zifan Ye, Mathilde Bras, Younes Belkada, and
  Thomas Wolf. 2023.
\newblock \href {https://arxiv.org/abs/2211.05100} {{BLOOM}: a 176{B}-parameter
  open-access multilingual language model}.
\newblock \emph{Preprint}, arXiv:2211.05100.

\bibitem[{Dai et~al.(2022)Dai, Dong, Hao, Sui, Chang, and
  Wei}]{dai-etal-2022-knowledge}
Damai Dai, Li~Dong, Yaru Hao, Zhifang Sui, Baobao Chang, and Furu Wei. 2022.
\newblock \href {https://doi.org/10.18653/v1/2022.acl-long.581} {Knowledge
  neurons in pretrained transformers}.
\newblock In \emph{Proceedings of the 60th Annual Meeting of the Association
  for Computational Linguistics (Volume 1: Long Papers)}, pages 8493--8502,
  Dublin, Ireland. Association for Computational Linguistics.

\bibitem[{Dar et~al.(2023)Dar, Geva, Gupta, and
  Berant}]{dar-etal-2023-analyzing}
Guy Dar, Mor Geva, Ankit Gupta, and Jonathan Berant. 2023.
\newblock \href {https://doi.org/10.18653/v1/2023.acl-long.893} {Analyzing
  transformers in embedding space}.
\newblock In \emph{Proceedings of the 61st Annual Meeting of the Association
  for Computational Linguistics (Volume 1: Long Papers)}, pages 16124--16170,
  Toronto, Canada. Association for Computational Linguistics.

\bibitem[{DeepSeek-AI et~al.(2026)DeepSeek-AI, Xu, Lin, Xue, Wang, Xu, Wu,
  Zhang, Lin, Dong, Ling, Lu, Zhao, Deng, Hou, Xu, Shao, Ruan, Sun, Dai, Guo,
  Yang, Chen, Li, Ji, Li, Wei, Lin, Yuan, Xia, Dai, Hao, Chen, Cao, Meng, Li,
  Yu, Zhang, Xu, Li, Liang, Zhang, Luo, Wei, Yuan, Zhang, Luo, Chen, Ji, Zhang,
  Ding, Tang, Cao, Gao, Qu, Zeng, Yang, Zhu, Luo, Song, Yu, Huang, Cai, Liang,
  Zhou, Ye, Li, Xu, Hu, Yang, Chen, Yan, Chen, Zhou, Xiang, Yuan, Cheng, Zhou,
  Zhu, Yu, Sun, Ran, Jiang, Qiu, Li, Zheng, Song, Dong, Gao, Guan, Zhou, Huang,
  Yu, Wang, Zhang, Wang, Xia, Zhang, Zhao, Guo, Luo, Ma, Zhu, Wang, Cai, Zhang,
  Chen, Di, Xu, Mei, Wang, Zhang, Zhang, Tang, Li, Zhou, Han, Wang, Huang,
  Wang, Cong, Wang, Zhang, Wang, Zhu, Li, Chen, Du, Jiang, Tian, Xu, Lu, Xu,
  Ge, Zhang, Pan, Wang, Chen, Yin, Xu, Shen, Zhang, Chen, Liu, Lu, Sun, Zhou,
  Chen, Cai, Nie, Wu, Chen, Hu, Liu, Hu, Ma, Wang, Yu, Zhou, Pan, Yu, Zhou, Ni,
  Yun, Jin, Pei, Ye, Lin, Ji, Cui, Yue, Yu, Wang, Zhang, Xiao, Zeng, An, Zhao,
  Liu, Liang, Pang, Luo, Yao, Gao, Yang, Huang, Hou, Zhang, Ma, Gao, He, Wang,
  Wang, Bi, Liu, Wang, Chen, Zhang, Nie, Sun, Wang, Cheng, Liu, Xie, Liu, Liu,
  Yu, Li, Yang, Zhang, Chen, Wang, Su, Chen, Lin, Fu, Yan, Wang, Ma, Luo,
  Zhang, Xu, Ma, Huang, Li, Xu, Zhao, Sun, Wang, Qian, Shao, Yu, Zhang, Ding,
  Shi, Wu, Xiong, Ma, He, Tang, Zhou, Luo, Zhong, Piao, Wang, Zhang, Chen, Tan,
  Wei, Ma, Liu, Yang, Guo, Wu, Wu, Li, Cheng, Ou, Xu, Li, Wang, Yang, Xu, Wu,
  Meng, Zou, Zha, Xiong, Chen, Lin, Cao, Wang, Zhang, Yan, Lin, Gu, Luo, You,
  Liu, Zhou, Zhou, Huang, Wu, Wang, Zhao, Ren, Zhang, Sha, Fu, Ju, Xu, Xie,
  Zhang, Gao, Hao, Gou, Ma, Yan, Shao, Huang, Chen, Wu, Ren, Wu, Li, Zhang, Xu,
  Wang, Qu, Gu, Zhu, Li, Zhang, Xie, Gao, Wan, Pan, and Yao}]{2026DeepSeekV4}
DeepSeek-AI, Anyi Xu, Bang Lin, Bing Xue, Bing-Li Wang, Bin Xu, Bo~Wu, Bowei
  Zhang, Chao Lin, Chengyao Dong, Chen Ling, Chengda Lu, Chen Zhao, Chengqi
  Deng, Cheng Hou, Chen Xu, Chenze Shao, Chong Ruan, Conner Sun, Damai Dai,
  Daya Guo, De-Bin Yang, Deli Chen, Dong-Hui Li, Dong-Li Ji, Erhang Li,
  Fangchen Wei, Fangyun Lin, Fang Yuan, Fei Xia, Fucong Dai, Guangbo Hao,
  Guanting Chen, Guo Cao, Guo-Hui Meng, Guowei Li, Han Yu, Han Zhang, Hanwei
  Xu, Hao Li, Hao Liang, Haoling Zhang, Haoming Luo, Haoran Wei, Hao Yuan,
  Haowei Zhang, Haowen Luo, Hao Chen, Haozhe Ji, Heng Zhang, Honghui Ding,
  Hongxuan Tang, Huanqi Cao, Huazuo Gao, Hui Qu, Hui Zeng, J.~Yang, J.~Q. Zhu,
  Jianming Luo, Jia Song, Jia Yu, Jialiang Huang, Jialu Cai, Jian Liang,
  Jiangting Zhou, Jiasheng Ye, Jiashi Li, Jiaxin Xu, Jiewen Hu, Jieyu Yang, Jin
  Chen, Jin Yan, JingChang Chen, Jing Zhou, Jing Xiang, Jingyang Yuan, Jing
  Cheng, Jingzi Zhou, Jinhua Zhu, Jiping Yu, Joseph Sun, Junliang Ran, Jun
  Jiang, Junjie Qiu, Junlong Li, Junming Zheng, Jun-Mei Song, Kai Dong, Kaige
  Gao, Kang Guan, Kexing Zhou, Kezhao Huang, Kuai Yu, Lean Wang, Lecong Zhang,
  Lei Wang, Leyi Xia, Li~Zhang, Liang Zhao, Lihua Guo, Lin Luo, Lin Ma, Linyan
  Zhu, Litong Wang, Liyu Cai, Liyue Zhang, Longhao Chen, Mingxi Di, My~Xu, Max
  Mei, Miaojun Wang, Mingchuan Zhang, Minghua Zhang, Minghui Tang, Mingming Li,
  Mi~Zhou, Min Han, Ning Wang, Pan Huang, Pan Wang, Peixin Cong, Peiyi Wang,
  Peng Zhang, Qiancheng Wang, Qihao Zhu, Qingyang Li, Qinyu Chen, Qiushi Du,
  Qi~Jiang, Rui Tian, Rui-qin Xu, Ruijie Lu, Ruiling Xu, Ruiqi Ge, Ruisong
  Zhang, Ruizhe Pan, Runji Wang, Run~Da Chen, Runqiu Yin, Runxin Xu, Ruo-Han
  Shen, Ruoyu Zhang, Ruyi Chen, Sh. Liu, Shanghao Lu, Shangmian Sun, Shangyan
  Zhou, Shan-Shan Chen, Shaofei Cai, Shao Nie, Shao-Ping Wu, Shaoyuan Chen,
  Shengding Hu, Sheng Liu, Shiqiang Hu, Shirong Ma, Shiyu Wang, Shuiping Yu,
  Shunfeng Zhou, Shuting Pan, Shuying Yu, Songyang Zhou, Tao Ni, Tao Yun, Tian
  Jin, Tianhong Pei, Tian Ye, Tianle Lin, Tian Ji, Tianyi Cui, Tianyuan Yue,
  Tingting Yu, Tu-Tu Wang, W.~Y. Zhang, Wl~Xiao, Wangding Zeng, Wei An, Weilin
  Zhao, Wen Liu, Wenfeng Liang, Wenjie Pang, Wenjing Luo, Wenjin Yao, Wenjun
  Gao, Wenkai Yang, Wen-Wen Huang, Wenqing Hou, Wentao Zhang, Wenting Ma,
  Xi~Gao, Xiang He, Xiang Wang, Xianzu Wang, Xiao Bi, Xiaodong Liu, Xiaohan
  Wang, Xiaokang Chen, Xiaokang Zhang, Xiaotao Nie, Xiaowen Sun, Xiaoxiang
  Wang, Xin Cheng, Xin Liu, Xin Xie, Xingchao Liu, Xingchen Liu, Xingkai Yu,
  Xingyou Li, Xinyu Yang, Xinyu Zhang, Xu~Chen, Xuanyu Wang, Xuecheng Su,
  Xueyin Chen, Xuheng Lin, Xu~Fu, Yc~Yan, Y.~Q. Wang, Yw~Ma, Yanfen Luo, Yang
  Zhang, Yanhong Xu, Yanru Ma, Yanwen Huang, Yao Li, Yao Xu, Yao Zhao, Yaofeng
  Sun, Yaohui Wang, Yi~Qian, Yingxia Shao, Yi~Yu, Yichao Zhang, Yifan Ding,
  Yifan Shi, Yijia Wu, Yi-Yu Xiong, Yi~Ma, Ying He, Ying Tang, Ying Zhou,
  Yingjia Luo, Yinmin Zhong, Yishi Piao, Yisong Wang, Yixiang Zhang, Yixiao
  Chen, Yixuan Tan, Yixuan Wei, Yiyang Ma, Yiyuan Liu, Yong Yang, Yongqiang
  Guo, Yongtong Wu, Yu~Wu, Yukun Li, Yuan Cheng, Yuan Ou, Yuanfan Xu, Yuanhao
  Li, Yuduan Wang, Yuehan Yang, Yue Xu, Yuhan Wu, Yu~Meng, Yu~Zou, Yukun Zha,
  Yunfan Xiong, Yupeng Chen, Yuping Lin, Yu~Cao, Yuqian Wang, Yushun Zhang,
  Yuting Yan, Yutong Lin, Yuxian Gu, Yu-Wei Luo, Yu-mei You, Yuxuan Liu, Yuxuan
  Zhou, Yuyang Zhou, Yuzhen Huang, Zhanghua Wu, Ze~Wang, Ze~Zhao, Zehui Ren,
  Zekai Zhang, Zhangli Sha, Zhe Fu, Zhe Ju, Zhean Xu, Zhenda Xie, Zhengyan
  Zhang, Zhe Gao, Zhewen Hao, Zhibin Gou, Zhicheng Ma, Zhigang Yan, Zhihong
  Shao, Zhixiang Huang, Zhixuan Chen, Zhiyu Wu, Zhizhou Ren, Zhongyu Wu,
  Zhuoshu Li, Zhuping Zhang, Zian Xu, Zihao Wang, Zihua Qu, Zihui Gu, Zijia
  Zhu, Zi-Long Li, Zipeng Zhang, Ziwei Xie, Ziyi Gao, Ziyi Wan, Zizheng Pan,
  and Zong-xiang Yao. 2026.
\newblock \href {https://arxiv.org/abs/2606.19348} {Deep{S}eek-{V}4: towards
  highly efficient million-token context intelligence}.
\newblock \emph{Preprint}, arXiv:2606.19348.

\bibitem[{Dettmers et~al.(2022)Dettmers, Lewis, Belkada, and
  Zettlemoyer}]{Dettmers2022}
Tim Dettmers, Mike Lewis, Younes Belkada, and Luke Zettlemoyer. 2022.
\newblock \href
  {https://proceedings.neurips.cc/paper_files/paper/2022/file/c3ba4962c05c49636d4c6206a97e9c8a-Paper-Conference.pdf}
  {{LLM}.int8(): 8-bit matrix multiplication for {T}ransformers at scale}.
\newblock \emph{Advances in Neural Information Processing Systems}.

\bibitem[{Devlin et~al.(2019)Devlin, Chang, Lee, and
  Toutanova}]{devlin-etal-2019-bert}
Jacob Devlin, Ming-Wei Chang, Kenton Lee, and Kristina Toutanova. 2019.
\newblock \href {https://doi.org/10.18653/v1/N19-1423} {{BERT}: Pre-training of
  deep bidirectional transformers for language understanding}.
\newblock In \emph{Proceedings of the 2019 Conference of the North {A}merican
  Chapter of the Association for Computational Linguistics: Human Language
  Technologies, Volume 1 (Long and Short Papers)}, pages 4171--4186,
  Minneapolis, Minnesota. Association for Computational Linguistics.

\bibitem[{Dunefsky et~al.(2024)Dunefsky, Chlenski, and
  Nanda}]{Dunefsky2024Transcodersfindinterpretable}
Jacob Dunefsky, Philippe Chlenski, and Neel Nanda. 2024.
\newblock \href {https://doi.org/10.52202/079017-0768} {Transcoders find
  interpretable {LLM} feature circuits}.
\newblock In \emph{Advances in Neural Information Processing Systems},
  volume~37, pages 24375--24410. Curran Associates, Inc.

\bibitem[{Elhage et~al.(2022)Elhage, Hume, Olsson, Schiefer, Henighan, Kravec,
  Hatfield-Dodds, Lasenby, Drain, Chen, Grosse, McCandlish, Kaplan, Amodei,
  Wattenberg, and Olah}]{Elhage2022Toymodelssuperposition}
Nelson Elhage, Tristan Hume, Catherine Olsson, Nicholas Schiefer, Tom Henighan,
  Shauna Kravec, Zac Hatfield-Dodds, Robert Lasenby, Dawn Drain, Carol Chen,
  Roger Grosse, Sam McCandlish, Jared Kaplan, Dario Amodei, Martin Wattenberg,
  and Christopher Olah. 2022.
\newblock \href {https://transformer-circuits.pub/2022/toy_model/index.html}
  {Toy models of superposition}.
\newblock Transformer Circuits Thread.

\bibitem[{Elhage et~al.(2021)Elhage, Nanda, Olsson, Henighan, Joseph, Mann,
  Askell, Bai, Chen, Conerly, DasSarma, Drain, Ganguli, Hatfield-Dodds,
  Hernandez, Jones, Kernion, Lovitt, Ndousse, Amodei, Brown, Clark, Kaplan,
  McCandlish, and Olah}]{Elhage2021mathematicalframeworktransformer}
Nelson Elhage, Neel Nanda, Catherine Olsson, Tom Henighan, Nicholas Joseph, Ben
  Mann, Amanda Askell, Yuntao Bai, Anna Chen, Tom Conerly, Nova DasSarma, Dawn
  Drain, Deep Ganguli, Zac Hatfield-Dodds, Danny Hernandez, Andy Jones, Jackson
  Kernion, Liane Lovitt, Kamal Ndousse, Dario Amodei, Tom Brown, Jack Clark,
  Jared Kaplan, Sam McCandlish, and Chris Olah. 2021.
\newblock \href {https://transformer-circuits.pub/2021/framework/index.html} {A
  mathematical framework for transformer circuits}.
\newblock Transformer Circuits Thread.

\bibitem[{Elhelo and Geva(2025)}]{elhelo-geva-2025-inferring}
Amit Elhelo and Mor Geva. 2025.
\newblock \href {https://doi.org/10.18653/v1/2025.acl-long.866} {Inferring
  functionality of attention heads from their parameters}.
\newblock In \emph{Proceedings of the 63rd Annual Meeting of the Association
  for Computational Linguistics (Volume 1: Long Papers)}, pages 17701--17733,
  Vienna, Austria. Association for Computational Linguistics.

\bibitem[{Ethayarajh(2019)}]{ethayarajh-2019-contextual}
Kawin Ethayarajh. 2019.
\newblock \href {https://doi.org/10.18653/v1/D19-1006} {How contextual are
  contextualized word representations? {C}omparing the geometry of {BERT},
  {ELM}o, and {GPT}-2 embeddings}.
\newblock In \emph{Proceedings of the 2019 Conference on Empirical Methods in
  Natural Language Processing and the 9th International Joint Conference on
  Natural Language Processing (EMNLP-IJCNLP)}, pages 55--65, Hong Kong, China.
  Association for Computational Linguistics.

\bibitem[{Gemma(2024)}]{gemma_2024}
Team Gemma. 2024.
\newblock \href {https://doi.org/10.34740/KAGGLE/M/3301} {Gemma}.
\newblock Kaggle.

\bibitem[{Gemma(2025)}]{Gemma3}
Team Gemma. 2025.
\newblock \href
  {https://storage.googleapis.com/deepmind-media/gemma/Gemma3Report.pdf} {Gemma
  3 technical report}.
\newblock Technical report, Google DeepMind.

\bibitem[{Geva et~al.(2023)Geva, Bastings, Filippova, and
  Globerson}]{geva-etal-2023-dissecting}
Mor Geva, Jasmijn Bastings, Katja Filippova, and Amir Globerson. 2023.
\newblock \href {https://doi.org/10.18653/v1/2023.emnlp-main.751} {Dissecting
  recall of factual associations in auto-regressive language models}.
\newblock In \emph{Proceedings of the 2023 Conference on Empirical Methods in
  Natural Language Processing}, pages 12216--12235, Singapore. Association for
  Computational Linguistics.

\bibitem[{Geva et~al.(2022)Geva, Caciularu, Wang, and
  Goldberg}]{geva-etal-2022-transformer}
Mor Geva, Avi Caciularu, Kevin Wang, and Yoav Goldberg. 2022.
\newblock \href {https://doi.org/10.18653/v1/2022.emnlp-main.3} {Transformer
  feed-forward layers build predictions by promoting concepts in the vocabulary
  space}.
\newblock In \emph{Proceedings of the 2022 Conference on Empirical Methods in
  Natural Language Processing}, pages 30--45, Abu Dhabi, United Arab Emirates.
  Association for Computational Linguistics.

\bibitem[{Geva et~al.(2021)Geva, Schuster, Berant, and
  Levy}]{geva-etal-2021-transformer}
Mor Geva, Roei Schuster, Jonathan Berant, and Omer Levy. 2021.
\newblock \href {https://doi.org/10.18653/v1/2021.emnlp-main.446} {Transformer
  feed-forward layers are key-value memories}.
\newblock In \emph{Proceedings of the 2021 Conference on Empirical Methods in
  Natural Language Processing}, pages 5484--5495, Online and Punta Cana,
  Dominican Republic. Association for Computational Linguistics.

\bibitem[{Groeneveld et~al.(2024)Groeneveld, Beltagy, Walsh, Bhagia, Kinney,
  Tafjord, Jha, Ivison, Magnusson, Wang, Arora, Atkinson, Authur, Chandu,
  Cohan, Dumas, Elazar, Gu, Hessel, Khot, Merrill, Morrison, Muennighoff, Naik,
  Nam, Peters, Pyatkin, Ravichander, Schwenk, Shah, Smith, Strubell, Subramani,
  Wortsman, Dasigi, Lambert, Richardson, Zettlemoyer, Dodge, Lo, Soldaini,
  Smith, and Hajishirzi}]{groeneveld-etal-2024-olmo}
Dirk Groeneveld, Iz~Beltagy, Evan Walsh, Akshita Bhagia, Rodney Kinney, Oyvind
  Tafjord, Ananya Jha, Hamish Ivison, Ian Magnusson, Yizhong Wang, Shane Arora,
  David Atkinson, Russell Authur, Khyathi Chandu, Arman Cohan, Jennifer Dumas,
  Yanai Elazar, Yuling Gu, Jack Hessel, Tushar Khot, William Merrill, Jacob
  Morrison, Niklas Muennighoff, Aakanksha Naik, Crystal Nam, Matthew Peters,
  Valentina Pyatkin, Abhilasha Ravichander, Dustin Schwenk, Saurabh Shah,
  William Smith, Emma Strubell, Nishant Subramani, Mitchell Wortsman, Pradeep
  Dasigi, Nathan Lambert, Kyle Richardson, Luke Zettlemoyer, Jesse Dodge, Kyle
  Lo, Luca Soldaini, Noah Smith, and Hannaneh Hajishirzi. 2024.
\newblock \href {https://doi.org/10.18653/v1/2024.acl-long.841} {{OLM}o:
  Accelerating the science of language models}.
\newblock In \emph{Proceedings of the 62nd Annual Meeting of the Association
  for Computational Linguistics (Volume 1: Long Papers)}, pages 15789--15809,
  Bangkok, Thailand. Association for Computational Linguistics.

\bibitem[{Gur-Arieh et~al.(2025)Gur-Arieh, Mayan, Agassy, Geiger, and
  Geva}]{gur-arieh-etal-2025-enhancing}
Yoav Gur-Arieh, Roy Mayan, Chen Agassy, Atticus Geiger, and Mor Geva. 2025.
\newblock \href {https://doi.org/10.18653/v1/2025.acl-long.288} {Enhancing
  automated interpretability with output-centric feature descriptions}.
\newblock In \emph{Proceedings of the 63rd Annual Meeting of the Association
  for Computational Linguistics (Volume 1: Long Papers)}, pages 5757--5778,
  Vienna, Austria. Association for Computational Linguistics.

\bibitem[{Gurnee et~al.(2024)Gurnee, Horsley, Guo, Kheirkhah, Sun, Hathaway,
  Nanda, and Bertsimas}]{2024_Gurnee}
Wes Gurnee, Theo Horsley, Zifan~Carl Guo, Tara~Rezaei Kheirkhah, Qinyi Sun,
  Will Hathaway, Neel Nanda, and Dimitris Bertsimas. 2024.
\newblock \href {https://openreview.net/forum?id=ZeI104QZ8I} {Universal neurons
  in {GPT}2 language models}.
\newblock \emph{Transactions of Machine Learning Research}.

\bibitem[{Gurnee et~al.(2023)Gurnee, Nanda, Pauly, Harvey, Troitskii, and
  Bertsimas}]{2023_Gurnee}
Wes Gurnee, Neel Nanda, Matthew Pauly, Katherine Harvey, Dmitrii Troitskii, and
  Dimitris Bertsimas. 2023.
\newblock \href {https://openreview.net/forum?id=JYs1R9IMJr} {Finding neurons
  in a haystack: case studies with sparse probing}.
\newblock \emph{Transactions of Macine Learning Research}.

\bibitem[{He et~al.(2016)He, Zhang, Ren, and Sun}]{He2016Deepresiduallearning}
Kaiming He, Xiangyu Zhang, Shaoqing Ren, and Jian Sun. 2016.
\newblock \href {https://doi.org/10.1109/CVPR.2016.90} {Deep residual learning
  for image recognition}.
\newblock In \emph{2016 IEEE Conference on Computer Vision and Pattern
  Recognition (CVPR)}, pages 770--778.

\bibitem[{Heimersheim and Turner(2023)}]{2023_Heimersheim}
Stefan Heimersheim and Alex Turner. 2023.
\newblock \href
  {https://www.alignmentforum.org/posts/8mizBCm3dyc432nK8/residual-stream-norms-grow-exponentially-over-the-forward}
  {Residual stream norms grow exponentially over the forward pass}.

\bibitem[{Hendrycks and Gimpel(2023)}]{Hendrycks2016}
Dan Hendrycks and Kevin Gimpel. 2023.
\newblock \href {https://arxiv.org/abs/1606.08415} {Bridging nonlinearities and
  stochastic regularizers with gaussian error linear units}.
\newblock \emph{Preprint}, arXiv:1606.08415.

\bibitem[{Janiak et~al.(2024)Janiak, Rager, Dao, and
  Lau}]{janiak-etal-2024-adversarial}
Jett Janiak, Can Rager, James Dao, and Yeu-Tong Lau. 2024.
\newblock \href {https://doi.org/10.18653/v1/2024.blackboxnlp-1.15} {An
  adversarial example for direct logit attribution: Memory management in
  {GELU}-4{L}}.
\newblock In \emph{Proceedings of the 7th BlackboxNLP Workshop: Analyzing and
  Interpreting Neural Networks for NLP}, pages 232--237, Miami, Florida, US.
  Association for Computational Linguistics.

\bibitem[{Ji et~al.(2026)Ji, Martens, Zheng, Zhou, Moghadam, Zhang,
  Saratchandran, and Lucey}]{Ji2026Cuttingskip}
Yiping Ji, James Martens, Jianqiao Zheng, Ziqin Zhou, Peyman Moghadam, Xinyu
  Zhang, Hemanth Saratchandran, and Simon Lucey. 2026.
\newblock \href {https://openreview.net/forum?id=iJl3L059s6} {Cutting the skip:
  training residual-free transformers}.
\newblock In \emph{The Fourteenth International Conference on Learning
  Representations}.

\bibitem[{Jiang et~al.(2023)Jiang, Sablayrolles, Mensch, Bamford, Chaplot,
  de~las Casas, Bressand, Lengyel, Lample, Saulnier, Lavaud, Lachaux, Stock,
  Scao, Lavril, Wang, Lacroix, and Sayed}]{jiang2023mistral7b}
Albert~Q. Jiang, Alexandre Sablayrolles, Arthur Mensch, Chris Bamford,
  Devendra~Singh Chaplot, Diego de~las Casas, Florian Bressand, Gianna Lengyel,
  Guillaume Lample, Lucile Saulnier, Lélio~Renard Lavaud, Marie-Anne Lachaux,
  Pierre Stock, Teven~Le Scao, Thibaut Lavril, Thomas Wang, Timothée Lacroix,
  and William~El Sayed. 2023.
\newblock \href {https://arxiv.org/abs/2310.06825} {Mistral 7{B}}.
\newblock \emph{Preprint}, arXiv:2310.06825.

\bibitem[{Joshi et~al.(2025)Joshi, Ahmad, and
  Modi}]{joshi-etal-2025-calibration}
Abhinav Joshi, Areeb Ahmad, and Ashutosh Modi. 2025.
\newblock \href {https://doi.org/10.18653/v1/2025.emnlp-main.742} {Calibration
  across layers: Understanding calibration evolution in {LLM}s}.
\newblock In \emph{Proceedings of the 2025 Conference on Empirical Methods in
  Natural Language Processing}, pages 14697--14725, Suzhou, China. Association
  for Computational Linguistics.

\bibitem[{Kadavath et~al.(2022)Kadavath, Conerly, Askell, Henighan, Drain,
  Perez, Schiefer, Hatfield-Dodds, DasSarma, Tran-Johnson, Johnston, El-Showk,
  Jones, Elhage, Hume, Chen, Bai, Bowman, Fort, Ganguli, Hernandez, Jacobson,
  Kernion, Kravec, Lovitt, Ndousse, Olsson, Ringer, Amodei, Brown, Clark,
  Joseph, Mann, McCandlish, Olah, and
  Kaplan}]{Kadavath2022Languagemodelsmostly}
Saurav Kadavath, Tom Conerly, Amanda Askell, Tom Henighan, Dawn Drain, Ethan
  Perez, Nicholas Schiefer, Zac Hatfield-Dodds, Nova DasSarma, Eli
  Tran-Johnson, Scott Johnston, Sheer El-Showk, Andy Jones, Nelson Elhage,
  Tristan Hume, Anna Chen, Yuntao Bai, Sam Bowman, Stanislav Fort, Deep
  Ganguli, Danny Hernandez, Josh Jacobson, Jackson Kernion, Shauna Kravec,
  Liane Lovitt, Kamal Ndousse, Catherine Olsson, Sam Ringer, Dario Amodei, Tom
  Brown, Jack Clark, Nicholas Joseph, Ben Mann, Sam McCandlish, Chris Olah, and
  Jared Kaplan. 2022.
\newblock \href {https://arxiv.org/abs/2207.05221} {Language models (mostly)
  know what they know}.
\newblock \emph{Preprint}, arXiv:2207.05221.

\bibitem[{Kantamneni et~al.(2025)Kantamneni, Engels, Rajamanoharan, Tegmark,
  and Nanda}]{Kantamneni2025}
Subhash Kantamneni, Joshua Engels, Senthooran Rajamanoharan, Max Tegmark, and
  Neel Nanda. 2025.
\newblock \href
  {https://raw.githubusercontent.com/mlresearch/v267/main/assets/kantamneni25a/kantamneni25a.pdf}
  {Are sparse autoencoders useful? {A} case study in sparse probing}.
\newblock In \emph{Proceedings of the 42nd International Conference on Machine
  Learning}.

\bibitem[{Kong et~al.(2026)Kong, Ning, Adler, and
  Shavit}]{Kong2026Negativepreactivations}
Linghao Kong, Angelina Ning, Micah Adler, and Nir Shavit. 2026.
\newblock \href {https://openreview.net/forum?id=RzcCrU0tXP} {Negative
  pre-activations differentiate syntax}.
\newblock In \emph{The Fourteenth International Conference on Learning
  Representations}.

\bibitem[{Kovaleva et~al.(2021)Kovaleva, Kulshreshtha, Rogers, and
  Rumshisky}]{kovaleva-etal-2021-bert}
Olga Kovaleva, Saurabh Kulshreshtha, Anna Rogers, and Anna Rumshisky. 2021.
\newblock \href {https://doi.org/10.18653/v1/2021.findings-acl.300} {{BERT}
  busters: Outlier dimensions that disrupt transformers}.
\newblock In \emph{Findings of the Association for Computational Linguistics:
  ACL-IJCNLP 2021}, pages 3392--3405, Online. Association for Computational
  Linguistics.

\bibitem[{Lad et~al.(2024)Lad, Gurnee, and
  Tegmark}]{lad2024remarkablerobustnessllmsstages}
Vedang Lad, Wes Gurnee, and Max Tegmark. 2024.
\newblock \href {https://arxiv.org/abs/2406.19384} {The remarkable robustness
  of {LLM}s: stages of inference?}
\newblock \emph{Preprint}, arXiv:2406.19384.

\bibitem[{Leask et~al.(2025)Leask, Bussmann, Pearce, Bloom, Tigges, Moubayed,
  Sharkey, and Nanda}]{Leask2025}
Patrick Leask, Bart Bussmann, Michael Pearce, Joseph Bloom, Curt Tigges,
  Noura~Al Moubayed, Lee Sharkey, and Neel Nanda. 2025.
\newblock \href {https://openreview.net/forum?id=9ca9eHNrdH} {Sparse
  autoencoders do not find canonical units of analysis}.
\newblock In \emph{The Thirteenth International Conference on Learning
  Representations}.

\bibitem[{Lee(2023)}]{Lee2023}
Minhyeok Lee. 2023.
\newblock \href {https://doi.org/10.1155/2023/4229924} {Mathematical analysis
  and performance evaluation of the {GELU} activation function in deep
  learning}.
\newblock \emph{Journal of Mathematics}.

\bibitem[{Li et~al.(2023)Li, Hopkins, Bau, Viégas, Pfister, and
  Wattenberg}]{Li2023Emergentworldrepresentations}
Kenneth Li, Aspen~K. Hopkins, David Bau, Fernanda Viégas, Hanspeter Pfister,
  and Martin Wattenberg. 2023.
\newblock \href {https://openreview.net/forum?id=DeG07_TcZvT} {Emergent world
  representations: Exploring a sequence model trained on a synthetic task}.
\newblock In \emph{The Eleventh International Conference on Learning
  Representations}.

\bibitem[{Lv et~al.(2024)Lv, Chen, Zhang, Wang, Liu, Wen, Xie, and
  Yan}]{2024_Lv}
Ang Lv, Yuhan Chen, Kaiyi Zhang, Yulong Wang, Lifeng Liu, Ji-Rong Wen, Jian
  Xie, and Rui Yan. 2024.
\newblock \href {https://arxiv.org/abs/2403.19521} {Interpreting key mechanisms
  of factual recall in transformer-based language models}.
\newblock \emph{Preprint}, arXiv:2403.19521.

\bibitem[{McDougall et~al.(2024)McDougall, Conmy, Rushing, McGrath, and
  Nanda}]{mcdougall-etal-2024-copy}
Callum~Stuart McDougall, Arthur Conmy, Cody Rushing, Thomas McGrath, and Neel
  Nanda. 2024.
\newblock \href {https://doi.org/10.18653/v1/2024.blackboxnlp-1.22} {Copy
  suppression: Comprehensively understanding a motif in language model
  attention heads}.
\newblock In \emph{Proceedings of the 7th BlackboxNLP Workshop: Analyzing and
  Interpreting Neural Networks for NLP}, pages 337--363, Miami, Florida, US.
  Association for Computational Linguistics.

\bibitem[{McGrath et~al.(2023)McGrath, Rahtz, Kramár, Mikulik, and
  Legg}]{McGrath2023hydraeffect}
Thomas McGrath, Matthew Rahtz, János Kramár, Vladimir Mikulik, and Shane
  Legg. 2023.
\newblock \href {https://arxiv.org/abs/2307.15771} {The hydra effect: emergent
  self-repair in language model computations}.
\newblock \emph{Preprint}, arXiv:2307.15771.

\bibitem[{Meta(2024{\natexlab{a}})}]{Llama3.1}
Meta. 2024{\natexlab{a}}.
\newblock \href {https://huggingface.co/collections/meta-llama/llama-31} {Llama
  3.1}.
\newblock Huggingface collection.

\bibitem[{Meta(2024{\natexlab{b}})}]{Llama3.2}
Meta. 2024{\natexlab{b}}.
\newblock \href {https://huggingface.co/collections/meta-llama/llama-32} {Llama
  3.2}.
\newblock Huggingface collection.

\bibitem[{Miller and Neo(2023)}]{2023_Miller}
Joseph Miller and Clement Neo. 2023.
\newblock \href
  {https://www.lesswrong.com/posts/cgqh99SHsCv3jJYDS/we-found-an-neuron-in-gpt-2}
  {We found an neuron in {GPT}-2}.

\bibitem[{Millidge and Black(2022)}]{2022_Millidge}
Beren Millidge and Sid Black. 2022.
\newblock \href
  {https://www.lesswrong.com/posts/mkbGjzxD8d8XqKHzA/the-singular-value-decompositions-of-transformer-weight}
  {The singular value decompositions of transformer weight matrices are highly
  interpretable}.

\bibitem[{Mueller et~al.(2025)Mueller, Geiger, Wiegreffe, Arad, Arcuschin,
  Belfki, Chan, Fiotto-Kaufman, Haklay, Hanna, Huang, Gupta, Nikankin, Orgad,
  Prakash, Reusch, Sankaranarayanan, Shao, Stolfo, Tutek, Zur, Bau, and
  Belinkov}]{Mueller2025}
Aaron Mueller, Atticus Geiger, Sarah Wiegreffe, Dana Arad, Iván Arcuschin,
  Adam Belfki, Yik~Siu Chan, Jaden Fiotto-Kaufman, Tal Haklay, Michael Hanna,
  Jing Huang, Rohan Gupta, Yaniv Nikankin, Hadas Orgad, Nikhil Prakash, Anja
  Reusch, Aruna Sankaranarayanan, Shun Shao, Alessandro Stolfo, Martin Tutek,
  Amir Zur, David Bau, and Yonatan Belinkov. 2025.
\newblock \href
  {https://raw.githubusercontent.com/mlresearch/v267/main/assets/mueller25a/mueller25a.pdf}
  {{MIB}: a mechanistic interpretability benchmark}.
\newblock In \emph{Proceedings of the 42nd International Conference on Machine
  Learning}.

\bibitem[{Nanda(2022)}]{Nanda2022}
Neel Nanda. 2022.
\newblock \href {https://neuroscope.io} {Neuroscope}.
\newblock Website.

\bibitem[{Nanda and Bloom(2022)}]{nanda2022transformerlens}
Neel Nanda and Joseph Bloom. 2022.
\newblock \href
  {https://transformerlensorg.github.io/TransformerLens/generated/code/transformer_lens.html}
  {Transformer{L}ens}.
\newblock \url{https://github.com/TransformerLensOrg/TransformerLens}.

\bibitem[{Niu et~al.(2024)Niu, Liu, Zu, and Penn}]{2024_Niu}
Jingcheng Niu, Andrew Liu, Zining Zu, and Gerald Penn. 2024.
\newblock \href {https://openreview.net/forum?id=2HJRwwbV3G} {What does the
  knowledge neuron thesis have to do with knowledge?}
\newblock In \emph{The Twelfth International Conference on Learning
  Representations}.

\bibitem[{nostalgebraist(2020)}]{2020_LogitLens}
nostalgebraist. 2020.
\newblock \href
  {https://www.lesswrong.com/posts/AcKRB8wDpdaN6v6ru/interpreting-gpt-the-logit-lens}
  {Interpreting {GPT}: The logit lens}.

\bibitem[{Olmo et~al.(2026)Olmo, :, Ettinger, Bertsch, Kuehl, Graham, Heineman,
  Groeneveld, Brahman, Timbers, Ivison, Morrison, Poznanski, Lo, Soldaini,
  Jordan, Chen, Noukhovitch, Lambert, Walsh, Dasigi, Berry, Malik, Shah, Geng,
  Arora, Gupta, Anderson, Xiao, Murray, Romero, Graf, Asai, Bhagia, Wettig,
  Liu, Rangapur, Anastasiades, Huang, Schwenk, Trivedi, Magnusson, Lochner,
  Liu, Miranda, Sap, Morgan, Schmitz, Guerquin, Wilson, Huff, Bras, Xin, Shao,
  Skjonsberg, Shen, Li, Wilde, Pyatkin, Merrill, Chang, Gu, Zeng, Sabharwal,
  Zettlemoyer, Koh, Farhadi, Smith, and Hajishirzi}]{Olmo3}
Team Olmo, :, Allyson Ettinger, Amanda Bertsch, Bailey Kuehl, David Graham,
  David Heineman, Dirk Groeneveld, Faeze Brahman, Finbarr Timbers, Hamish
  Ivison, Jacob Morrison, Jake Poznanski, Kyle Lo, Luca Soldaini, Matt Jordan,
  Mayee Chen, Michael Noukhovitch, Nathan Lambert, Pete Walsh, Pradeep Dasigi,
  Robert Berry, Saumya Malik, Saurabh Shah, Scott Geng, Shane Arora, Shashank
  Gupta, Taira Anderson, Teng Xiao, Tyler Murray, Tyler Romero, Victoria Graf,
  Akari Asai, Akshita Bhagia, Alexander Wettig, Alisa Liu, Aman Rangapur, Chloe
  Anastasiades, Costa Huang, Dustin Schwenk, Harsh Trivedi, Ian Magnusson,
  Jaron Lochner, Jiacheng Liu, Lester James~V. Miranda, Maarten Sap, Malia
  Morgan, Michael Schmitz, Michal Guerquin, Michael Wilson, Regan Huff,
  Ronan~Le Bras, Rui Xin, Rulin Shao, Sam Skjonsberg, Shannon~Zejiang Shen,
  Shuyue~Stella Li, Tucker Wilde, Valentina Pyatkin, Will Merrill, Yapei Chang,
  Yuling Gu, Zhiyuan Zeng, Ashish Sabharwal, Luke Zettlemoyer, Pang~Wei Koh,
  Ali Farhadi, Noah~A. Smith, and Hannaneh Hajishirzi. 2026.
\newblock \href {https://arxiv.org/abs/2512.13961} {Olmo 3}.
\newblock \emph{Preprint}, arXiv:2512.13961.

\bibitem[{OpenAI(2024)}]{OpenAI2023GPT4}
OpenAI. 2024.
\newblock \href {https://arxiv.org/abs/2303.08774} {{GPT}-4 technical report}.
\newblock \emph{Preprint}, arXiv:2303.08774.

\bibitem[{Park et~al.(2024)Park, Choe, and Veitch}]{pmlr-v235-park24c}
Kiho Park, Yo~Joong Choe, and Victor Veitch. 2024.
\newblock \href {https://proceedings.mlr.press/v235/park24c.html} {The linear
  representation hypothesis and the geometry of large language models}.
\newblock In \emph{Proceedings of the 41st International Conference on Machine
  Learning}, volume 235 of \emph{Proceedings of Machine Learning Research},
  pages 39643--39666. PMLR.

\bibitem[{Radford et~al.(2019)Radford, Wu, Child, Luan, Amodei, and
  Sutskever}]{radford2019language}
Alec Radford, Jeff Wu, Rewon Child, David Luan, Dario Amodei, and Ilya
  Sutskever. 2019.
\newblock Language models are unsupervised multitask learners.

\bibitem[{Raffel et~al.(2020)Raffel, Shazeer, Roberts, Lee, Narang, Matena,
  Zhou, Li, and Liu}]{2020t5}
Colin Raffel, Noam Shazeer, Adam Roberts, Katherine Lee, Sharan Narang, Michael
  Matena, Yanqi Zhou, Wei Li, and Peter~J. Liu. 2020.
\newblock \href {http://jmlr.org/papers/v21/20-074.html} {Exploring the limits
  of transfer learning with a unified text-to-text transformer}.
\newblock \emph{Journal of Machine Learning Research}, 21(140):1--67.

\bibitem[{Ramachandran et~al.(2018)Ramachandran, Zoph, and
  Le}]{Ramachandran2017}
Prajit Ramachandran, Barret Zoph, and Quoc~V. Le. 2018.
\newblock \href {https://openreview.net/forum?id=Hkuq2EkPf} {Searching for
  activation functions}.

\bibitem[{Rushing and Nanda(2024)}]{pmlr-v235-rushing24a}
Cody Rushing and Neel Nanda. 2024.
\newblock \href {https://proceedings.mlr.press/v235/rushing24a.html}
  {Explorations of self-repair in language models}.
\newblock In \emph{Proceedings of the 41st International Conference on Machine
  Learning}, volume 235 of \emph{Proceedings of Machine Learning Research},
  pages 42836--42855. PMLR.

\bibitem[{Sanh et~al.(2019)Sanh, Debut, Chaumond, and
  Wolf}]{Sanh2019DistilBERT}
Victor Sanh, Lysandre Debut, Julien Chaumond, and Thomas Wolf. 2019.
\newblock \href
  {https://www.emc2-ai.org/assets/docs/neurips-19/emc2-neurips19-paper-33.pdf}
  {Distil{BERT}, a distilled version of {BERT}: smaller, faster, cheaper and
  lighter}.
\newblock In \emph{NeurIPS $EMC^2$ Workshop}.

\bibitem[{Saphra and Wiegreffe(2024)}]{saphra-wiegreffe-2024-mechanistic}
Naomi Saphra and Sarah Wiegreffe. 2024.
\newblock \href {https://doi.org/10.18653/v1/2024.blackboxnlp-1.30}
  {Mechanistic?}
\newblock In \emph{Proceedings of the 7th BlackboxNLP Workshop: Analyzing and
  Interpreting Neural Networks for NLP}, pages 480--498, Miami, Florida, US.
  Association for Computational Linguistics.

\bibitem[{Sharkey et~al.(2022)Sharkey, Braun, and Millidge}]{Sharkey2022}
Lee Sharkey, Dan Braun, and Beren Millidge. 2022.
\newblock \href
  {https://www.alignmentforum.org/posts/z6QQJbtpkEAX3Aojj/interim-research-report-taking-features-out-of-superposition}
  {[{I}nterim research report] {T}aking features out of superposition with
  sparse autoencoders}.

\bibitem[{Shazeer(2020)}]{2020_Shazeer}
Noam Shazeer. 2020.
\newblock \href {https://arxiv.org/abs/2002.05202} {{GLU} variants improve
  transformer}.
\newblock \emph{Preprint}, arXiv:2002.05202.

\bibitem[{Soldaini et~al.(2024)Soldaini, Kinney, Bhagia, Schwenk, Atkinson,
  Authur, Bogin, Chandu, Dumas, Elazar, Hofmann, Jha, Kumar, Lucy, Lyu,
  Lambert, Magnusson, Morrison, Muennighoff, Naik, Nam, Peters, Ravichander,
  Richardson, Shen, Strubell, Subramani, Tafjord, Walsh, Zettlemoyer, Smith,
  Hajishirzi, Beltagy, Groeneveld, Dodge, and Lo}]{soldaini-etal-2024-dolma}
Luca Soldaini, Rodney Kinney, Akshita Bhagia, Dustin Schwenk, David Atkinson,
  Russell Authur, Ben Bogin, Khyathi Chandu, Jennifer Dumas, Yanai Elazar,
  Valentin Hofmann, Ananya Jha, Sachin Kumar, Li~Lucy, Xinxi Lyu, Nathan
  Lambert, Ian Magnusson, Jacob Morrison, Niklas Muennighoff, Aakanksha Naik,
  Crystal Nam, Matthew Peters, Abhilasha Ravichander, Kyle Richardson, Zejiang
  Shen, Emma Strubell, Nishant Subramani, Oyvind Tafjord, Evan Walsh, Luke
  Zettlemoyer, Noah Smith, Hannaneh Hajishirzi, Iz~Beltagy, Dirk Groeneveld,
  Jesse Dodge, and Kyle Lo. 2024.
\newblock \href {https://doi.org/10.18653/v1/2024.acl-long.840} {Dolma: an open
  corpus of three trillion tokens for language model pretraining research}.
\newblock In \emph{Proceedings of the 62nd Annual Meeting of the Association
  for Computational Linguistics (Volume 1: Long Papers)}, pages 15725--15788,
  Bangkok, Thailand. Association for Computational Linguistics.

\bibitem[{Stacey et~al.(2026)Stacey, Orgad, Inui, Heinzerling, and
  Moosavi}]{Stacey2026Hiddenfailuresrobustness}
Joe Stacey, Hadas Orgad, Kentaro Inui, Benjamin Heinzerling, and Nafise~Sadat
  Moosavi. 2026.
\newblock \href {https://arxiv.org/abs/2604.11662} {A robust evaluation of
  probe robustness: lessons for reliable {OOD} uncertainty quantification}.
\newblock \emph{Preprint}, arXiv:2604.11662.

\bibitem[{Stolfo et~al.(2024)Stolfo, Wu, Gurnee, Belinkov, Song, Sachan, and
  Nanda}]{2024_Stolfo}
Alessandro Stolfo, Ben Wu, Wes Gurnee, Yonatan Belinkov, Xingyi Song, Mrinmaya
  Sachan, and Neel Nanda. 2024.
\newblock \href
  {https://proceedings.neurips.cc/paper_files/paper/2024/hash/e21955c93dede886af1d0d362c756757-Abstract-Conference.html}
  {Confidence regulation neurons in language models}.
\newblock \emph{Advances in Neural Information Processing Systems}.

\bibitem[{Sun et~al.(2024)Sun, Chen, Kolter, and
  Liu}]{Sun2024Massiveactivationslarge}
Mingjie Sun, Xinlei Chen, J.~Zico Kolter, and Zhuang Liu. 2024.
\newblock \href {https://openreview.net/forum?id=F7aAhfitX6} {Massive
  activations in large language models}.
\newblock \emph{COLM}.

\bibitem[{Timkey and van Schijndel(2021)}]{timkey-van-schijndel-2021-bark}
William Timkey and Marten van Schijndel. 2021.
\newblock \href {https://doi.org/10.18653/v1/2021.emnlp-main.372} {All bark and
  no bite: Rogue dimensions in transformer language models obscure
  representational quality}.
\newblock In \emph{Proceedings of the 2021 Conference on Empirical Methods in
  Natural Language Processing}, pages 4527--4546, Online and Punta Cana,
  Dominican Republic. Association for Computational Linguistics.

\bibitem[{Touvron et~al.(2023{\natexlab{a}})Touvron, Lavril, Izacard, Martinet,
  Lachaux, Lacroix, Rozière, Goyal, Hambro, Azhar, Rodriguez, Joulin, Grave,
  and Lample}]{llama}
Hugo Touvron, Thibaut Lavril, Gautier Izacard, Xavier Martinet, Marie-Anne
  Lachaux, Timothée Lacroix, Baptiste Rozière, Naman Goyal, Eric Hambro,
  Faisal Azhar, Aurelien Rodriguez, Armand Joulin, Edouard Grave, and Guillaume
  Lample. 2023{\natexlab{a}}.
\newblock \href {https://arxiv.org/abs/2302.13971} {Llama: Open and efficient
  foundation language models}.
\newblock \emph{Preprint}, arXiv:2302.13971.

\bibitem[{Touvron et~al.(2023{\natexlab{b}})Touvron, Martin, Stone, Albert,
  Almahairi, Babaei, Bashlykov, Batra, Bhargava, Bhosale, Bikel, Blecher,
  Ferrer, Chen, Cucurull, Esiobu, Fernandes, Fu, Fu, Fuller, Gao, Goswami,
  Goyal, Hartshorn, Hosseini, Hou, Inan, Kardas, Kerkez, Khabsa, Kloumann,
  Korenev, Koura, Lachaux, Lavril, Lee, Liskovich, Lu, Mao, Martinet, Mihaylov,
  Mishra, Molybog, Nie, Poulton, Reizenstein, Rungta, Saladi, Schelten, Silva,
  Smith, Subramanian, Tan, Tang, Taylor, Williams, Kuan, Xu, Yan, Zarov, Zhang,
  Fan, Kambadur, Narang, Rodriguez, Stojnic, Edunov, and
  Scialom}]{Touvron2023Llama2}
Hugo Touvron, Louis Martin, Kevin Stone, Peter Albert, Amjad Almahairi, Yasmine
  Babaei, Nikolay Bashlykov, Soumya Batra, Prajjwal Bhargava, Shruti Bhosale,
  Dan Bikel, Lukas Blecher, Cristian~Canton Ferrer, Moya Chen, Guillem
  Cucurull, David Esiobu, Jude Fernandes, Jeremy Fu, Wenyin Fu, Brian Fuller,
  Cynthia Gao, Vedanuj Goswami, Naman Goyal, Anthony Hartshorn, Saghar
  Hosseini, Rui Hou, Hakan Inan, Marcin Kardas, Viktor Kerkez, Madian Khabsa,
  Isabel Kloumann, Artem Korenev, Punit~Singh Koura, Marie-Anne Lachaux,
  Thibaut Lavril, Jenya Lee, Diana Liskovich, Yinghai Lu, Yuning Mao, Xavier
  Martinet, Todor Mihaylov, Pushkar Mishra, Igor Molybog, Yixin Nie, Andrew
  Poulton, Jeremy Reizenstein, Rashi Rungta, Kalyan Saladi, Alan Schelten, Ruan
  Silva, Eric~Michael Smith, Ranjan Subramanian, Xiaoqing~Ellen Tan, Binh Tang,
  Ross Taylor, Adina Williams, Jian~Xiang Kuan, Puxin Xu, Zheng Yan, Iliyan
  Zarov, Yuchen Zhang, Angela Fan, Melanie Kambadur, Sharan Narang, Aurelien
  Rodriguez, Robert Stojnic, Sergey Edunov, and Thomas Scialom.
  2023{\natexlab{b}}.
\newblock \href {https://arxiv.org/abs/2307.09288} {Llama 2: open foundation
  and fine-tuned chat models}.
\newblock \emph{Preprint}, arXiv:2307.09288.

\bibitem[{Vaswani et~al.(2017)Vaswani, Shazeer, Parmar, Uszkoreit, Jones,
  Gomez, Kaiser, and Polosukhin}]{2017_Vaswani}
Ashish Vaswani, Noam~M. Shazeer, Niki Parmar, Jakob Uszkoreit, Llion Jones,
  Aidan~N. Gomez, Lukasz Kaiser, and Illia Polosukhin. 2017.
\newblock \href
  {https://proceedings.neurips.cc/paper_files/paper/2017/hash/3f5ee243547dee91fbd053c1c4a845aa-Abstract.html}
  {Attention is all you need}.
\newblock In \emph{Advances in Neural Information Processing Systems}.

\bibitem[{Vershynin(2025)}]{Vershynin2025}
Roman Vershynin. 2025.
\newblock \href {https://www.math.uci.edu/~rvershyn/papers/HDP-book/HDP-2.pdf}
  {\emph{High dimensional probability}}, 2nd edition.
\newblock Cambridge University Press.

\bibitem[{Voita et~al.(2024)Voita, Ferrando, and
  Nalmpantis}]{voita-etal-2024-neurons}
Elena Voita, Javier Ferrando, and Christoforos Nalmpantis. 2024.
\newblock \href {https://doi.org/10.18653/v1/2024.findings-acl.75} {Neurons in
  large language models: Dead, n-gram, positional}.
\newblock In \emph{Findings of the Association for Computational Linguistics:
  ACL 2024}, pages 1288--1301, Bangkok, Thailand. Association for Computational
  Linguistics.

\bibitem[{Wang and Komatsuzaki(2021)}]{gpt-j}
Ben Wang and Aran Komatsuzaki. 2021.
\newblock \href {https://huggingface.co/EleutherAI/gpt-j-6b} {{GPT-J-6B: A 6
  Billion Parameter Autoregressive Language Model}}.
\newblock \url{https://github.com/kingoflolz/mesh-transformer-jax}.

\bibitem[{Wendler et~al.(2024)Wendler, Veselovsky, Monea, and
  West}]{wendler-etal-2024-llamas}
Chris Wendler, Veniamin Veselovsky, Giovanni Monea, and Robert West. 2024.
\newblock \href {https://doi.org/10.18653/v1/2024.acl-long.820} {Do llamas work
  in {E}nglish? on the latent language of multilingual transformers}.
\newblock In \emph{Proceedings of the 62nd Annual Meeting of the Association
  for Computational Linguistics (Volume 1: Long Papers)}, pages 15366--15394,
  Bangkok, Thailand. Association for Computational Linguistics.

\bibitem[{Wu et~al.(2025)Wu, Arora, Geiger, Wang, Huang, Jurafsky, Manning, and
  Potts}]{Wu2025}
Zhengxuan Wu, Aryaman Arora, Atticus Geiger, Zheng Wang, Jing Huang, Dan
  Jurafsky, Christopher~D. Manning, and Christopher Potts. 2025.
\newblock \href
  {https://raw.githubusercontent.com/mlresearch/v267/main/assets/wu25a/wu25a.pdf}
  {Ax{B}ench: Steering {LLM}s? {E}ven simple baselines outperform sparse
  autoencoders}.
\newblock In \emph{Proceedings of the 42nd International Conference on Machine
  Learning}.

\bibitem[{Yang et~al.(2025)Yang, Li, Yang, Zhang, Hui, Zheng, Yu, Gao, Huang,
  Lv, Zheng, Liu, Zhou, Huang, Hu, Ge, Wei, Lin, Tang, Yang, Tu, Zhang, Yang,
  Yang, Zhou, Zhou, Lin, Dang, Bao, Yang, Yu, Deng, Li, Xue, Li, Zhang, Wang,
  Zhu, Men, Gao, Liu, Luo, Li, Tang, Yin, Ren, Wang, Zhang, Ren, Fan, Su,
  Zhang, Zhang, Wan, Liu, Wang, Cui, Zhang, Zhou, and Qiu}]{Yang2025Qwen3}
An~Yang, Anfeng Li, Baosong Yang, Beichen Zhang, Binyuan Hui, Bo~Zheng, Bowen
  Yu, Chang Gao, Chengen Huang, Chenxu Lv, Chujie Zheng, Dayiheng Liu, Fan
  Zhou, Fei Huang, Feng Hu, Hao Ge, Haoran Wei, Huan Lin, Jialong Tang, Jian
  Yang, Jianhong Tu, Jianwei Zhang, Jianxin Yang, Jiaxi Yang, Jing Zhou,
  Jingren Zhou, Junyang Lin, Kai Dang, Keqin Bao, Kexin Yang, Le~Yu, Lianghao
  Deng, Mei Li, Mingfeng Xue, Mingze Li, Pei Zhang, Peng Wang, Qin Zhu, Rui
  Men, Ruize Gao, Shixuan Liu, Shuang Luo, Tianhao Li, Tianyi Tang, Wenbiao
  Yin, Xingzhang Ren, Xinyu Wang, Xinyu Zhang, Xuancheng Ren, Yang Fan, Yang
  Su, Yichang Zhang, Yinger Zhang, Yu~Wan, Yuqiong Liu, Zekun Wang, Zeyu Cui,
  Zhenru Zhang, Zhipeng Zhou, and Zihan Qiu. 2025.
\newblock \href {https://arxiv.org/abs/2505.09388} {Qwen3 technical report}.
\newblock \emph{Preprint}, arXiv:2505.09388.

\bibitem[{Yang et~al.(2024)Yang, Yang, Hui, Zheng, Yu, Zhou, Li, Li, Liu,
  Huang, Dong, Wei, Lin, Tang, Wang, Yang, Tu, Zhang, Ma, Yang, Xu, Zhou, Bai,
  He, Lin, Dang, Lu, Chen, Yang, Li, Xue, Ni, Zhang, Wang, Peng, Men, Gao, Lin,
  Wang, Bai, Tan, Zhu, Li, Liu, Ge, Deng, Zhou, Ren, Zhang, Wei, Ren, Liu, Fan,
  Yao, Zhang, Wan, Chu, Liu, Cui, Zhang, Guo, and Fan}]{qwen2}
An~Yang, Baosong Yang, Binyuan Hui, Bo~Zheng, Bowen Yu, Chang Zhou, Chengpeng
  Li, Chengyuan Li, Dayiheng Liu, Fei Huang, Guanting Dong, Haoran Wei, Huan
  Lin, Jialong Tang, Jialin Wang, Jian Yang, Jianhong Tu, Jianwei Zhang,
  Jianxin Ma, Jianxin Yang, Jin Xu, Jingren Zhou, Jinze Bai, Jinzheng He,
  Junyang Lin, Kai Dang, Keming Lu, Keqin Chen, Kexin Yang, Mei Li, Mingfeng
  Xue, Na~Ni, Pei Zhang, Peng Wang, Ru~Peng, Rui Men, Ruize Gao, Runji Lin,
  Shijie Wang, Shuai Bai, Sinan Tan, Tianhang Zhu, Tianhao Li, Tianyu Liu,
  Wenbin Ge, Xiaodong Deng, Xiaohuan Zhou, Xingzhang Ren, Xinyu Zhang, Xipin
  Wei, Xuancheng Ren, Xuejing Liu, Yang Fan, Yang Yao, Yichang Zhang, Yu~Wan,
  Yunfei Chu, Yuqiong Liu, Zeyu Cui, Zhenru Zhang, Zhifang Guo, and Zhihao Fan.
  2024.
\newblock \href {https://arxiv.org/abs/2407.10671} {Qwen2 technical report}.
\newblock \emph{Preprint}, arXiv:2407.10671.

\bibitem[{Zhang et~al.(2022)Zhang, Roller, Goyal, Artetxe, Chen, Chen, Dewan,
  Diab, Li, Lin, Mihaylov, Ott, Shleifer, Shuster, Simig, Koura, Sridhar, Wang,
  and Zettlemoyer}]{zhang2022opt}
Susan Zhang, Stephen Roller, Naman Goyal, Mikel Artetxe, Moya Chen, Shuohui
  Chen, Christopher Dewan, Mona Diab, Xian Li, Xi~Victoria Lin, Todor Mihaylov,
  Myle Ott, Sam Shleifer, Kurt Shuster, Daniel Simig, Punit~Singh Koura, Anjali
  Sridhar, Tianlu Wang, and Luke Zettlemoyer. 2022.
\newblock \href {https://arxiv.org/abs/2205.01068} {{OPT}: open pre-trained
  transformer language models}.
\newblock \emph{Preprint}, arXiv:2205.01068.

\bibitem[{Zhao et~al.(2026)Zhao, Choenni, Saxena, and
  Titov}]{zhao-etal-2026-finding}
Xiutian Zhao, Rochelle Choenni, Rohit Saxena, and Ivan Titov. 2026.
\newblock \href {https://doi.org/10.18653/v1/2026.eacl-long.155} {Finding
  culture-sensitive neurons in vision-language models}.
\newblock In \emph{Proceedings of the 19th Conference of the {E}uropean Chapter
  of the {A}ssociation for {C}omputational {L}inguistics (Volume 1: Long
  Papers)}, pages 3366--3381, Rabat, Morocco. Association for Computational
  Linguistics.

\bibitem[{Zhu et~al.(2025)Zhu, Huang, Huang, Zeng, Mao, Wu, Min, and
  Zhou}]{Zhu2025Hyperconnections}
Defa Zhu, Hongzhi Huang, Zihao Huang, Yutao Zeng, Yunyao Mao, Banggu Wu, Qiyang
  Min, and Xun Zhou. 2025.
\newblock \href {https://openreview.net/forum?id=9FqARW7dwB}
  {Hyper-connections}.
\newblock In \emph{The Thirteenth International Conference on Learning
  Representations}.

\end{thebibliography}

\appendix

\section{Responsible NLP Statements}\label{ap:responsible}

\subsection{Computational complexity}
All our experiments can be run on a single NVIDIA RTX A6000 (48GB).
We use TransformerLens \citep{nanda2022transformerlens}.

The main analysis, computing the weight cosines, needs less than a minute per model.

Other parts were more expensive:
\begin{itemize}
	\item For the ablations (\cref{sec:ablation}), each run on Dolma took approximately 8 hours. This is to be multiplied by 6 neuron classes, times 2 for the respective baselines, times 3 for the different numbers of neurons ablated, plus one clean run, leading to a total of 37 runs, i.e. roughly 300 GPU hours.
	\item For the activation-based analysis in \cref{sec:cs},
	we needed a single run of $\approx 25$ h to store the max/min activating examples for all neurons,
	and then $\approx 45$ s per neuron ($\approx 5$ min) to recompute its activations on the relevant texts and visualize them.
	\item Another expensive part is computing the randomness regions based on mismatched cosines
	(\cref{sec:baseline,ap:baseline}).
	The time complexity is $O(n^2)$ in the number of neurons per layer, since we have to consider every pair of neurons.
	Since however we found that this baseline is hardly different from the more "naive" Gaussian one,
	we suggest that future work could just leave out this step.
	\item Finally, our weight processing (\cref{ap:preprocessing}) makes model loading last about a minute.
	A possible solution in future work would be to save the preprocessed weights.
\end{itemize}

\subsection{LLM use}
We used LLM assistants to help with programming.

\section{Impact Statement}
This paper presents work whose goal is to advance the field
of machine learning interpretability.
We believe our work advances the field in the following ways:
First, it provides guidance to future research on GLU-based neurons.
Second, analyzing the input-output behavior of neurons,
rather than just their input or just their output behavior,
is a crucial missing link for understanding the mechanisms within models.
Third, we find a small class of neurons with disproportionate influence.
This can guide future research towards analyzing these neurons in particular,
since they seem especially important
and understanding them yields a good cost-benefit factor.

Like many researchers in the field,
we believe that discovering the underlying structure of models will have several benefits.
First, ideally, any scientific
field should have a deep understanding of the models it
uses; results that are obtained using blackbox models are
hard to understand, replicate and generalize. Second, once
we understand our models better, we will be better able to
address failure modes. For example, once we understand
how unaligned behavior like bias and hallucinations comes
about, it will be easier to address them, e.g., by changing
the model architecture. Third, interpretability can support
explainability. If we understand how a recommendation or
answer came about, we can better assess its validity.

\section{Models and datasets used}
\label{ap:models}

\subsection{Non-GLU models}
\label{ap:models-non-GLU}
\begin{itemize}
	\item three encoder-decoder models from the T5 family \citep{2020t5}: small, base, and large;
	\item two encoder-only models from the BERT family \citep{devlin-etal-2019-bert}: bert-base-cased and bert-large-cased;
	\item Othello-GPT \citep{Li2023Emergentworldrepresentations}, a decoder-only model trained on a non-language task;
	\item 18 decoder-only language models:
	BLOOM-560m, BLOOM-1b1, BLOOM-1b7, BLOOM-7b1 \citep{BigScience2022BLOOM};
	\href{https://huggingface.co/distilbert/distilgpt2}{DistilGPT-2} \citep{Sanh2019DistilBERT};
	GPT2-small, GPT2-medium, GPT2-large, GPT2-XL \citep{radford2019language};
	GPT-J-6B \cite{gpt-j};
	Pythia-14m, Pythia-1b, Pythia-6.9b, Pythia-12b \citep{Biderman2023Pythia};
	OPT-125m, OPT-1.3b, OPT-6.7b, OPT-13B \citep{zhang2022opt}.
\end{itemize}
These models use the GELU activation function, except T5 and OPT which use ReLU.

\shortpar{T5}
Apache 2.0 license. This encoder-decoder model was pretrained on English text and then finetuned for some English-centric tasks, but also translation from English to French, Romanian, and German.

\shortpar{BERT}
Apache 2.0 license, English.

\shortpar{Othello-GPT}
MIT license. This model was not trained on natural language, but on Othello game transcripts.

\shortpar{BLOOM}
This model was released under a custom license, the BigScience Responsible AI License (RAIL).\footnote{\url{https://huggingface.co/spaces/bigscience/license}}
Training data of BLOOM-1b7 contains 45 natural languages and 12 programming languages in varying proportions, see \citet{BigScience2022BLOOM} for details.

\shortpar{GPT2}
MIT license. The training data is not public.

\shortpar{DistilGPT-2}
Apache-2.0 license. This is a distilled version of GPT2, trained on OpenWebText, a dataset of English text.

\shortpar{GPT-J and Pythia}
Apache 2.0 license. English-only.

\shortpar{OPT}
The training data consists of predominantly English text. The model is released under a \href{https://github.com/facebookresearch/metaseq/blob/main/projects/OPT/MODEL_LICENSE.md}{custom license} that limits use to non-commercial research purposes (along with some other restrictions such as military, nuclear, surveillance, or biometry uses).

\subsection{GLU models}
\label{ap:models-GLU}

List of models:
Gemma-2-2B, Gemma-2-9B \cite{gemma_2024},
Llama-2-7B \cite{Touvron2023Llama2}, -3.1-8B \cite{Llama3.1}, -3.2-1B, -3.2-3B \cite{Llama3.2},
OLMo-1B, \href{https://huggingface.co/allenai/OLMo-7B-0424-hf}{OLMo-7B-0424} \cite{groeneveld-etal-2024-olmo}, 
Mistral-7B \cite{jiang2023mistral7b},
Qwen2.5-0.5B, Qwen2.5-7B \cite{qwen2},
Yi-6B \cite{ai2025yiopenfoundationmodels}.
These models use SwiGLU,
except for Gemma, which uses GEGLU.

\shortpar{Gemma}
To download the model one needs to explicitly accept the
\href{https://ai.google.dev/gemma/terms}{terms of use}.
NLP research is explicitly listed as an intended usage \citep{gemma_2024}.

Gemma 1 \citep{gemma_2024} and Gemma 2 (model card \href{https://ai.google.dev/gemma/docs/core/model_card_2}{here}) are primarily English and code.
Gemma 3 was also trained on multilingual data \citep{Gemma3}.

\shortpar{Llama}
Inference code and weights under an ad hoc
\href{https://github.com/meta-llama/llama/blob/main/LICENSE}{license}.
There is also an
\href{https://github.com/meta-llama/llama/blob/main/USE_POLICY.md}{``Acceptable Use Policy''}.
Our work is well within those terms.

In Llama 1,
languages mostly include English and programming languages,
but also Wikipedia dumps from
``bg, ca, cs, da, de, en, es, fr, hr, hu, it,
nl, pl, pt, ro, ru, sl, sr, sv, uk''
\citep{llama}.
Llama 2 is mostly English; a more precise distribution of languages is described in table 10 of \citet{Touvron2023Llama2}.
In Llama 3.1 and 3.2, supported languages are English, German, French, Italian, Portuguese, Hindi, Spanish, and Thai \citep{Llama3.1,Llama3.2},
but at least Llama 3.2 has been trained on a broader range of languages without officially supporting them \citep{Llama3.2}.
There is no more detailed public information about the training data.

\shortpar{OLMo and Dolma}
Training and inference code, weights (OLMo), and data (Dolma) under Apache 2.0 license.
``The Science of Language Models'' is explicitly mentioned as an intended use case.
Dolma is quality-filtered and designed to contain only English and programming languages
(though we came across some French sentences as well, see \cref{tab:cs2})
\citep{groeneveld-etal-2024-olmo,soldaini-etal-2024-dolma}.

\shortpar{Mistral}
Inference code and weights are released under the Apache 2.0 license,
but accessing them requires accepting the
\href{https://mistral.ai/terms}{terms}.
Languages are not explicitly mentioned in the paper,
but clearly include English and code
\citep{jiang2023mistral7b}.

\shortpar{Qwen}
Inference code and weights under Apache 2.0 license.
Supports ``over 29 languages, including
Chinese, English, French, Spanish, Portuguese, German, Italian, Russian, Japanese, Korean, Vietnamese, Thai, Arabic,
and more''
\citep{qwen2}.

\shortpar{Yi}
Inference code and weights under Apache 2.0 license.
Trained on English and Chinese
\citep{ai2025yiopenfoundationmodels}.

\section{Weight preprocessing}
\label{ap:preprocessing}

\subsection{Original TransformerLens preprocessing}

TransformerLens v2 \cite{nanda2022transformerlens}.
applies preprocessing steps to the weights
to make them more interpretable without changing model behavior.
The steps that affect MLP weights are
"LayerNorm folding" and "Centering writing weights".
For details, see the TransformerLens documentation at \url{https://github.com/TransformerLensOrg/TransformerLens/blob/main/further_comments.md}.
All these processing steps filter out some parts of the model weights that don't influence model behavior.
Thus, analyzing the raw model weights (without processing) would just lead to more noisy results.

\begin{figure*}
	\includegraphics[width=\textwidth]
	{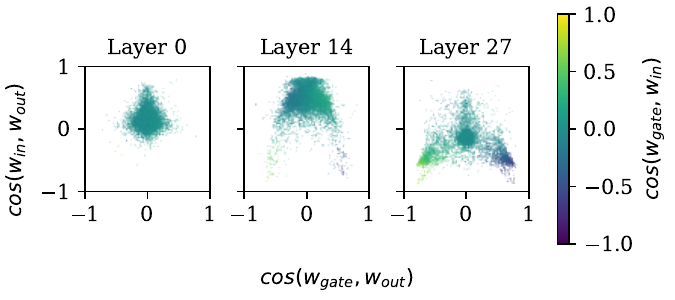}
	\caption{Equivalent of \cref{fig:wcos_selected}, but \textit{without} weight processing.
		(We don't include the randomness regions.)}
	\label{fig:wcos-raw}
\end{figure*}

\subsection{Our additional preprocessing: \wp{}}
\label{ap:preprocessing2}

We propose an additional preprocessing step specific to gated activation functions:
For each neuron,
we multiply $\win$ and $\wout$ by the sign of $\cos(\wgate,\win)$.
We call this step \wp{}.

\wp{} \textbf{does not affect model behavior}:
In \cref{eq:neuron},
if we replace $\win$ by $-\win$ and $\wout$ by $-\wout$, the two minus signs cancel out, so the neuron output remains the same.
We call this the \textbf{symmetry property} of gated activation functions.

We find neurons \textbf{easier to interpret} after applying \wp{}, for the following three reasons:

\shortpar{Reasoning about activation causes}
The two reading weight vectors,
$\wgate$ and $\win$,
now always have a non-negative cosine similarity,
i.e., they do not point in opposite directions.
This makes it easier to reason about what causes a neuron to activate
(there are less minus signs to worry about).
This is especially relevant for the case studies (\cref{sec:cs}).

\shortpar{Treating equivalent neurons the same way}
By the symmetry property, for any neuron we can construct an equivalent one that implements the same function.
Any sound interpretability method should treat these two neurons the same way,
and \wp{} guarantees that this is the case.
For example,
in plots like \cref{fig:wcos_selected},
neurons that belong together are in the same area of the plot thanks to \wp{}.

\shortpar{Easier definition of conditional ablations}
For the conditional ablations introduced in \cref{sec:conditional},
the four cases correspond to real distinctions.
Without \wp, equivalent cases would be more complicated to define:
e.g. "gate+\_post+" would be defined as 
"$\xgate>0$ and $\xpost$ has the same sign as $\cos(\wgate,\win)$".

\section{Details on method}
\label{ap:alldefs}

\Cref{tab:alldefs} shows the complete class definitions.

\begin{table*}\centering
\small
	\caption{
			Decision table defining input-output (IO) classes in GLU models.
			See \cref{sec:taxonomy-glu} for justification. The threshold $\tau$ used in practice was $0.5$.\\
			Note: What may seem an inconsistency in the typical-atypical distinctions is deliberate:
			When $\wout$ is aligned to both $\wgate$ and $\win$ (as in the \textit{strengthening} and \textit{weakening} fields), we expect also $\wgate$ and $\win$ to be aligned with each other; hence our calling it \textit{typical} when $\cos(\wgate,\win)$ is high.
			On the other hand, if $\wout$ is aligned with only one of the other two weight vectors, we expect the other two to be misaligned with each other.
		}
	\label{tab:alldefs}
	\begin{tabular}{p{.17\linewidth}c|cc|cc}
			&$|\cos(\wgate,\wout)|$
			& \multicolumn{2}{c}{$\approx 1$ (or $>\tau$)}
			& \multicolumn{2}{c}{$\approx 0$ (or $<\tau$)}\\
			$\cos(\win,\wout)$ & &&&&\\
			
			\hline

			\rowcolor{gray!25}			
			\multicolumn{2}{l|}{$ \approx +1$ (or $>+\tau$)}
			& \multicolumn{2}{c}{strengthening}
			
			& \multicolumn{2}{c}{conditional strengthening}\\
					&&  {\scriptsize $|\cos(\wgate,\win)| >\tau$} &
					{\scriptsize $|\cos(\wgate,\win)| <\tau$}&
					{\scriptsize $|\cos(\wgate,\win)| <\tau$}
					&  {\scriptsize $|\cos(\wgate,\win)| >\tau$}\\
					&&typical & atypical
					&typical & atypical\\

			\rowcolor{gray!25}
			\multicolumn{2}{l|}{$ \approx -1$ (or $<-\tau$)}
			& \multicolumn{2}{c}{weakening}
			& \multicolumn{2}{c}{conditional weakening}\\
					&& {\scriptsize $|\cos(\wgate,\win)| >\tau$} &
					{\scriptsize $|\cos(\wgate,\win)| <\tau$}&
					{\scriptsize $|\cos(\wgate,\win)| <\tau$}
					&  {\scriptsize $|\cos(\wgate,\win)| >\tau$}\\
					&&typical & atypical
					&typical & atypical\\

			\rowcolor{gray!25}
			\multicolumn{2}{l|}{$ \approx 0$ (or $\in [-\tau,+\tau]$)}
			& \multicolumn{2}{c}{proportional change}
			& \multicolumn{2}{c}{orthogonal output}\\
					&& {\scriptsize $|\cos(\wgate,\win)| <\tau$}
					&  {\scriptsize $|\cos(\wgate,\win)| >\tau$}&&\\
					&&typical & atypical &&
					\\
		\end{tabular}
	
\end{table*}

\section{Random baselines}
\label{ap:baseline}
\begin{figure*}
	\centering
	\includegraphics
	[width=.67\textwidth]
	{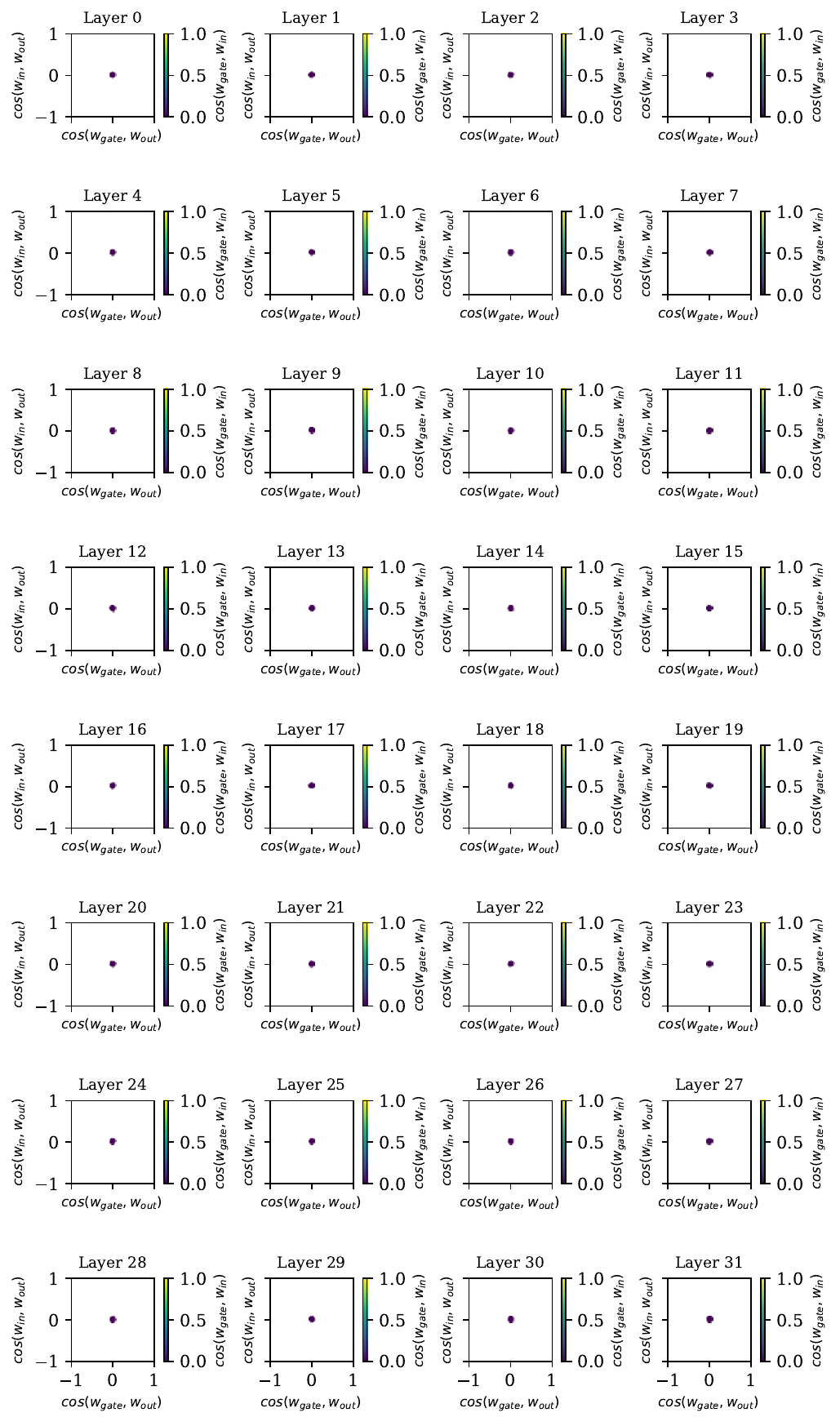}
	\caption{
		Equivalent of
		\cref{fig:wcos_selected}
		for the randomly initialized OLMo-7B model (training checkpoint 0).
		Whatever doesn't look like this, is significant.
		}
	\label{fig:olmo0}
\end{figure*}

Here we describe our two baselines:
random initialization and mismatched cosines.

In a randomly initialized model, all cosine similarities would be very close to zero:
In $n$ dimensions, absolute cosine similarities behave like $1/\sqrt{n}$
(\citealp{Vershynin2025}, p. 68).
More precisely, the cosines follow a beta distribution
with parameters $(d_\text{model}-1)/2, (d_\text{model}-1)/2$,
rescaled to the range $[-1,1]$.\footnote{
	\url{https://stats.stackexchange.com/questions/85916/distribution-of-scalar-products-of-two-random-unit-vectors-in-d-dimensions}
}
Taking, e.g., $d_\text{model}=4096$ (as e.g. in OLMo-7B),
we get a $95\%$ randomness range of approximately $[-0.03,0.03]$.
This is empirically confirmed on
the first training checkpoint of OLMo-7B-0424 (\cref{fig:olmo0}).

Inspired by work on outlier dimensions in the \textit{activations} of Transformers
\cite{ethayarajh-2019-contextual, kovaleva-etal-2021-bert, timkey-van-schijndel-2021-bark, Dettmers2022, Sun2024Massiveactivationslarge},
we suspected that a similar phenomenon might be at work in the \textit{weights},
making cosine similarities artificially high.
To account for this possibility, we construct a second baseline specific to each model layer:
We compute all the (e.g.) $\cos(\win,\wout)$ of a layer,
even if the two weights belong to different neurons.
If a cosine similarity is higher than most of these mismatched cosines,
it is likely not due to an outlier dimension common to all neurons of the layer,
but reflects something specific to this neuron.

\section{Ablation experiments}
\label{ap:ablations}

\subsection{Hypotheses and choice of metrics}

We originally had two hypotheses (which turned out to be wrong, see \cref{sec:ablation}):

\begin{itemize}
	\item
	We hypothesized that \textit{conditional strengthening} neurons might contribute to \textit{subject enrichment} \citep{geva-etal-2023-dissecting},
	a crucial step of factual recall that involves MLPs writing appropriate attributes for the given subject.
	Both phenomena occur in roughly the same layers,
	and similar $\win$ and $\wout$ could correspond to related concepts.
	\item
	We expected that \textit{weakening} neurons would make the output distribution flatter,
	i.e. \textit{increase the entropy}.
	This could happen
	by reducing the probability of high-ranking tokens
	(weakening directions corresponding to tokens)
	or by increasing the probability of very low-ranking tokens
	(weakening directions corresponding to negations of tokens).
\end{itemize}

This is why we tested the two metrics of
\textit{attribute rate} (a proxy of subject enrichment)
and \textit{entropy}.
We additionally considered the \textit{loss},
and, following Gurnee et al.'s  analysis of entropy neurons \citep{2024_Gurnee},
\textit{rank} of the correct token and \textit{scale} of the final hidden state.

\subsection{Details on attribute rate}
\label{ap:attribute}
Our investigation of attribute rate closely follows \citet{geva-etal-2023-dissecting}.
It requires a dataset of subject-attribute mappings
that we didn't have access to.
In order to replicate this dataset, we closely followed the procedure described in their paper,
which assumes attributes are tokens that appear in the same Wikipedia paragraph as the subject (excluding stopwords).
We used the Wikipedia dump from October 20, 2021, instead of October 13,
since there is an official dump made at this date.\footnote{A list of dumps by date is available at
	\href{ https://archive.org/search?query=subject\%3A\%22enwiki\%22+AND+subject\%3A\%22data+dumps\%22+AND+collection\%3A\%22wikimediadownloads\%22&sort=-date
}{this URL}.}
To improve replicability,
we publish our complete code as well as our subject-attribute dataset.

\paragraph{Results}

\begin{figure}
	\centering
	\includegraphics
	[width=.7\linewidth]
	{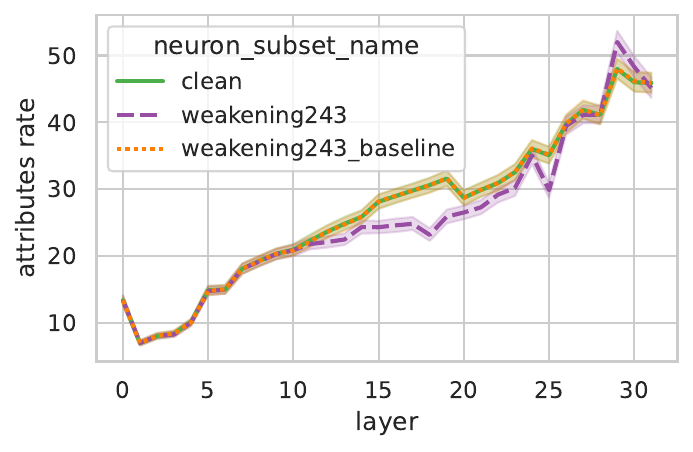}
	\caption{Effect on attribute rate of mean-ablating 243 weakening neurons (weakening243),
		or 243 random neurons from the same layers (weakening243\_baseline).
		The
		baseline has no sizable influence.
		In contrast, there is a small but clearly visible effect when ablating weakening neurons, already from layer $\approx 10$ onward,
		even though weakening neurons are few and mostly in late layers.
		In the \href{https://github.com/sjgerstner/RW_functionalities_results}{supplementary material}
		we show results for other neuron classes, all of which are indistinguishable from the "clean" line.
	}
	\label{fig:attributes}
\end{figure}

Ablating weakening neurons has a small but clearly visible effect in layers $\approx 10$ and onward
(see \cref{fig:attributes}),
whereas the same number of random neurons from the same layers (the orange dotted line in the same figure)
or from any other given class
has no visible effect at all (see \href{https://github.com/sjgerstner/RW_functionalities_results}{supplementary material}).
This is particularly interesting since there are very few weakening neurons in these early-middle layers.
The direction of the effect is different across layers,
so we cannot conclude that
"weakening neurons are responsible for attribute rate";
instead, weakening neurons are crucial components in general,
and their role is not limited to any specific stage of inference.

\subsection{Mean ablation}
\label{ap:mean-ablation}
\subsubsection{Method}
\shortpar{Computing the means}
We pre-computed the mean activation of every neuron on the same 20M token subset of Dolma that we also used for the actual ablation experiment.

\shortpar{Conditional ablations}
For conditional ablations, we replace the neuron activation by the mean value it would have in the corresponding case (not the mean activation of the neuron overall).
For example, in the case "gate-\_post+":
\begin{itemize}
	\item we replace the activation ($\xpost$) only when the condition "gate-\_post+" is fulfilled, i.e. when $\xgate<0$ and $\xpost>0$ -- this is just the definition of conditional ablation;
	\item the value that we replace it with is the mean value of $\xpost$ \emph{across the cases in which $\xgate<0$ and $\xpost>0$}.
\end{itemize}

\subsubsection{Comparison of mean and zero ablation results}
Mean ablation recovers effects of weakening neurons that would go unnoticed when just using zero ablation.
We hypothesize that these additional effects happen when activations are relatively close to zero but far away from their mean.
This remains to be tested in future work.

\section{Double checking}
\label{sec:doublecheck}
In our case studies,
we observe that
many neurons
have the property of \textbf{double checking}:
The
two reading weight vectors ($\wgate$ and $\win$) are
approximately orthogonal, but still intuitively represent
the same concept.

We characterize double checking as follows:
The sets of meaningful vectors \textit{most similar} to $\wgate$ and $\win$
have a high overlap.
More formally, let $U = \{u_0, ..., u_{d_{\text{vocab}}}\}$ be the set of unembedding vectors;
then
\[\argmax_{u\in U} \cos(u,\wgate) \approx \argmax_{u\in U} \cos(u,\win).\]

This phenomenon is \textit{possible} because
random vectors in high dimensions are ``lone stars'' (\citealp{Vershynin2025}, p. 68).
If this is the case for the unembedding vectors,
it is plausible that we can find $\wgate,\win$
that are reasonably similar to a $u_i$ but not to any other $u_j$.
These $\wgate,\win$ can even be (approximately) orthogonal to each other,
as in the following three-dimensional toy example:
$u_1=(1,0,0), u_2=(0,1,0), \wgate=(1,0,1), \win=(1,0,-1)$.

However the phenomenon is \textit{unlikely} to occur in random vectors,
and hence is a significant finding:
If choosing $\wgate,\win$ randomly,
we would expect them to be approximately orthogonal to \textit{all} unembedding vectors;
and even if both were somewhat similar to an unembedding,
we certainly wouldn't expect it to be the same unembedding for both.

We would also not naively expect this phenomenon in a trained network:
If the role of both $\wgate$ and $\win$ is to detect a concept
(e.g. a token prediction)
represented by a vector $u$,
then we would get the best performance with $\wgate=\win=u$,
i.e., $\wgate,\win$ would not be orthogonal.

Double checking is therefore likely to be a useful feature for the model.
We hypothesize that this is because
it shrinks the region in model space that activates the neuron positively.
If (say) $\win=\wgate= (1,0)$, the neuron activates whenever the (normalized) residual input $x$ satisfies $x \cdot (1,0) > 0$;
this happens on the whole half-space $x_1>0$.
If however $\wgate = (1,0)$ and $\win = (0,1)$, the neuron activates positively only in the first quadrant ($x_1,x_2>0$).

This behavior thus enables more precise concept detection.
This may explain
why conditional neurons are more frequent than their unconditional counterparts.

\begin{table*}\centering
	\caption{Overview of prediction/suppression neurons chosen for case studies in \cref{ap:cs}
	}
	\label{tab:cs}
	\rowcolors{2}{white}{gray!25}
	\begin{tabular}{l|l|rrr}
		Neuron & RW category & $\cos(\wgate,\win)$ & $\cos(\wgate,\wout)$ & $\cos(\win,\wout)$ \\
		\href{https://gluscope.github.io/OLMo-7B-0424/L28/N4737/vis.html}{28.4737} & strengthening  & 0.5290 & 0.5048 & 0.7060\\
		\href{https://gluscope.github.io/OLMo-7B-0424/L28/N9766/vis.html}{28.9766} & conditional strengthening  & 0.4764 & 0.4119 & 0.5982\\ 
		\href{https://gluscope.github.io/OLMo-7B-0424/L31/N9634/vis.html}{31.9634} & weakening  & 0.7164 & -0.7218 & -0.8542\\
		\href{https://gluscope.github.io/OLMo-7B-0424/L29/N10900/vis.html}{29.10900} & conditional weakening & 0.4988 & -0.4992 & -0.5775\\
		\href{https://gluscope.github.io/OLMo-7B-0424/L30/N10972/vis.html}{30.10972} & proportional change & 0.4543 & -0.5814 & -0.4182\\
		\href{https://gluscope.github.io/OLMo-7B-0424/L29/N4180/vis.html}{29.4180} & orthogonal output & 0.0272 & 0.4057 & 0.0669\\
	\end{tabular}
\end{table*}

\begin{table*}\centering
	\caption{
		Description of the weight vectors of the selected
		\textit{prediction}
		neurons, by top tokens or similarity to $\wout$.
		The question mark, ?, signals unknown unicode characters.
		The last column presents the (shortened) text samples on which the respective neuron activates most strongly (positively or negatively).
	}
	\label{tab:cs2}
	\rowcolors{2}{white}{gray!25}
	\begin{tabular}
		{p{.12\textwidth}|p{.08\textwidth}p{.08\textwidth}|p{.08\textwidth}p{.08\textwidth}|p{.08\textwidth}
		}

		Neuron,\newline RW class & \multicolumn{2}{c|}{$\wgate$} & \multicolumn{2}{c|}{$\win$} & $\wout$ %& Top activations
		\\
		
		\href{https://gluscope.github.io/OLMo-7B-0424/L28/N4737/vis.html}{28.4737}\newline strengthening &
		\multicolumn{2}{c|}{$\approx \wout$} &
		\multicolumn{2}{c|}{$\approx \wout$} &
		pos:\newline \textit{\blank review\newline \blank Review}
		\\
		
		\href{https://gluscope.github.io/OLMo-7B-0424/L28/N9766/vis.html}{28.9766}\newline conditional strengthening &
		pos:\newline \textit{well\newline \blank well
		} & neg:\newline \textit{\blank far\newline \blank high
		} &
		\multicolumn{2}{c|}{$\approx \wout$} &
		pos:\newline \textit{\blank well\newline well
		}
		\\
		
		\href{https://gluscope.github.io/OLMo-7B-0424/L31/N9634/vis.html}{31.9634}\newline weakening &
		\multicolumn{2}{c|}{$\approx-\wout$} &
		\multicolumn{2}{c|}{$\approx-\wout$} &
		pos:\newline \textit{\blank again\newline \blank Again
		}
		\\
		
		\href{https://gluscope.github.io/OLMo-7B-0424/L29/N10900/vis.html}{29.10900}\newline conditional weakening &
		pos:\newline \textit{\blank today\newline \blank nowadays
		}& neg:\newline \textit{\blank these\newline these
		} &
		\multicolumn{2}{c|}{$\approx-\wout$} &
		pos:\newline \textit{\blank these\newline \blank These
		}
		\\
		
		\href{https://gluscope.github.io/OLMo-7B-0424/L30/N10972/vis.html}{30.10972}\newline proportional change &
		\multicolumn{2}{c|}{$\approx \wout$} &
		pos:\newline \textit{\blank when\newline when
		}& neg:\newline \textit{\blank timing\newline \blank dates
		} &
		neg:\newline \textit{\blank when\newline when
		}
		\\
		
		\href{https://gluscope.github.io/OLMo-7B-0424/L29/N4180/vis.html}{29.4180}\newline  orthogonal output&
		pos:\newline \textit{\blank here\newline \blank therein
		}& neg:\newline \textit{\blank there\newline \blank we
		} &
		pos: ? & neg:\newline \textit{\blank here\newline \blank in
		}& 
		neg:\newline \textit{\blank there\newline there
		}
		\\
		
	\end{tabular}
	
\end{table*}

\begin{table*}\centering
	\caption{
		Description of the weight vectors of the selected
		\textit{prototypical}
		neurons, by top tokens or similarity to $\wout$.
		The question mark, ?, signals unknown unicode characters.
		The last column presents the (shortened) text samples on which the respective neuron activates most strongly (positively or negatively).
	}
	\label{tab:cs-prototypical}
	\rowcolors{2}{white}{gray!25}
	\begin{tabular}{p{.12\textwidth}|p{.11\textwidth}p{.11\textwidth}|p{.11\textwidth}p{.11\textwidth}|p{.11\textwidth}p{.11\textwidth}
		}
		
		Neuron,\newline RW class & \multicolumn{2}{c|}{$\wgate$} & \multicolumn{2}{c|}{$\win$} & \multicolumn{2}{c}{$\wout$} %& Top activations
		\\
		
		\href{https://gluscope.github.io/OLMo-7B-0424/L25/N9997/vis.html}{25.9997} \newline strengthening &
		\multicolumn{2}{c|}{$\approx \wout$} &
		\multicolumn{2}{c|}{$\approx \wout$} &
		pos:\newline \textit{\blank S\newline S} &
		neg:\newline \textit{\blank Chocolate\newline \blank Cour}
		\\
		
		\href{https://gluscope.github.io/OLMo-7B-0424/L5/N10602/vis.html}{5.10602}\newline conditional strengthening &
		pos:\newline \textit{t\newline as} & neg:\newline \textit{deep\newline hum} &
		\multicolumn{2}{c|}{$\approx \wout$} &
		pos:\newline \textit{as\newline t} & neg:\newline \textit{ating\newline \blank their}
		\\
		
		\href{https://gluscope.github.io/OLMo-7B-0424/L31/N7117/vis.html}{31.7117}\newline weakening &
		\multicolumn{2}{c|}{$\approx-\wout$} &
		\multicolumn{2}{c|}{$\approx-\wout$} &
		pos:\newline \textit{\blank by\newline by} & neg:\newline \textit{ani\newline iw}
		\\
		
		\href{https://gluscope.github.io/OLMo-7B-0424/L23/N6543/vis.html}{23.6543}\newline conditional weakening &
		pos:\newline \textit{the\newline a}& neg:\newline \textit{ham\newline aden} &
		\multicolumn{2}{c|}{$\approx-\wout$} &
		pos:\newline \textit{\blank Op\newline \blank AB} & neg:\newline \textit{\blank rom\newline \blank c}
		\\
		
		\href{https://gluscope.github.io/OLMo-7B-0424/L25/N7415/vis.html}{25.7415}\newline proportional change &
		\multicolumn{2}{c|}{$\approx \wout$} &
		pos:\newline \textit{berry\newline rod
		}& neg:\newline \textit{a\newline the
		} &
		pos:\newline \textit{?\newline \blank Hart} & neg:\newline \textit{\blank Nine\newline jin}
		\\
		
	\end{tabular}
	
\end{table*}

\section{Case studies}
\label{ap:cs}

\subsection{Neuron choice}
We used two different methods to find interesting neurons:

\textbf{First}, we selected among \textit{prediction neurons} in the sense of \citet{2024_Gurnee}.
These are defined as neurons whose $\cos(W_U,\wout)$ has a high kurtosis;
in other words, they boost predictions of a small set of tokens while leaving other token scores virtually unchanged.
Specifically, from each discrete RW class we chose the neuron with the highest kurtosis.
This first method guarantees finding interpretable neurons in terms of output behavior,
though not necessarily an interpretable \textit{overall} behavior.
See \cref{tab:cs}
for an overview of neurons chosen by this method.

A downside is that prediction neurons tend to appear in later layers only.
Therefore this neuron choice does not help understand what happens in early layers,
especially why there are so many conditional strengthening neurons.
We therefore also use a \textbf{second} method:
We just select the most prototypical neuron from each class.
For example, for conditional strengthening, we take the neuron with the highest $\cos(\win,\wout)$ among those neurons whose $\cos(\wgate,\wout)$ is within the randomness range (\cref{sec:baseline,ap:baseline}).
This method led to choosing the neurons
\href{https://gluscope.github.io/OLMo-7B-0424/L5/N10602/vis.html}{5.10602} (conditional strengthening),
\href{https://gluscope.github.io/OLMo-7B-0424/L23/N6543/vis.html}{23.6543} (conditional weakening),
\href{https://gluscope.github.io/OLMo-7B-0424/L25/N7415/vis.html}{25.7415} (proportional change),
\href{https://gluscope.github.io/OLMo-7B-0424/L25/N9997/vis.html}{25.9997} (strengthening),
\href{https://gluscope.github.io/OLMo-7B-0424/L31/N7117/vis.html}{31.7117} (weakening).

\subsection{Methods}
Additionally to our RW analysis, we use two well-established neuron analysis methods:

First, we project neuron weights to vocabulary space
with the unembedding matrix $W_U$ and
inspect high-scoring tokens.

Second, we find text examples on which the neurons are strongly activated (positively or negatively).
For each neuron we save the 16 strongest positive and negative activations, respectively.

\subsection{Detailed analysis of weakening neuron 31.9634}
\label{ap:cs-weakening}
Here we say a bit more about the neuron analyzed in \cref{sec:cs}.

Judging by the weights,
we would predict the following:
The neuron activates
positively when
the residual stream contains the ``minus \textit{again}'' direction,
and then weakens that direction
by writing "plus \textit{again}".
The neuron activates
negatively when 
the residual stream contains
information both for and against predicting \textit{again},
and then weakens the \textit{again} direction.
Given that $\wgate$ and $\win$ are highly similar
($\cos(\wgate,\win)=0.7164$),
we would expect that it is easier for the neuron to activate positively (with $\xgate$ and $\xin$ of the same sign).

When actually recording activations of the neuron,
we get a more complex picture:
First of all, the neuron often activates negatively.
Strong negative activations are often on punctuation,
and the actual next token is often \textit{meanwhile}
or \textit{instead}
(and not \textit{again}).
On the positive side,
the strongest activations do not have any obvious semantic relationship to \textit{again}.
We also observed
weaker positive activations
when \textit{again} is a plausible continuation,
e.g., on the token \textit{\textbf{once}} (as in \textit{once again}).
These are cases with negative $\xgate$ values (and also $\xin<0$, hence positive activations) --
a case that we found to be important in \cref{sec:conditional}.
In
these cases, \textit{again} is already weakly present in the
residual stream before the last MLP,
and the neuron
reinforces \textit{again}.

Thus the behavior of this particular weakening neuron is interpretable in the $\xgate<0$ case,
echoing our finding from \cref{sec:conditional} that this case is surprisingly relevant to model behavior.
The $\xgate>0$ case is less interpretable for this particular neuron,
even though this case is more frequent and can lead to stronger activations.
Nevertheless we have some hypotheses for the strong activations as well:
For strong positive activations (which showed no clear pattern),
we hypothesize that sometimes the residual stream ends up near ``minus \textit{again}''
for semantically unrelated reasons
(there are many more possible concepts than dimensions,
so the corresponding directions cannot be fully orthogonal;
see \citealp{Elhage2022Toymodelssuperposition});
in these cases the neuron would reduce the unjustified presence of this ``minus \textit{again}'' direction.
With strong negative activations (where the next token was often \textit{meanwhile} or \textit{instead}),
the neuron may ensure only these tokens are predicted,
and not the relatively similar \textit{again}.

\subsection{Results and analysis for prediction neurons}

See \cref{tab:cs2}.

\textbf{Strengthening neuron \href{https://gluscope.github.io/OLMo-7B-0424/L28/N4737/vis.html}{28.4737}}
predicts \textit{review} (and related tokens) if
activated positively, which happens if \textit{review} is already
present in the residual stream. The maximally positive
activations are in standard contexts that continue with \textit{review} or similar,
such as
the newline after the description of an e-book
(the next paragraph often is the beginning of a review).
On the other hand,
strong negative activations (with $\xgate>0$, $\xin<0$)
often occur in contexts where the next token is or could be something like \textit{blog} or \textit{post}, self-referencing the text.
Other negative activations, with $\xgate<0$ and $\xin>0$,
occur more broadly in contexts semantically talking about reviews
(not just on the exact token before \textit{review}).

\textbf{Conditional strengthening neuron \href{https://gluscope.github.io/OLMo-7B-0424/L28/N9766/vis.html}{28.9766}'s}
RW functionality concerns
\textit{well} and similar tokens.
28.9766
promotes them if activated positively, which happens
when both $\wgate$ and $\win$ indicate that \emph{well} is
represented in the residual stream.
This is a case of double checking.
The maximally positive activation in our sample occurs on
\textit{\textbf{Oh}},
in a context in which \textit{Oh, well} makes sense
(and is the actual continuation).

\textbf{Weakening neuron \href{https://gluscope.github.io/OLMo-7B-0424/L31/N9634/vis.html}{31.9634}.}
See \cref{ap:cs-weakening}.

\textbf{Conditional weakening neuron \href{https://gluscope.github.io/OLMo-7B-0424/L29/N10900/vis.html}{29.10900}.}
Gate and
linear input weight vectors act as two independent
ways of checking for the absence of the token \textit{these} in the
residual stream.
This is a case of double checking (see \cref{sec:doublecheck}).
At the same time, the gate and in weights check for predictions like \textit{today, nowadays}.
When such predictions are present,
the neuron
promotes \textit{these}.
This is a plausible choice in these cases because of the expression \textit{these days}.
An example is
\textit{social media tools change and come and go at the drop of a \textbf{hat}}.
(This sentence talks about a characteristic of current times,
so \textit{these days} would indeed be a plausible continuation.)

\textbf{Proportional change neuron \href{https://gluscope.github.io/OLMo-7B-0424/L30/N10972/vis.html}{30.10972}}
predicts
the token \textit{when} if activated negatively. This happens if
\textit{when} is absent from the residual stream (gate
condition) and is proportional to the presence of time-related
tokens (-$\win$).
An example for a large negative activation is
\textit{puts you on multiple webpages \textbf{at}}.\footnote{
	The actual sentence ends with \textit{as soon as} and comes from a now-dead webpage.
	%http://raf-ranking.com/building-backlinks-how-to-attract-massive-visitors-to-your-blog-for-free/
	We also found one occurrence of \textit{at when} in what seems to be a paraphrase of the same text,
	on https://www.docdroid.net/RgxdG5s/fantastic-tips-for-bloggers-of-all-amountsoystcpdf-pdf .
	We suspect that both texts are machine-generated paraphrases of an original text containing \textit{at once}
	(\textit{when} and \textit{as soon as} can be synonyms of \textit{once} in other contexts),
	and that the model has (also) seen a paraphrased version with \textit{at when}.
	In fact many of the largest negative activations are on \textit{at} in contexts calling for \textit{at once}.
}
Conversely, if \textit{when} is absent, and time-related tokens are absent too,
the neuron activates positively and suppresses \textit{when} further.

\textbf{Orthogonal output neuron
	\href{https://gluscope.github.io/OLMo-7B-0424/L29/N4180/vis.html}{29.4180}
}
predicts \textit{there} (positive activation) if
the residual stream contains a component that we
interpret as ``complement of place expected''
(e.g., \textit{here}, \textit{therein}).
Both
$\wgate$ and $\win$ check for (different aspects of)
this component being present, another case of
double checking.
The largest positive activation is on
\textit{here \textbf{or}}.

Overall, these neurons all promote a specific set of tokens
(we chose them that way),
but under very different circumstances.
The (conditional) strengthening neurons
are the most straightforward to interpret,
because their input and output clearly correspond to the same concept.
In contrast, weakening neurons inherently involve
(an apparent) conflict
between the intermediate model prediction and what the neuron promotes.

\subsection{Results and analysis for prototypical examples}
See \cref{tab:cs-prototypical}.

\textbf{Strengthening neuron \href{https://gluscope.github.io/OLMo-7B-0424/L25/N9997/vis.html}{25.9997}}
is all about strengthening an \textit{S} as a next token.
The strong positive activations ($\xgate,\xin>0$)
usually have an \textit{S} as next token,
but in very specific contexts such as abbreviations or names of fictional characters (\textit{Jan\textbf{os} Slynt}).
This may be due to tokenization (in more common contexts the \textit{s} will not be a standalone token),
or these activations might have the specific role of strengthening the \textit{s} prediction in memorized contexts.
Note that if this were the case, the neuron would play a role in the model's memory of these contexts, but would not be responsible for it alone.
The other activations (either $\xgate$ or $\xin$ negative)
are not readily interpretable,
but also much smaller.

\textbf{Conditional strengthening neuron \href{https://gluscope.github.io/OLMo-7B-0424/L5/N10602/vis.html}{5.10602}}
activates ($\xgate>0$) on those tokens that often start negated auxiliary verbs:
\textit{don, aren, won, didn}.
Correspondingly, the top token of $W_U \wgate$ is \textit{t}
(but interestingly not an apostrophe).
On the other hand, $\win$ detects alternative predictions:
\textit{ate, ating} etc. (as in \textit{donate}) lead to a negative activation,
and \textit{as} (as in \textit{arenas}) leads to a positive activation.
Correspondingly the strongest positive activations are on the \textit{aren} of \textit{arenas}
(but the strongest negative activations are not always in a \textit{donate} context,
perhaps because both \textit{don't} and \textit{donate} can appear in the same slots).
These alternative predictions
(\textit{aren}->\textit{as} or \textit{don}->\textit{ating}, respectively)
are then strengthened by $\wout$.

\textbf{Weakening neuron \href{https://gluscope.github.io/OLMo-7B-0424/L31/N7117/vis.html}{31.7117}}
is in many ways similar to the other weakening neuron we investigated (see \cref{ap:cs-weakening}).
Based on the weights, this neuron detects the intermediate prediction "minus \textit{by}" and writes \textit{by}
(just like the other neuron detects "minus \textit{again}" and writes \textit{again}).
When examining the activations, we also get a more complex picture, that is similar to the other neuron:
Negative activations ($\xgate>0, \xin<0$) are surprisingly frequent;
strong activations (positive or negative) are not particularly interpretable;
but negative-gate activations ($\xgate<0, \xin<0$, corresponding to the neuron strengthening a \textit{by} prediction) are more interpretable in that \textit{by} is often (though not always) the next token.
There are however also some differences:
In the case $\xgate,\xin>0$,
sometimes the preceding, current, or next token is a \textit{by}.
Perhaps, in these cases a previous model component indicated that \textit{by} should not be repeated,
and the neuron weakens this signal.
A similar observation can be made about the weaker negative activations with $\xgate<0,\xin>0$:
the token \textit{by} is around, but usually not the correct prediction.
Here the residual stream contains a contradictory signal, leading to a negative activation of the neuron, which then writes "minus \textit{by}".

\textbf{Conditional weakening neuron \href{https://gluscope.github.io/OLMo-7B-0424/L23/N6543/vis.html}{23.6543}:}
About the only interpretable thing is that $\xgate<0, \xin>0$ (weak negative) activations tend to occur on the penultimate token of personal names, in contexts like "X said/commented/...".
Based on the weights, it is not particularly interpretable what effect the neuron has in these or any other contexts.

\textbf{Proportional change neuron \href{https://gluscope.github.io/OLMo-7B-0424/L25/N7415/vis.html}{25.7415}:}
The weights are not particularly interpretable on their own.
The activations seem to be polysemantic:
several distinct patterns emerge.
Among the top activations with $\xin>0$ (whether or not $\xgate$ is positive or negative),
many are on a token (parenthesis or slash) announcing a metric conversion,
e.g. the parenthesis in \textit{180 °C (350 °F)}.
Others are on the last token of multi-token proper nouns.
So $\win$ corresponds to these concepts.
As for $\wgate$,
the top positive $\xgate$ values mostly happen on punctuation marks starting a line (often the comment signs \verb|\\| or \verb|#| in code).
The most negative $\xgate$ values tend to happen on the penultimate token of some arbitrary-looking token strings (such as proper nouns, typoed words, or chemical compounds).
So possibly $\wgate$ could be interpreted as "something new should start vs. the current thing should be ended".
The neuron modulates this "start of something" concept
proportionally to the presence of $\win$:
When a metric conversion is expected or a proper noun has just ended,
the neuron strengthens this prediction that something should start.

\subsection{More case studies}
These are various neurons that popped out to us as possibly interesting,
for not very systematic reasons,
for example because they strongly activated on a specific named entity.
All of them are in OLMo-7B.

We encourage the readers to explore the activations of these neurons on their own (links below) and compare this with simple weight-based analyses (e.g. logit lens on weight vectors).

Conditional strengthening neurons:
\href{https://gluscope.github.io/OLMo-7B-0424/L0/N1480/vis.html}{0.1480},
\href{https://gluscope.github.io/OLMo-7B-0424/L4/N1940/vis.html}{4.1940},
\href{https://gluscope.github.io/OLMo-7B-0424/L4/N3720/vis.html}{4.3720},
\href{https://gluscope.github.io/OLMo-7B-0424/L4/N4801/vis.html}{4.4801},
\href{https://gluscope.github.io/OLMo-7B-0424/L4/N5772/vis.html}{4.5772},
\href{https://gluscope.github.io/OLMo-7B-0424/L4/N6517/vis.html}{4.6517},
\href{https://gluscope.github.io/OLMo-7B-0424/L4/N6799/vis.html}{4.6799},
\href{https://gluscope.github.io/OLMo-7B-0424/L4/N7667/vis.html}{4.7667},
\href{https://gluscope.github.io/OLMo-7B-0424/L4/N9983/vis.html}{4.9983},
\href{https://gluscope.github.io/OLMo-7B-0424/L4/N10859/vis.html}{4.10859},
\href{https://gluscope.github.io/OLMo-7B-0424/L4/N10882/vis.html}{4.10882},
\href{https://gluscope.github.io/OLMo-7B-0424/L4/N10995/vis.html}{4.10995},
\href{https://gluscope.github.io/OLMo-7B-0424/L22/N2589/vis.html}{22.2589},
\href{https://gluscope.github.io/OLMo-7B-0424/L24/N4880/vis.html}{24.4880},
\href{https://gluscope.github.io/OLMo-7B-0424/L24/N6771/vis.html}{24.6771},
\href{https://gluscope.github.io/OLMo-7B-0424/L25/N2723/vis.html}{25.2723},
\href{https://gluscope.github.io/OLMo-7B-0424/L25/N10496/vis.html}{25.10496};

Weakening neurons:
\href{https://gluscope.github.io/OLMo-7B-0424/L30/N9996/vis.html}{30.9996},
\href{https://gluscope.github.io/OLMo-7B-0424/L31/N9216/vis.html}{31.9216};

Conditional weakening neurons:
\href{https://gluscope.github.io/OLMo-7B-0424/L24/N10431/vis.html}{24.10431};

Proportional change neurons:
\href{https://gluscope.github.io/OLMo-7B-0424/L25/N7032/vis.html}{25.7032},
\href{https://gluscope.github.io/OLMo-7B-0424/L25/N8607/vis.html}{25.8607},
\href{https://gluscope.github.io/OLMo-7B-0424/L29/N8118/vis.html}{29.8118},
\href{https://gluscope.github.io/OLMo-7B-0424/L31/N5490/vis.html}{31.5490},
\href{https://gluscope.github.io/OLMo-7B-0424/L31/N6275/vis.html}{31.6275},
\href{https://gluscope.github.io/OLMo-7B-0424/L31/N8342/vis.html}{31.8342};

Orthogonal output neurons:
\href{https://gluscope.github.io/OLMo-7B-0424/L0/N1758/vis.html}{0.1758},
\href{https://gluscope.github.io/OLMo-7B-0424/L0/N3338/vis.html}{0.3338},
\href{https://gluscope.github.io/OLMo-7B-0424/L0/N3872/vis.html}{0.3872},
\href{https://gluscope.github.io/OLMo-7B-0424/L0/N7829/vis.html}{0.7829},
\href{https://gluscope.github.io/OLMo-7B-0424/L0/N7966/vis.html}{0.7966},
\href{https://gluscope.github.io/OLMo-7B-0424/L29/N2568/vis.html}{29.2568},
\href{https://gluscope.github.io/OLMo-7B-0424/L29/N3327/vis.html}{29.3327},
\href{https://gluscope.github.io/OLMo-7B-0424/L29/N4101/vis.html}{29.4101},
\href{https://gluscope.github.io/OLMo-7B-0424/L29/N6417/vis.html}{29.6417},
\href{https://gluscope.github.io/OLMo-7B-0424/L29/N9734/vis.html}{29.9734},
\href{https://gluscope.github.io/OLMo-7B-0424/L30/N2667/vis.html}{30.2667},
\href{https://gluscope.github.io/OLMo-7B-0424/L30/N3143/vis.html}{30.3143},
\href{https://gluscope.github.io/OLMo-7B-0424/L30/N3883/vis.html}{30.3883},
\href{https://gluscope.github.io/OLMo-7B-0424/L30/N4577/vis.html}{30.4577},
\href{https://gluscope.github.io/OLMo-7B-0424/L30/N5372/vis.html}{30.5372},
\href{https://gluscope.github.io/OLMo-7B-0424/L30/N8535/vis.html}{30.8535},
\href{https://gluscope.github.io/OLMo-7B-0424/L31/N2135/vis.html}{31.2135},
\href{https://gluscope.github.io/OLMo-7B-0424/L31/N10424/vis.html}{31.10424}.

\section{Additional figures and tables}\label{ap:hr}
These final figures and tables show additional results:
\begin{itemize}
	\item \Cref{ap:hr-freq}
	shows more results on activation frequencies.
	\item \Cref{ap:hr-ablations} shows additional results on neuron ablations.
	\item In \cref{ap:hr-distributions}, we show our analyses of IO functionalities by layer (\cref{sec:stat}) for all the models we investigated, both non-GLU and GLU.
\end{itemize}

To keep this appendix at a manageable size, we don't include all the plots produced by our experiments. We publish the other plots as supplementary material at \url{https://github.com/sjgerstner/RW_functionalities_results}.

\subsection{Activation frequencies}
\label{ap:hr-freq}
\label{ap:freq}

\begin{figure*}
	\centering
	\includegraphics
	[height=.95\textheight]
	{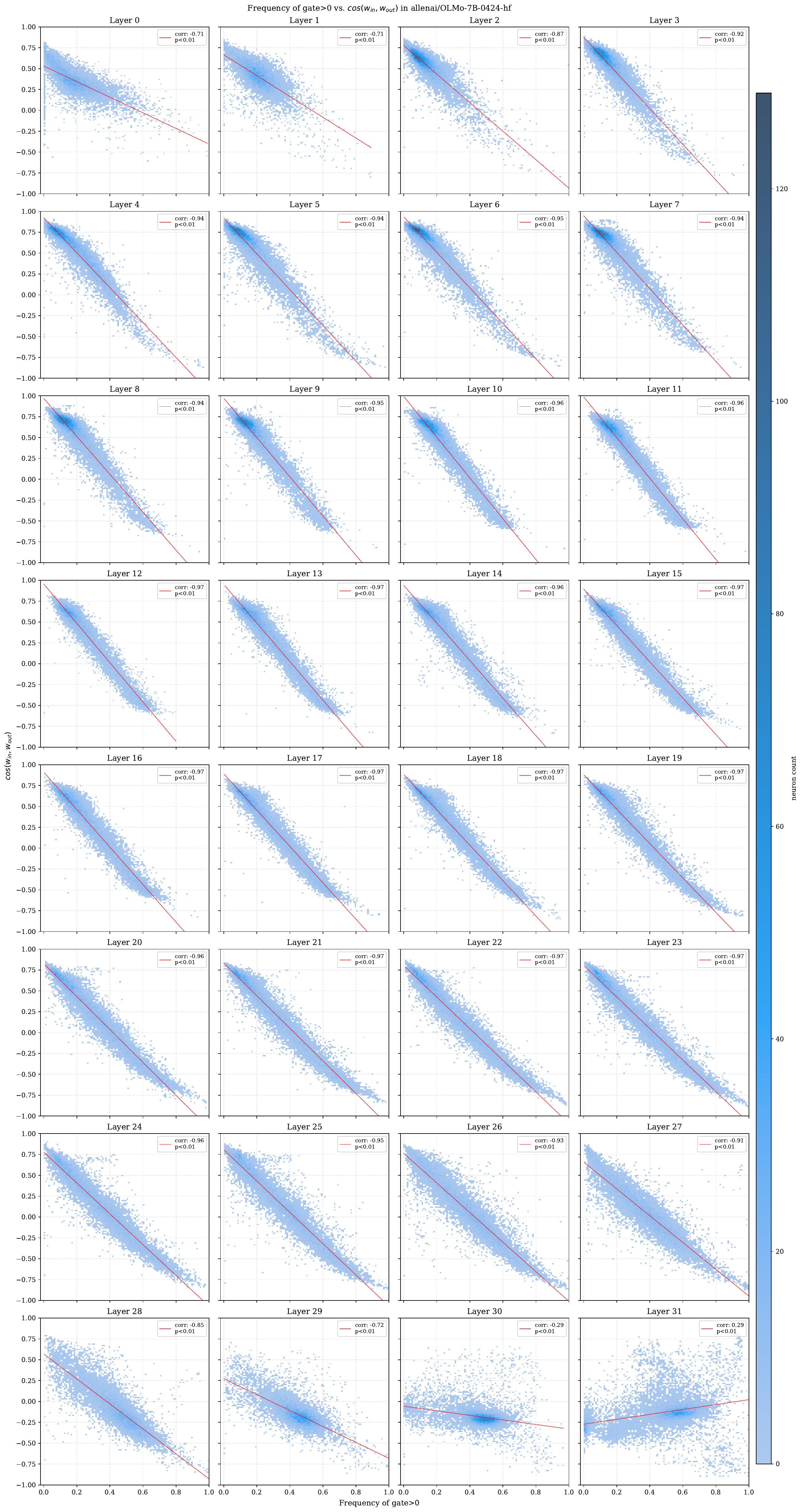}
	\caption{
		Like \cref{fig:freq}
		but for all layers separately.
	}
	\label{fig:freq-all}
\end{figure*}

\Cref{fig:freq-all}
shows activation frequencies vs. IO cosines in OLMo-7B, on all layers separately. See the \href{https://github.com/sjgerstner/RW_functionalities_results}{supplementary material} for other models, tables by discrete class, and plots against $|\cos(\wgate,\wout)|$ or $\cos(\wgate,\win)$.

The last layer displays a different pattern than the rest
(last subplot in \cref{fig:freq-all}).
Here the correlation is positive ($+0.29$),
and we can distinguish two clusters of neurons:
One cluster has a medium-negative $\cos(\win,\wout)$ (around $-0.3$)
and activates very rarely;
another one is much more spread out (both in terms of $\cos(\win,\wout)$ and activation frequency),
centers at a weaker negative cosine similarity ($-0.1$ to $-0.2$)
and activates a bit more than half of the time.
The presence of these two clusters leads to the slightly positive correlation.
Comparing with the other plots suggests that
the first cluster mostly corresponds to weakening neurons and atypical proportional change neurons.

We do not find such striking patterns with gate-out or gate-in similarities.

\clearpage

\subsection{Neuron ablations}
\label{ap:hr-ablations}
\label{ap:hr-entropy}

\begin{figure}[h]
	\centering
	\includegraphics
	[width=\linewidth]
	{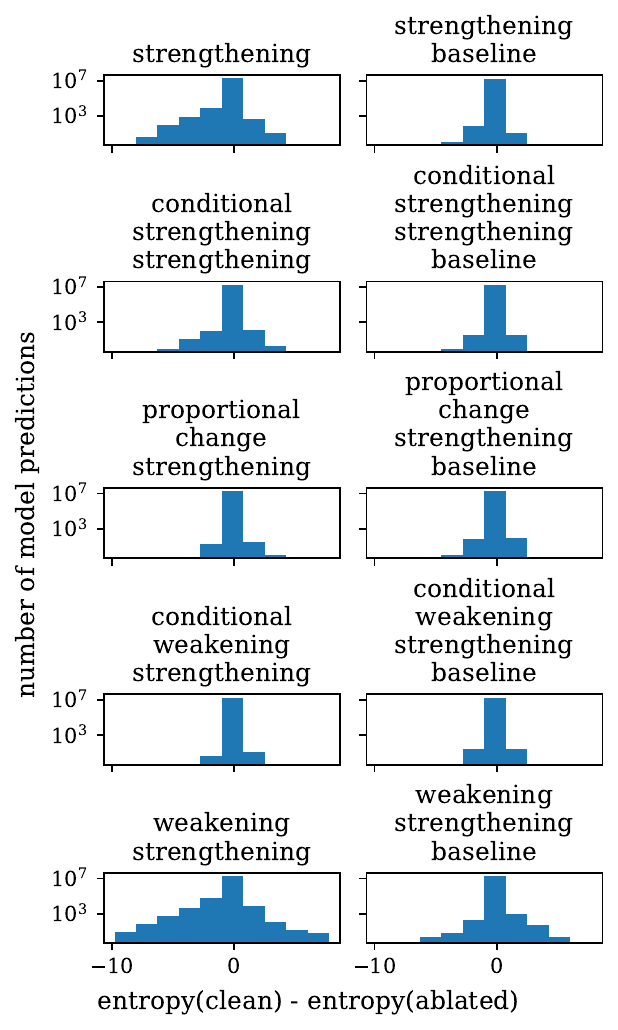}
	\caption{Effect on entropy of zero-ablations of various neuron classes (ablating as many neurons as there are strengthening neurons).}
	\label{fig:entropy-nstrengthening-zero}
\end{figure}

\begin{figure}
	\centering
	\includegraphics
	[width=\linewidth]
	{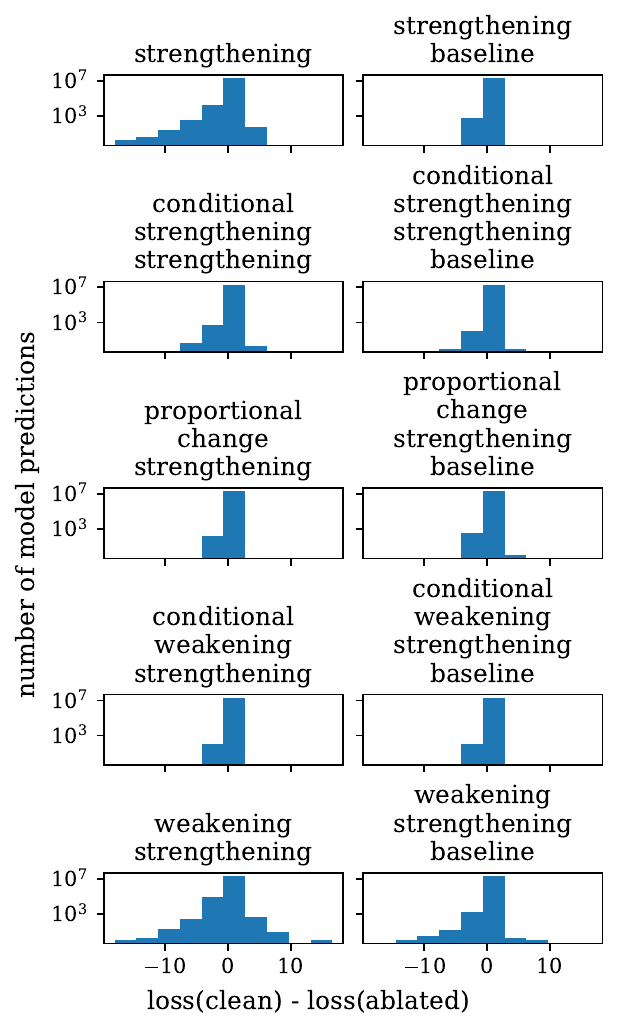}
	\caption{Effect on loss of zero-ablations of various neuron classes (ablating as many neurons as there are strengthening neurons).}
	\label{fig:loss-nstrengthening-zero}
\end{figure}

\begin{figure}
	\centering
	\includegraphics
	[width=\linewidth]
	{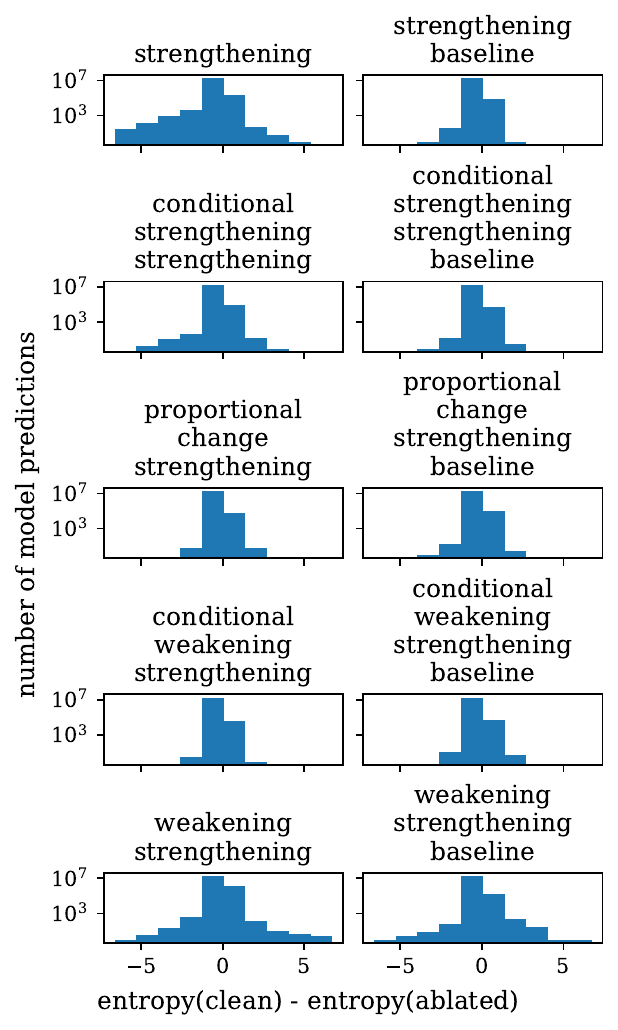}
	\caption{Effect on entropy of mean-ablations of various neuron classes (ablating as many neurons as there are strengthening neurons).}
	\label{fig:entropy-nstrengthening-mean}
\end{figure}

\begin{figure}
	\centering
	\includegraphics
	[width=\linewidth]
	{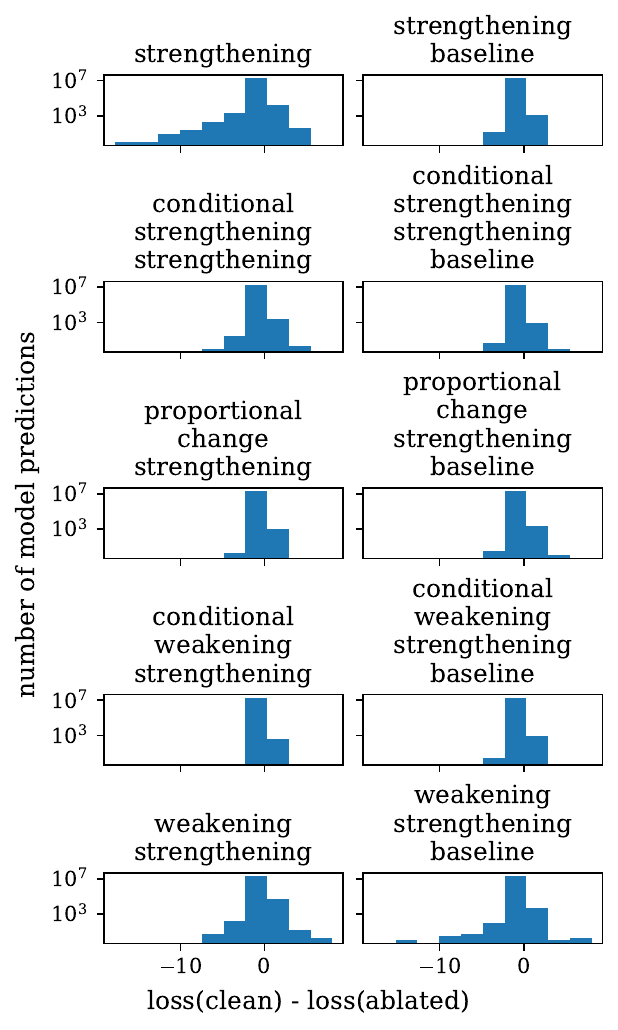}
	\caption{Effect on loss of mean-ablations of various neuron classes (ablating as many neurons as there are strengthening neurons).}
	\label{fig:loss-nstrengthening-mean}
\end{figure}

\begin{figure}
	\centering
	\includegraphics
	[width=\linewidth]
	{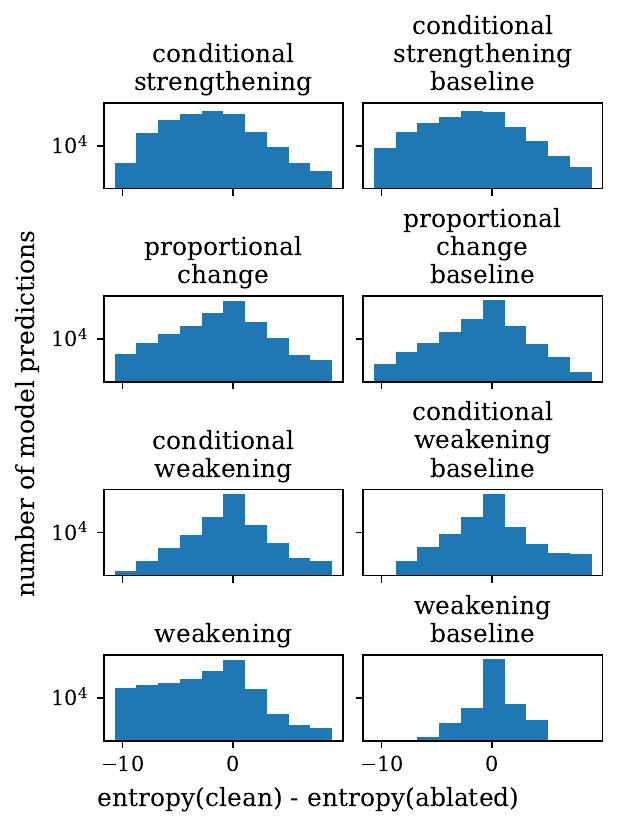}
	\caption{Effect on entropy of zero-ablations of various neuron classes (ablating as many neurons as there are weakening neurons).}
	\label{fig:entropy-nweakening-zero}
\end{figure}

\begin{figure}
	\centering
	\includegraphics
	[width=\linewidth]
	{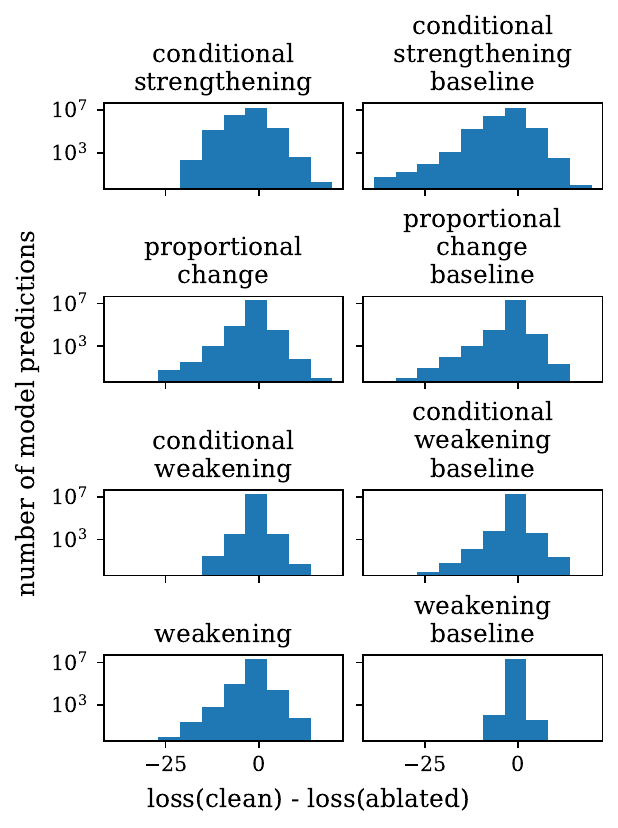}
	\caption{Effect on loss of zero-ablations of various neuron classes (ablating as many neurons as there are weakening neurons).}
	\label{fig:loss-nweakening-zero}
\end{figure}

\begin{figure}
	\centering
	\includegraphics
	[width=\linewidth]
	{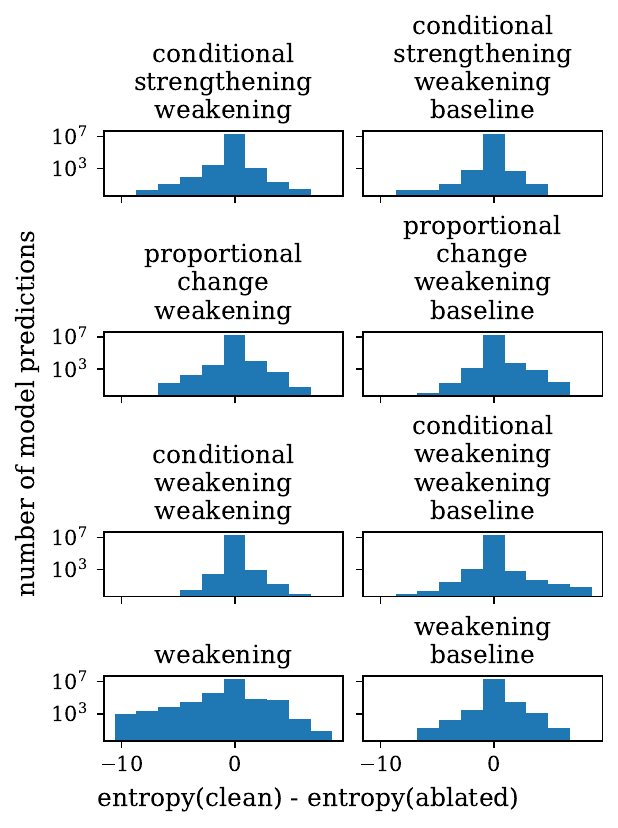}
	\caption{Effect on entropy of mean-ablations of various neuron classes (ablating as many neurons as there are weakening neurons).}
	\label{fig:entropy-nweakening-mean}
\end{figure}

\begin{figure}
	\centering
	\includegraphics
	[width=\linewidth]
	{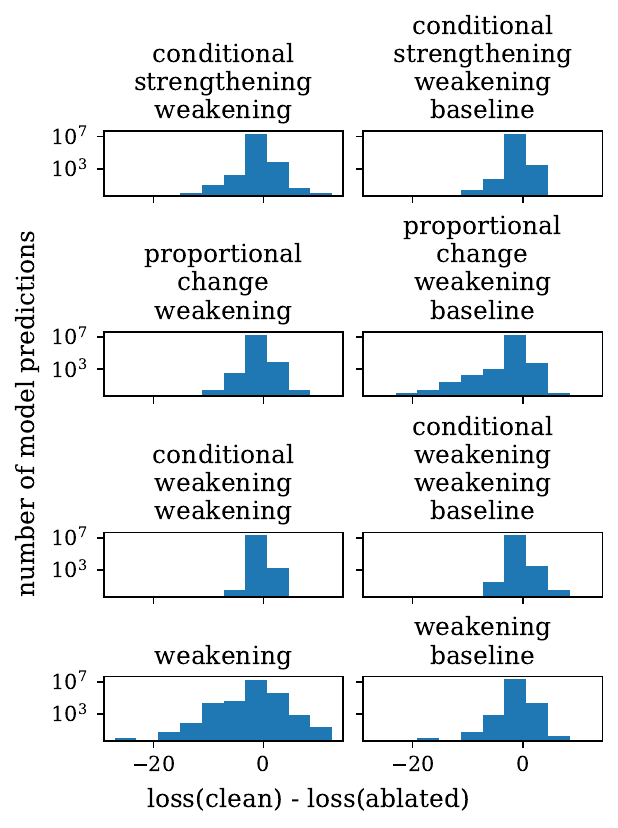}
	\caption{Effect on loss of mean-ablations of various neuron classes (ablating as many neurons as there are weakening neurons).}
	\label{fig:loss-nweakening-mean}
\end{figure}

\begin{figure}
	\centering
	\includegraphics
	[width=\linewidth]
	{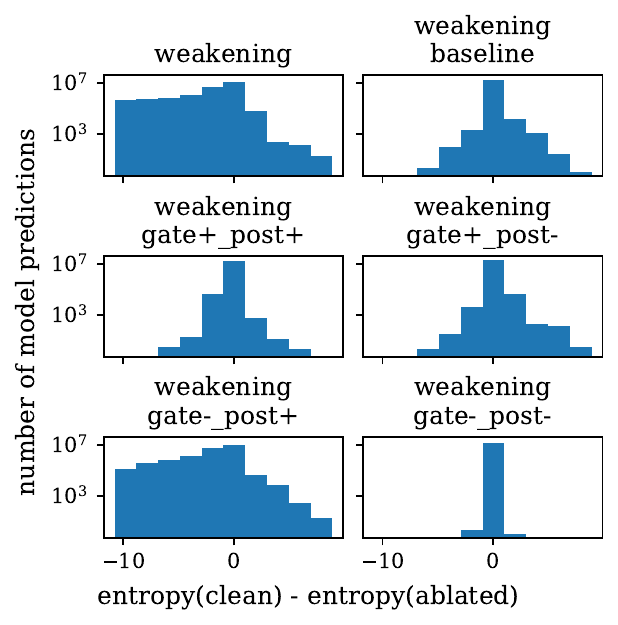}
	\caption{Effect on entropy of conditional zero-ablations of weakening neurons.}
	\label{fig:entropy-weakening}
\end{figure}

\begin{figure}
	\centering
	\includegraphics
	[width=\linewidth]
	{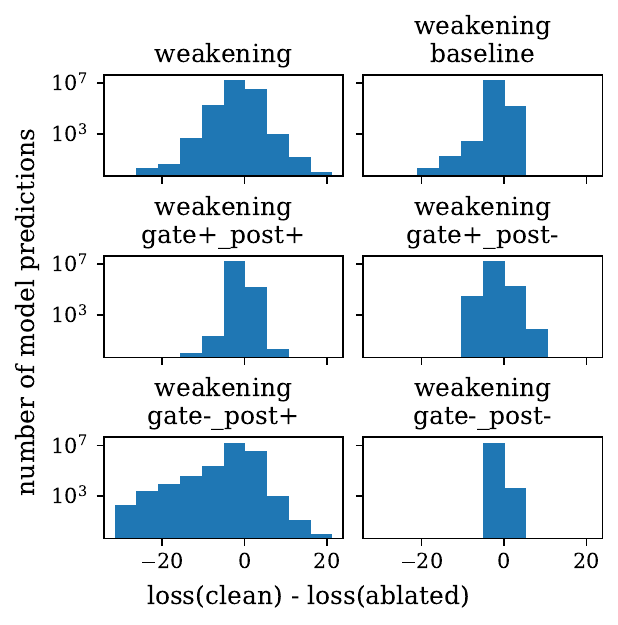}
	\caption{Effect on loss of conditional zero-ablations of weakening neurons.}
	\label{fig:loss-weakening}
\end{figure}

\begin{figure}
	\centering
	\includegraphics
	[width=\linewidth]
	{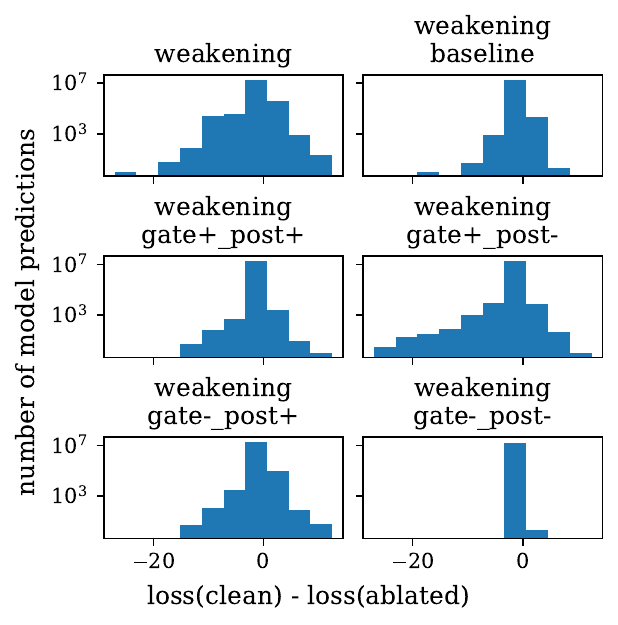}
	\caption{Effect on loss of conditional mean-ablations of weakening neurons.}
	\label{fig:loss-weakening-mean}
\end{figure}

This section (\cref{fig:entropy-nstrengthening-zero,fig:loss-nstrengthening-zero,fig:entropy-nstrengthening-mean,fig:loss-nstrengthening-mean,fig:entropy-nweakening-zero,fig:loss-nweakening-zero,fig:entropy-nweakening-mean,fig:loss-nweakening-mean,fig:entropy-weakening,fig:loss-weakening,fig:loss-weakening-mean}) contains results for entropy and loss, on OLMo-7B.
The \href{https://github.com/sjgerstner/RW_functionalities_results}{supplementary material} shows the effect of ablations on attribute rate (as described in \cref{ap:attribute}), rank of correct output token, and scale of last hidden state vector, as well as equivalent results on Llama-3.2-3B.



\clearpage

\subsection{Distributions of neuron weight cosines by model and layer}
\label{ap:hr-distributions}

\subsubsection{Non-GLU models}
\label{ap:hr-non-glu}

\begin{figure}
	\includegraphics[width=3.25in]{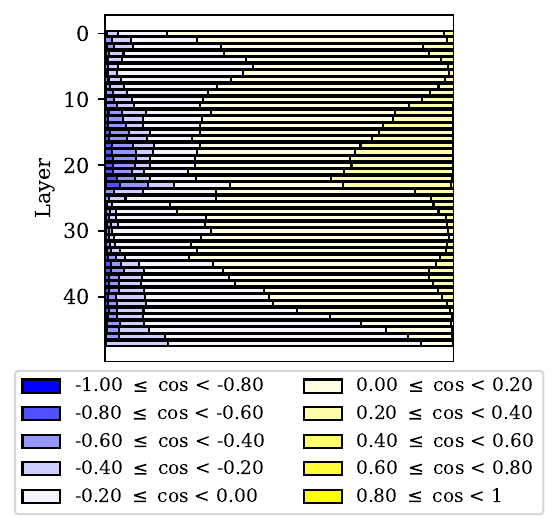}
	\caption{
		Distribution of neurons by layer and input-output weight cosines in T5-large models (visualization of \cref{tab:t5-large_half_coarse}).
		Layers 0-23 correspond to the encoder, and 24-47 to the decoder.
	}
	\label{fig:t5}
\end{figure}

\begin{figure}
	\includegraphics[width=3.25in]{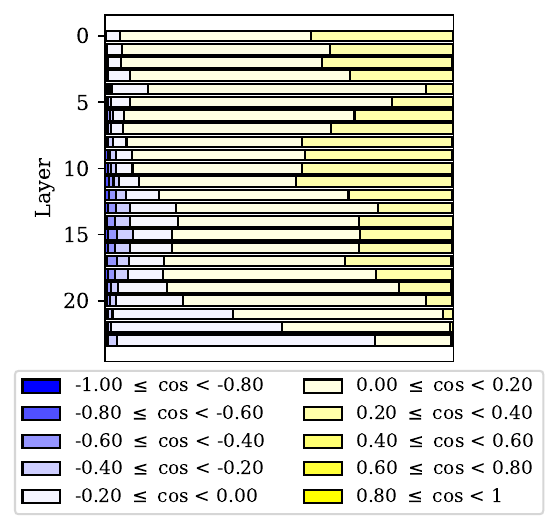}
	\caption{
		Distribution of neurons by layer and input-output weight cosines in BERT-large-cased (visualization of \cref{tab:bert-large-cased_half_coarse}).
		}
	\label{fig:bert}
\end{figure}

\begin{figure}
	\includegraphics[width=\columnwidth]{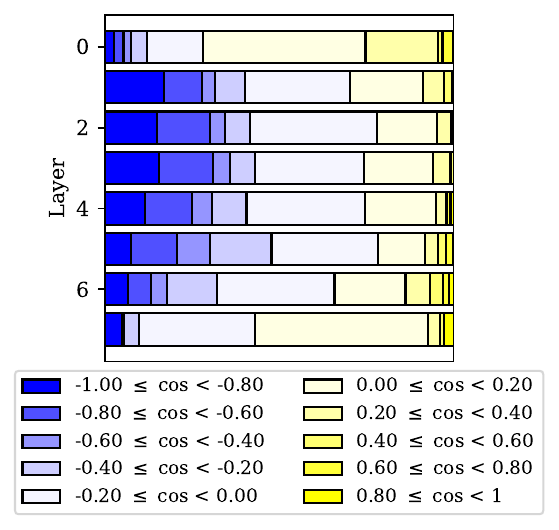}
	\caption{Distribution of neurons by layer and input-output weight cosines in Othello-GPT (visualization of \cref{tab:othello-gpt_half_coarse}).}
	\label{fig:othello-gpt}
\end{figure}

\begin{figure}
	\includegraphics[width=3.25in]{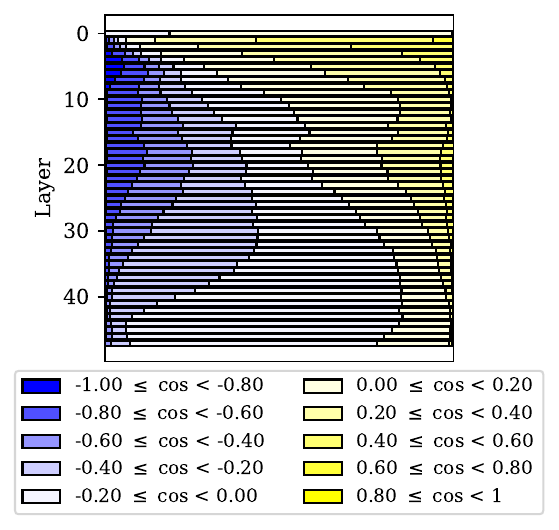}
	\caption{
		Distribution of neurons by layer and input-output weight cosines in GPT2-XL (visualization of \cref{tab:gpt2-xl_half_coarse}).
		}
	\label{fig:gpt}
\end{figure}

Here we show results for a few selected models:
T5-large (\cref{tab:t5-large_half_coarse,fig:t5}),
BERT-large-cased (\cref{tab:bert-large-cased_half_coarse,fig:bert}), Othello-GPT (\cref{tab:othello-gpt_half_coarse,fig:othello-gpt}),
and GPT2-XL (\cref{tab:gpt2-xl_half_coarse,fig:gpt}).
For other models see the \href{https://github.com/sjgerstner/RW_functionalities_results}{supplementary material}.

We can see that many neurons have cosine similarities substantially different from zero.
In particular, there is a sizable number of \textbf{weakening neurons} with cosine similarities below $-0.8$ (
mostly early-middle layers in GPT2,
but it varies across models to some extent).

On the other hand, there are very few strengthening neurons with cosine similarities above $+0.8$.
Some of the table columns stop at $\cos<0.6$ or $\cos<0.8$.
This is not a bug, but reflects the rarity of strengthening neurons:
in these models, there is not a single neuron with a higher IO cosine similarity (so no strengthening neurons in the stricter sense of the word), but there are many neurons with very low negative IO cosine similarities (weakening neurons).

There is also a large number of neurons with moderately non-zero cosine similarities:
cosines between e.g. $0.2$ and $0.6$ (
mostly early layers in GPT2, but again it varies);
and moderately negative cosines between e.g. $-0.2$ and $-0.6$ (
mostly middle-to-late layers in GPT2, but variable overall).

Some of these phenomena have been briefly observed before \citep{Elhage2021mathematicalframeworktransformer,2024_Gurnee},
but to our knowledge we are the first to systematically report them.

\paragraph{T5 models (encoder-decoder).}
\Cref{tab:t5-large_half_coarse,fig:t5}.

Note that in these visualizations the encoder and decoder layers are stacked: The first half of the layers correspond to the encoder module, the second half to the decoder module. Thus, in each of these tables and figures, the upper half corresponds to the encoder and the lower half to the decoder.

\begin{table*}
\caption{Distribution of neuron IO cosines by layer in t5-large. See \cref{fig:t5} for a visualization. Layers 0-23 correspond to the encoder, and 24-47 to the decoder.}
\label{tab:t5-large_half_coarse}
\begin{tabular}{l|r|r|r|r|r|r|r|r}
	& \multicolumn{1}{p{.06\textwidth}|}{-1.00 $\leq$ cos < -0.80} & \multicolumn{1}{p{.06\textwidth}|}{-0.80 $\leq$ cos < -0.60} & \multicolumn{1}{p{.06\textwidth}|}{-0.60 $\leq$ cos < -0.40} & \multicolumn{1}{p{.06\textwidth}|}{-0.40 $\leq$ cos < -0.20} & \multicolumn{1}{p{.06\textwidth}|}{-0.20 $\leq$ cos < 0.00} & \multicolumn{1}{p{.06\textwidth}|}{0.00 $\leq$ cos < 0.20} & \multicolumn{1}{p{.06\textwidth}|}{0.20 $\leq$ cos < 0.40} & \multicolumn{1}{p{.06\textwidth}}{0.40 $\leq$ cos < 0.60} \\
	Layer &  &  &  &  &  &  &  &  \\
	\hline
0 & 0 & 0 & 26 & 134 & 568 & 3253 & 115 & 0 \\
1 & 0 & 0 & 87 & 222 & 779 & 2928 & 80 & 0 \\
2 & 0 & 24 & 79 & 233 & 1028 & 2377 & 355 & 0 \\
3 & 0 & 8 & 45 & 167 & 1180 & 2411 & 285 & 0 \\
4 & 0 & 10 & 32 & 155 & 1465 & 2284 & 150 & 0 \\
5 & 0 & 4 & 33 & 122 & 1586 & 2284 & 67 & 0 \\
6 & 0 & 11 & 30 & 102 & 1476 & 2419 & 58 & 0 \\
7 & 0 & 4 & 40 & 150 & 1289 & 2510 & 103 & 0 \\
8 & 0 & 8 & 57 & 164 & 1087 & 2603 & 177 & 0 \\
9 & 0 & 18 & 75 & 168 & 951 & 2607 & 277 & 0 \\
10 & 0 & 16 & 90 & 210 & 838 & 2572 & 370 & 0 \\
11 & 1 & 14 & 129 & 204 & 766 & 2454 & 527 & 1 \\
12 & 1 & 43 & 158 & 280 & 770 & 2159 & 682 & 3 \\
13 & 0 & 43 & 173 & 232 & 692 & 2243 & 710 & 3 \\
14 & 2 & 37 & 212 & 196 & 675 & 2149 & 821 & 4 \\
15 & 1 & 56 & 222 & 252 & 591 & 2094 & 874 & 6 \\
16 & 0 & 48 & 212 & 231 & 538 & 2112 & 948 & 7 \\
17 & 2 & 82 & 250 & 239 & 542 & 1888 & 1088 & 5 \\
18 & 1 & 79 & 291 & 197 & 512 & 1863 & 1149 & 4 \\
19 & 2 & 92 & 268 & 166 & 533 & 1816 & 1215 & 4 \\
20 & 4 & 89 & 263 & 176 & 527 & 1839 & 1194 & 4 \\
21 & 2 & 98 & 216 & 142 & 517 & 1832 & 1285 & 4 \\
22 & 13 & 127 & 210 & 185 & 539 & 1580 & 1430 & 12 \\
23 & 13 & 168 & 326 & 309 & 653 & 1325 & 1268 & 34 \\
\hline
24 & 0 & 5 & 107 & 336 & 855 & 2343 & 450 & 0 \\
25 & 0 & 4 & 48 & 191 & 1065 & 2525 & 263 & 0 \\
26 & 0 & 5 & 23 & 53 & 682 & 3098 & 235 & 0 \\
27 & 0 & 16 & 49 & 69 & 713 & 3072 & 177 & 0 \\
28 & 0 & 13 & 33 & 84 & 1059 & 2819 & 88 & 0 \\
29 & 0 & 25 & 43 & 84 & 1028 & 2840 & 76 & 0 \\
30 & 0 & 11 & 39 & 38 & 1162 & 2784 & 62 & 0 \\
31 & 0 & 14 & 33 & 64 & 1018 & 2910 & 57 & 0 \\
32 & 0 & 19 & 37 & 60 & 901 & 3004 & 75 & 0 \\
33 & 0 & 21 & 57 & 97 & 911 & 2923 & 87 & 0 \\
34 & 0 & 36 & 92 & 104 & 761 & 2947 & 156 & 0 \\
35 & 1 & 71 & 116 & 214 & 872 & 2619 & 203 & 0 \\
36 & 1 & 74 & 149 & 241 & 920 & 2420 & 291 & 0 \\
37 & 4 & 43 & 120 & 266 & 1021 & 2355 & 287 & 0 \\
38 & 2 & 49 & 121 & 253 & 1109 & 2319 & 243 & 0 \\
39 & 4 & 50 & 112 & 290 & 1410 & 2047 & 183 & 0 \\
40 & 1 & 34 & 96 & 336 & 1464 & 2028 & 137 & 0 \\
41 & 5 & 38 & 100 & 341 & 1493 & 2042 & 77 & 0 \\
42 & 7 & 47 & 89 & 312 & 1802 & 1800 & 39 & 0 \\
43 & 0 & 32 & 107 & 308 & 2203 & 1423 & 22 & 1 \\
44 & 0 & 29 & 153 & 277 & 2548 & 1060 & 29 & 0 \\
45 & 0 & 29 & 139 & 356 & 2783 & 774 & 15 & 0 \\
46 & 0 & 15 & 156 & 538 & 2857 & 517 & 12 & 1 \\
47 & 0 & 1 & 16 & 726 & 2970 & 373 & 8 & 2 \\
\hline
\textbf{Total} & 67 & 1760 & 5559 & 10274 & 53709 & 106644 & 18500 & 95 \\
\end{tabular}
\end{table*}

\paragraph{BERT models (encoder-only).}
\Cref{tab:bert-large-cased_half_coarse,fig:bert}.

\begin{table*}
\caption{Distribution of neuron IO cosines by layer in bert-large-cased. See \cref{fig:bert} for a visualization.}
\label{tab:bert-large-cased_half_coarse}
\begin{tabular}{l|r|r|r|r|r|r|r|r|r}
	& \multicolumn{1}{p{.06\textwidth}|}{-1.00 $\leq$ cos < -0.80} & \multicolumn{1}{p{.06\textwidth}|}{-0.80 $\leq$ cos < -0.60} & \multicolumn{1}{p{.06\textwidth}|}{-0.60 $\leq$ cos < -0.40} & \multicolumn{1}{p{.06\textwidth}|}{-0.40 $\leq$ cos < -0.20} & \multicolumn{1}{p{.06\textwidth}|}{-0.20 $\leq$ cos < 0.00} & \multicolumn{1}{p{.06\textwidth}|}{0.00 $\leq$ cos < 0.20} & \multicolumn{1}{p{.06\textwidth}|}{0.20 $\leq$ cos < 0.40} & \multicolumn{1}{p{.06\textwidth}|}{0.40 $\leq$ cos < 0.60} & \multicolumn{1}{p{.06\textwidth}}{0.60 $\leq$ cos < 0.80} \\
	Layer &  &  &  &  &  &  &  & & \\
	\hline
0 & 0 & 1 & 4 & 10 & 160 & 2249 & 1671 & 1 & 0 \\
1 & 1 & 4 & 5 & 20 & 176 & 2442 & 1442 & 6 & 0 \\
2 & 7 & 10 & 4 & 13 & 158 & 2354 & 1527 & 22 & 1 \\
3 & 2 & 6 & 7 & 22 & 255 & 2584 & 1212 & 7 & 1 \\
4 & 4 & 25 & 26 & 35 & 422 & 3262 & 321 & 1 & 0 \\
5 & 5 & 12 & 25 & 29 & 225 & 3081 & 714 & 5 & 0 \\
6 & 2 & 27 & 35 & 37 & 125 & 2707 & 1158 & 5 & 0 \\
7 & 1 & 14 & 21 & 41 & 142 & 2443 & 1426 & 7 & 1 \\
8 & 1 & 16 & 20 & 56 & 162 & 2058 & 1770 & 11 & 2 \\
9 & 6 & 27 & 32 & 69 & 188 & 2028 & 1734 & 10 & 2 \\
10 & 3 & 32 & 34 & 68 & 188 & 1994 & 1760 & 14 & 3 \\
11 & 6 & 41 & 55 & 70 & 230 & 1844 & 1834 & 16 & 0 \\
12 & 1 & 53 & 74 & 124 & 389 & 2221 & 1218 & 15 & 1 \\
13 & 0 & 37 & 97 & 167 & 531 & 2375 & 877 & 12 & 0 \\
14 & 1 & 27 & 96 & 177 & 559 & 2125 & 1103 & 6 & 2 \\
15 & 1 & 39 & 107 & 181 & 463 & 2207 & 1079 & 18 & 1 \\
16 & 0 & 35 & 90 & 173 & 486 & 2201 & 1094 & 15 & 2 \\
17 & 0 & 28 & 111 & 145 & 416 & 2117 & 1255 & 22 & 2 \\
18 & 0 & 33 & 90 & 145 & 415 & 2503 & 894 & 15 & 1 \\
19 & 0 & 29 & 47 & 74 & 578 & 2724 & 620 & 22 & 2 \\
20 & 0 & 27 & 37 & 63 & 793 & 2848 & 306 & 21 & 1 \\
21 & 0 & 20 & 22 & 49 & 1416 & 2467 & 111 & 10 & 1 \\
22 & 1 & 11 & 20 & 43 & 2003 & 1977 & 32 & 9 & 0 \\
23 & 0 & 8 & 32 & 103 & 3032 & 893 & 26 & 2 & 0 \\
\textbf{Total} & 42 & 562 & 1091 & 1914 & 13512 & 55704 & 25184 & 272 & 23 \\
\end{tabular}
\end{table*}

\paragraph{Othello-GPT (decoder, non-language).}
\Cref{tab:othello-gpt_half_coarse,fig:othello-gpt}.

\begin{table*}
\caption{Distribution of neuron IO cosines by layer in othello-gpt. See \cref{fig:othello-gpt} for a visualization.}
\label{tab:othello-gpt_half_coarse}
\begin{tabular}{l|r|r|r|r|r|r|r|r|r|r}
	& \multicolumn{1}{p{.06\textwidth}|}{-1.00 $\leq$ cos < -0.80} & \multicolumn{1}{p{.06\textwidth}|}{-0.80 $\leq$ cos < -0.60} & \multicolumn{1}{p{.06\textwidth}|}{-0.60 $\leq$ cos < -0.40} & \multicolumn{1}{p{.06\textwidth}|}{-0.40 $\leq$ cos < -0.20} & \multicolumn{1}{p{.06\textwidth}|}{-0.20 $\leq$ cos < 0.00} & \multicolumn{1}{p{.06\textwidth}|}{0.00 $\leq$ cos < 0.20} & \multicolumn{1}{p{.06\textwidth}|}{0.20 $\leq$ cos < 0.40} & \multicolumn{1}{p{.06\textwidth}|}{0.40 $\leq$ cos < 0.60} & \multicolumn{1}{p{.06\textwidth}|}{0.60 $\leq$ cos < 0.80} & \multicolumn{1}{p{.06\textwidth}}{0.80 $\leq$ cos < 1.00} \\
	Layer &  &  &  &  &  &  &  &  &  &  \\
	\hline
0 & 55 & 55 & 43 & 96 & 328 & 954 & 424 & 28 & 63 & 2 \\
1 & 347 & 223 & 77 & 174 & 619 & 429 & 125 & 47 & 7 & 0 \\
2 & 309 & 310 & 86 & 150 & 741 & 356 & 79 & 15 & 1 & 1 \\
3 & 319 & 315 & 102 & 147 & 639 & 407 & 101 & 18 & 0 & 0 \\
4 & 238 & 275 & 119 & 200 & 695 & 417 & 59 & 7 & 20 & 18 \\
5 & 154 & 269 & 195 & 361 & 626 & 273 & 81 & 47 & 37 & 5 \\
6 & 137 & 132 & 94 & 296 & 690 & 417 & 142 & 77 & 39 & 24 \\
7 & 99 & 6 & 9 & 86 & 683 & 1016 & 69 & 22 & 3 & 55 \\
\textbf{Total} & 1658 & 1585 & 725 & 1510 & 5021 & 4269 & 1080 & 261 & 170 & 105 \\
\end{tabular}
\end{table*}

\paragraph{Decoder-only language models (non-GLU).}

\Cref{tab:gpt2-xl_half_coarse,fig:gpt}.

\begin{table*}
\caption{Distribution of neuron IO cosines by layer in gpt2-xl. See \cref{fig:gpt} for a visualization.}
\label{tab:gpt2-xl_half_coarse}
\begin{tabular}{l|r|r|r|r|r|r|r|r|r|r}
	& \multicolumn{1}{p{.06\textwidth}|}{-1.00 $\leq$ cos < -0.80} & \multicolumn{1}{p{.06\textwidth}|}{-0.80 $\leq$ cos < -0.60} & \multicolumn{1}{p{.06\textwidth}|}{-0.60 $\leq$ cos < -0.40} & \multicolumn{1}{p{.06\textwidth}|}{-0.40 $\leq$ cos < -0.20} & \multicolumn{1}{p{.06\textwidth}|}{-0.20 $\leq$ cos < 0.00} & \multicolumn{1}{p{.06\textwidth}|}{0.00 $\leq$ cos < 0.20} & \multicolumn{1}{p{.06\textwidth}|}{0.20 $\leq$ cos < 0.40} & \multicolumn{1}{p{.06\textwidth}|}{0.40 $\leq$ cos < 0.60} & \multicolumn{1}{p{.06\textwidth}|}{0.60 $\leq$ cos < 0.80} & \multicolumn{1}{p{.06\textwidth}}{0.80 $\leq$ cos < 1.00} \\
	Layer &  &  &  &  &  &  &  &  &  &  \\
	\hline
0 & 0 & 0 & 0 & 0 & 1188 & 5180 & 32 & 0 & 0 & 0 \\
1 & 15 & 58 & 88 & 89 & 137 & 526 & 1866 & 3246 & 374 & 1 \\
2 & 35 & 133 & 107 & 120 & 257 & 1053 & 2820 & 1865 & 10 & 0 \\
3 & 136 & 233 & 152 & 142 & 365 & 1494 & 2925 & 947 & 6 & 0 \\
4 & 320 & 304 & 187 & 219 & 422 & 1646 & 2685 & 610 & 7 & 0 \\
5 & 354 & 375 & 258 & 274 & 552 & 1964 & 2277 & 341 & 5 & 0 \\
6 & 296 & 504 & 279 & 326 & 659 & 1970 & 2112 & 251 & 3 & 0 \\
7 & 179 & 534 & 307 & 373 & 872 & 2191 & 1780 & 151 & 13 & 0 \\
8 & 94 & 542 & 365 & 474 & 1026 & 2258 & 1519 & 115 & 7 & 0 \\
9 & 46 & 563 & 413 & 586 & 1319 & 2332 & 1066 & 68 & 7 & 0 \\
10 & 43 & 641 & 454 & 652 & 1438 & 2154 & 963 & 47 & 8 & 0 \\
11 & 34 & 626 & 528 & 705 & 1495 & 2032 & 928 & 43 & 9 & 0 \\
12 & 20 & 651 & 556 & 777 & 1472 & 1943 & 919 & 58 & 4 & 0 \\
13 & 39 & 639 & 665 & 804 & 1390 & 1834 & 982 & 44 & 3 & 0 \\
14 & 30 & 635 & 770 & 976 & 1268 & 1524 & 1059 & 134 & 4 & 0 \\
15 & 40 & 484 & 819 & 1000 & 1414 & 1617 & 930 & 94 & 2 & 0 \\
16 & 41 & 568 & 809 & 896 & 1261 & 1698 & 1011 & 115 & 1 & 0 \\
17 & 27 & 697 & 779 & 823 & 1078 & 1591 & 1215 & 190 & 0 & 0 \\
18 & 35 & 739 & 829 & 879 & 1137 & 1372 & 1185 & 224 & 0 & 0 \\
19 & 34 & 668 & 885 & 930 & 1085 & 1487 & 1070 & 240 & 1 & 0 \\
20 & 26 & 703 & 891 & 981 & 1125 & 1352 & 1083 & 237 & 2 & 0 \\
21 & 32 & 621 & 891 & 1039 & 1240 & 1365 & 982 & 229 & 1 & 0 \\
22 & 27 & 558 & 872 & 1051 & 1297 & 1393 & 976 & 225 & 1 & 0 \\
23 & 19 & 472 & 913 & 1161 & 1426 & 1384 & 831 & 193 & 1 & 0 \\
24 & 15 & 417 & 1005 & 1271 & 1508 & 1317 & 715 & 152 & 0 & 0 \\
25 & 20 & 353 & 927 & 1398 & 1647 & 1270 & 617 & 167 & 1 & 0 \\
26 & 18 & 282 & 942 & 1537 & 1705 & 1192 & 581 & 142 & 1 & 0 \\
27 & 17 & 257 & 804 & 1629 & 1908 & 1138 & 535 & 111 & 1 & 0 \\
28 & 12 & 200 & 770 & 1803 & 1893 & 1111 & 479 & 131 & 1 & 0 \\
29 & 7 & 149 & 718 & 1853 & 2043 & 1049 & 467 & 114 & 0 & 0 \\
30 & 13 & 132 & 695 & 1979 & 2174 & 932 & 371 & 101 & 3 & 0 \\
31 & 13 & 121 & 591 & 2077 & 2239 & 924 & 359 & 75 & 1 & 0 \\
32 & 11 & 122 & 479 & 2172 & 2340 & 878 & 345 & 51 & 2 & 0 \\
33 & 8 & 101 & 393 & 2231 & 2557 & 782 & 268 & 59 & 1 & 0 \\
34 & 10 & 73 & 337 & 2219 & 2660 & 796 & 251 & 51 & 3 & 0 \\
35 & 3 & 78 & 250 & 2095 & 2927 & 758 & 240 & 44 & 5 & 0 \\
36 & 6 & 64 & 194 & 2105 & 3012 & 751 & 225 & 40 & 3 & 0 \\
37 & 4 & 53 & 159 & 1889 & 3318 & 753 & 197 & 25 & 2 & 0 \\
38 & 7 & 44 & 122 & 1746 & 3504 & 784 & 168 & 21 & 4 & 0 \\
39 & 5 & 36 & 84 & 1525 & 3795 & 807 & 117 & 27 & 4 & 0 \\
40 & 8 & 39 & 90 & 1146 & 4152 & 832 & 119 & 6 & 8 & 0 \\
41 & 5 & 22 & 65 & 861 & 4505 & 851 & 73 & 13 & 5 & 0 \\
42 & 9 & 29 & 57 & 623 & 4679 & 940 & 53 & 7 & 3 & 0 \\
43 & 11 & 21 & 70 & 401 & 4960 & 877 & 44 & 10 & 6 & 0 \\
44 & 6 & 21 & 85 & 270 & 5024 & 949 & 29 & 13 & 3 & 0 \\
45 & 5 & 20 & 99 & 265 & 4888 & 1064 & 29 & 22 & 7 & 1 \\
46 & 0 & 13 & 126 & 262 & 4717 & 1225 & 28 & 16 & 12 & 1 \\
47 & 2 & 5 & 113 & 339 & 4538 & 1316 & 55 & 21 & 11 & 0 \\
\textbf{Total} & 2137 & 14600 & 21992 & 48973 & 101616 & 66656 & 39581 & 11086 & 556 & 3 \\
\end{tabular}
\end{table*}

\clearpage

\subsubsection{GLU models}
\label{ap:hr-glu}

Here we only include the results for OLMo-1B, OLMo-7B, Llama-3.2-3B, and Yi-6B.
See the \href{https://github.com/sjgerstner/RW_functionalities_results}{supplementary material}
for additional models and plot types.

We note a few additional patterns that appear only in some of the investigated models:
\begin{itemize}
	\item In Yi and the OLMo models, the prevalence of conditional strengthening neurons starts even earlier, at the very first layer.
	A particularly interesting example is Yi:
	In layer 0 an enormous 68\% of all neurons are conditional strengthening,
	then almost none,
	then there is a second wave around layers 11-17 (out of 32) which have around 25\% of conditional strengthening neurons each.
	\item In some models, especially the OLMo ones, there is a non-negligible number of conditional weakening neurons. They tend to appear in middle-to-late layers, shortly after the conditional strengthening wave.
	The clearest example is OLMo-1B, with a peak of 1418 conditional weakening neurons out of 8192 (17\%) in layer 9 out of 16.
\end{itemize}

The following patterns could be random,
but still show that the model has \textit{not} learned something:
\begin{itemize}
	\item
	For almost all neurons
	the cosine similarities are still clearly below $1$ (the dots do not fill out the edges in \cref{fig:wcos_selected}).
	This echoes and extends
	Gurnee et al.'s findings \citep{2024_Gurnee}
	that in GPT2  the IO
	cosine similarity is approximately bounded by $\pm
	0.8$.
	In other words, we almost never get the \textit{prototypical} cases
	of conditional strengthening / weakening etc.,
	as defined in \cref{sec:theory}.
	This might be an effect of randomness
	(strong cosine similarities are less likely),
	but could also suggest that
	even input manipulator neurons
	add some novel information
	to the residual stream.
	\item
	We also observe that for the vast majority of neurons,
	$\cos(\wgate,\win) \approx 0$:
	This can be seen in the boxplots in the \href{https://github.com/sjgerstner/RW_functionalities_results}{supplementary material},
	as well as the purple color in \cref{fig:wcos_selected}.
	Thus most neurons operate on two input directions in the
	residual stream (not a single one), resulting in higher
	expressivity and more complex semantics.
	If not random, this could be related to double checking;
	see \cref{sec:doublecheck}.
\end{itemize}

\begin{figure*}
	\centering
	\includegraphics
	[width=\linewidth]
	{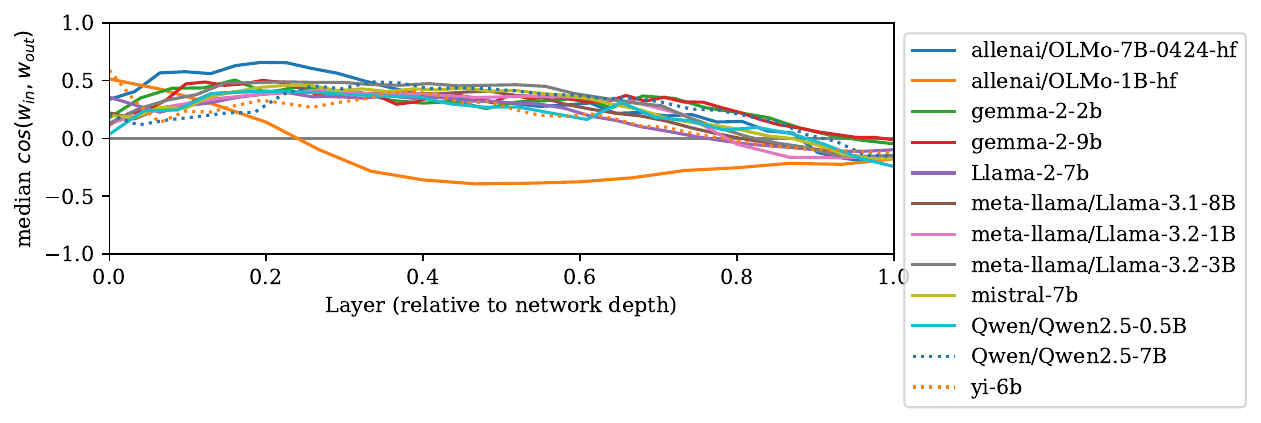}
	\caption{
		Median of $\cos(\win,\wout)$ by layer (x-axis)
		for all 12 models investigated.
		Unlike \cref{fig:medians}
		we also include the models of 1B parameters and below.
		All models follow the same general pattern,
		but OLMo-1B switches to negative values earlier than the others.
	}
	\label{fig:all_medians}
\end{figure*}

\begin{table*}
\caption{Distribution of neuron IO classes by layer and category in allenai/OLMo-1B-hf. See \cref{fig:coarse} for a visualization.}
\label{tab:allenai/OLMo-1B-hf}
\begin{tabular}{l|r|r|r|r|r|r|r|r|r|r|r}
 & \multicolumn{1}{p{.06\textwidth}|}{strength-ening} & \multicolumn{1}{p{.06\textwidth}|}{atypical strengthening} & \multicolumn{1}{p{.06\textwidth}|}{condi-tional strengthening} & \multicolumn{1}{p{.06\textwidth}|}{atypical conditional strengthening} & \multicolumn{1}{p{.06\textwidth}|}{propor-tional change} & \multicolumn{1}{p{.06\textwidth}|}{atypical proportional change} & \multicolumn{1}{p{.06\textwidth}|}{ortho-gonal output} & \multicolumn{1}{p{.06\textwidth}|}{weak-ening} & \multicolumn{1}{p{.06\textwidth}|}{atypical weakening} & \multicolumn{1}{p{.06\textwidth}|}{condi-tional weakening} & \multicolumn{1}{p{.06\textwidth}}{atypical conditional weakening} \\
Layer &  &  &  &  &  &  &  &  &  &  &  \\
\hline
0 & 0 & 0 & 4365 & 1 & 22 & 4 & 3798 & 1 & 0 & 1 & 0 \\
1 & 0 & 6 & 3018 & 0 & 99 & 5 & 5051 & 8 & 2 & 2 & 1 \\
2 & 0 & 2 & 2390 & 0 & 581 & 3 & 4976 & 10 & 11 & 215 & 4 \\
3 & 2 & 7 & 1927 & 2 & 1368 & 2 & 4286 & 17 & 38 & 541 & 2 \\
4 & 1 & 3 & 861 & 1 & 1435 & 4 & 4748 & 31 & 52 & 1051 & 5 \\
5 & 2 & 8 & 325 & 0 & 1256 & 2 & 5516 & 18 & 42 & 1023 & 0 \\
6 & 1 & 5 & 165 & 0 & 937 & 1 & 6026 & 8 & 15 & 1034 & 0 \\
7 & 1 & 2 & 138 & 0 & 685 & 3 & 6044 & 4 & 9 & 1306 & 0 \\
8 & 0 & 2 & 127 & 0 & 594 & 1 & 6228 & 7 & 8 & 1225 & 0 \\
9 & 1 & 0 & 160 & 0 & 543 & 4 & 6038 & 12 & 14 & 1418 & 2 \\
10 & 5 & 0 & 180 & 0 & 649 & 6 & 5932 & 20 & 27 & 1370 & 3 \\
11 & 7 & 5 & 219 & 2 & 558 & 15 & 6033 & 23 & 27 & 1294 & 9 \\
12 & 11 & 5 & 190 & 3 & 567 & 21 & 6107 & 20 & 27 & 1234 & 7 \\
13 & 3 & 0 & 53 & 0 & 466 & 12 & 7118 & 14 & 11 & 504 & 11 \\
14 & 0 & 0 & 8 & 1 & 37 & 5 & 8083 & 17 & 14 & 22 & 5 \\
15 & 7 & 1 & 90 & 2 & 212 & 16 & 7768 & 8 & 15 & 72 & 1 \\
\textbf{Total} & 41 & 46 & 14216 & 12 & 10009 & 104 & 93752 & 218 & 312 & 12312 & 50 \\
\end{tabular}
\end{table*}

\begin{table*}
\caption{Distribution of neuron IO classes by layer and category in allenai/OLMo-7B-0424-hf. See \cref{fig:coarse} for a visualization.}
\label{tab:allenai/OLMo-7B-0424-hf}
\begin{tabular}{l|r|r|r|r|r|r|r|r|r|r|r}
 & \multicolumn{1}{p{.06\textwidth}|}{strength-ening} & \multicolumn{1}{p{.06\textwidth}|}{atypical strengthening} & \multicolumn{1}{p{.06\textwidth}|}{condi-tional strengthening} & \multicolumn{1}{p{.06\textwidth}|}{atypical conditional strengthening} & \multicolumn{1}{p{.06\textwidth}|}{propor-tional change} & \multicolumn{1}{p{.06\textwidth}|}{atypical proportional change} & \multicolumn{1}{p{.06\textwidth}|}{ortho-gonal output} & \multicolumn{1}{p{.06\textwidth}|}{weak-ening} & \multicolumn{1}{p{.06\textwidth}|}{atypical weakening} & \multicolumn{1}{p{.06\textwidth}|}{condi-tional weakening} & \multicolumn{1}{p{.06\textwidth}}{atypical conditional weakening} \\
Layer &  &  &  &  &  &  &  &  &  &  &  \\
\hline
0 & 1 & 0 & 1397 & 619 & 34 & 8 & 8941 & 5 & 0 & 3 & 0 \\
1 & 1 & 2 & 2867 & 3 & 120 & 10 & 7981 & 23 & 1 & 0 & 0 \\
2 & 0 & 6 & 7379 & 6 & 270 & 3 & 3325 & 8 & 5 & 6 & 0 \\
3 & 1 & 10 & 6966 & 1 & 497 & 5 & 3439 & 5 & 5 & 78 & 1 \\
4 & 0 & 14 & 6223 & 0 & 601 & 2 & 4069 & 11 & 9 & 78 & 1 \\
5 & 0 & 38 & 7206 & 2 & 1022 & 1 & 2501 & 19 & 17 & 202 & 0 \\
6 & 0 & 18 & 7480 & 1 & 1156 & 3 & 2117 & 7 & 3 & 222 & 1 \\
7 & 0 & 16 & 7661 & 0 & 921 & 2 & 2217 & 6 & 7 & 178 & 0 \\
8 & 0 & 9 & 7076 & 0 & 839 & 2 & 2881 & 2 & 6 & 193 & 0 \\
9 & 0 & 1 & 6445 & 0 & 893 & 0 & 3458 & 3 & 3 & 205 & 0 \\
10 & 0 & 4 & 5526 & 0 & 710 & 0 & 4532 & 2 & 2 & 232 & 0 \\
11 & 0 & 1 & 4821 & 0 & 697 & 0 & 5145 & 7 & 2 & 335 & 0 \\
12 & 0 & 2 & 4279 & 0 & 642 & 0 & 5817 & 6 & 1 & 261 & 0 \\
13 & 0 & 2 & 3926 & 0 & 685 & 0 & 6018 & 9 & 3 & 365 & 0 \\
14 & 0 & 13 & 3876 & 0 & 887 & 0 & 5733 & 8 & 7 & 484 & 0 \\
15 & 0 & 0 & 3768 & 0 & 923 & 0 & 5791 & 3 & 2 & 521 & 0 \\
16 & 0 & 1 & 3420 & 0 & 773 & 2 & 6324 & 4 & 2 & 482 & 0 \\
17 & 0 & 4 & 3312 & 0 & 636 & 2 & 6536 & 11 & 1 & 506 & 0 \\
18 & 0 & 2 & 3488 & 1 & 687 & 0 & 6258 & 11 & 7 & 554 & 0 \\
19 & 0 & 6 & 3692 & 1 & 866 & 3 & 5632 & 12 & 7 & 789 & 0 \\
20 & 0 & 3 & 3030 & 0 & 1155 & 0 & 6095 & 11 & 8 & 706 & 0 \\
21 & 0 & 5 & 3305 & 0 & 1162 & 1 & 5618 & 9 & 11 & 897 & 0 \\
22 & 0 & 6 & 3075 & 0 & 1215 & 1 & 5838 & 7 & 8 & 858 & 0 \\
23 & 2 & 8 & 3090 & 2 & 1183 & 2 & 5740 & 17 & 15 & 949 & 0 \\
24 & 1 & 6 & 2342 & 4 & 1012 & 10 & 6823 & 15 & 8 & 786 & 1 \\
25 & 15 & 6 & 2515 & 7 & 1107 & 7 & 6434 & 19 & 15 & 882 & 1 \\
26 & 11 & 18 & 1719 & 10 & 997 & 3 & 7447 & 29 & 34 & 739 & 1 \\
27 & 1 & 2 & 1066 & 1 & 793 & 12 & 8636 & 14 & 25 & 457 & 1 \\
28 & 11 & 10 & 404 & 4 & 504 & 19 & 9748 & 19 & 18 & 265 & 6 \\
29 & 1 & 0 & 50 & 1 & 59 & 4 & 10764 & 36 & 16 & 67 & 10 \\
30 & 1 & 0 & 39 & 7 & 92 & 17 & 10730 & 50 & 14 & 51 & 7 \\
31 & 1 & 0 & 228 & 6 & 676 & 9 & 9788 & 138 & 59 & 89 & 14 \\
\textbf{Total} & 47 & 213 & 121671 & 676 & 23814 & 128 & 192376 & 526 & 321 & 12440 & 44 \\
\end{tabular}
\end{table*}

\begin{table*}
\caption{Distribution of neuron IO classes by layer and category in meta-llama/Llama-3.2-3B. See \cref{fig:bar} for a visualization.}
\label{tab:meta-llama/Llama-3.2-3B}
\begin{tabular}{l|r|r|r|r|r|r|r|r|r|r|r}
 & \multicolumn{1}{p{.06\textwidth}|}{strength-ening} & \multicolumn{1}{p{.06\textwidth}|}{atypical strengthening} & \multicolumn{1}{p{.06\textwidth}|}{condi-tional strengthening} & \multicolumn{1}{p{.06\textwidth}|}{atypical conditional strengthening} & \multicolumn{1}{p{.06\textwidth}|}{propor-tional change} & \multicolumn{1}{p{.06\textwidth}|}{atypical proportional change} & \multicolumn{1}{p{.06\textwidth}|}{ortho-gonal output} & \multicolumn{1}{p{.06\textwidth}|}{weak-ening} & \multicolumn{1}{p{.06\textwidth}|}{atypical weakening} & \multicolumn{1}{p{.06\textwidth}|}{condi-tional weakening} & \multicolumn{1}{p{.06\textwidth}}{atypical conditional weakening} \\
Layer &  &  &  &  &  &  &  &  &  &  &  \\
\hline
0 & 0 & 0 & 176 & 0 & 15 & 0 & 8000 & 0 & 0 & 1 & 0 \\
1 & 0 & 0 & 597 & 0 & 71 & 0 & 7516 & 1 & 0 & 4 & 3 \\
2 & 2 & 0 & 812 & 0 & 52 & 2 & 7309 & 0 & 3 & 10 & 2 \\
3 & 1 & 4 & 1495 & 0 & 67 & 2 & 6599 & 8 & 4 & 11 & 1 \\
4 & 0 & 1 & 3516 & 0 & 77 & 0 & 4567 & 5 & 11 & 11 & 4 \\
5 & 0 & 4 & 3495 & 1 & 102 & 1 & 4546 & 5 & 10 & 26 & 2 \\
6 & 0 & 12 & 3860 & 0 & 148 & 2 & 4127 & 6 & 12 & 20 & 5 \\
7 & 0 & 19 & 3778 & 0 & 312 & 1 & 4048 & 4 & 14 & 14 & 2 \\
8 & 0 & 19 & 3801 & 0 & 366 & 1 & 3966 & 8 & 9 & 20 & 2 \\
9 & 0 & 15 & 3420 & 0 & 421 & 2 & 4286 & 12 & 13 & 21 & 2 \\
10 & 0 & 18 & 3386 & 0 & 542 & 2 & 4203 & 9 & 7 & 22 & 3 \\
11 & 0 & 13 & 3644 & 0 & 474 & 1 & 3992 & 25 & 15 & 15 & 13 \\
12 & 0 & 4 & 3282 & 0 & 297 & 4 & 4551 & 17 & 14 & 15 & 8 \\
13 & 5 & 3 & 3361 & 0 & 173 & 9 & 4579 & 28 & 17 & 9 & 8 \\
14 & 0 & 0 & 3447 & 0 & 89 & 29 & 4557 & 37 & 14 & 12 & 7 \\
15 & 0 & 1 & 3057 & 0 & 103 & 19 & 4955 & 37 & 15 & 2 & 3 \\
16 & 0 & 0 & 1591 & 0 & 87 & 14 & 6441 & 16 & 25 & 12 & 6 \\
17 & 0 & 0 & 1004 & 0 & 73 & 4 & 7062 & 20 & 20 & 7 & 2 \\
18 & 0 & 0 & 627 & 0 & 65 & 6 & 7456 & 13 & 18 & 6 & 1 \\
19 & 0 & 1 & 497 & 0 & 71 & 8 & 7582 & 12 & 17 & 3 & 1 \\
20 & 0 & 0 & 165 & 0 & 62 & 1 & 7921 & 18 & 12 & 10 & 3 \\
21 & 0 & 0 & 78 & 0 & 54 & 1 & 8023 & 12 & 11 & 11 & 2 \\
22 & 0 & 0 & 37 & 0 & 67 & 4 & 8045 & 15 & 13 & 9 & 2 \\
23 & 0 & 0 & 44 & 0 & 134 & 11 & 7954 & 24 & 7 & 13 & 5 \\
24 & 0 & 1 & 39 & 0 & 249 & 9 & 7845 & 22 & 10 & 14 & 3 \\
25 & 11 & 2 & 19 & 1 & 507 & 14 & 7553 & 34 & 11 & 30 & 10 \\
26 & 2 & 0 & 35 & 0 & 753 & 44 & 7088 & 139 & 61 & 60 & 10 \\
27 & 3 & 0 & 34 & 1 & 891 & 81 & 6597 & 309 & 178 & 84 & 14 \\
\textbf{Total} & 24 & 117 & 49297 & 3 & 6322 & 272 & 171368 & 836 & 541 & 472 & 124 \\
\end{tabular}
\end{table*}

\begin{table*}
\caption{Distribution of neuron IO classes by layer and category in yi-6b. See \cref{fig:coarse} for a visualization.}
\label{tab:yi-6b}
\begin{tabular}{l|r|r|r|r|r|r|r|r|r|r|r}
 & \multicolumn{1}{p{.06\textwidth}|}{strength-ening} & \multicolumn{1}{p{.06\textwidth}|}{atypical strengthening} & \multicolumn{1}{p{.06\textwidth}|}{condi-tional strengthening} & \multicolumn{1}{p{.06\textwidth}|}{atypical conditional strengthening} & \multicolumn{1}{p{.06\textwidth}|}{propor-tional change} & \multicolumn{1}{p{.06\textwidth}|}{atypical proportional change} & \multicolumn{1}{p{.06\textwidth}|}{ortho-gonal output} & \multicolumn{1}{p{.06\textwidth}|}{weak-ening} & \multicolumn{1}{p{.06\textwidth}|}{atypical weakening} & \multicolumn{1}{p{.06\textwidth}|}{condi-tional weakening} & \multicolumn{1}{p{.06\textwidth}}{atypical conditional weakening} \\
Layer &  &  &  &  &  &  &  &  &  &  &  \\
\hline
0 & 400 & 63 & 7522 & 150 & 547 & 186 & 2137 & 2 & 0 & 0 & 1 \\
1 & 9 & 68 & 98 & 0 & 128 & 4 & 10688 & 12 & 0 & 1 & 0 \\
2 & 0 & 0 & 9 & 0 & 12 & 2 & 10978 & 5 & 2 & 0 & 0 \\
3 & 0 & 0 & 34 & 0 & 25 & 1 & 10938 & 8 & 2 & 0 & 0 \\
4 & 0 & 0 & 77 & 0 & 36 & 0 & 10889 & 4 & 2 & 0 & 0 \\
5 & 1 & 2 & 181 & 2 & 70 & 3 & 10729 & 13 & 3 & 4 & 0 \\
6 & 0 & 0 & 287 & 0 & 108 & 0 & 10588 & 17 & 8 & 0 & 0 \\
7 & 0 & 0 & 365 & 4 & 128 & 0 & 10481 & 17 & 11 & 2 & 0 \\
8 & 0 & 0 & 253 & 3 & 87 & 2 & 10653 & 4 & 5 & 1 & 0 \\
9 & 0 & 0 & 696 & 0 & 122 & 0 & 10150 & 13 & 15 & 12 & 0 \\
10 & 0 & 0 & 1200 & 0 & 148 & 2 & 9626 & 7 & 10 & 13 & 2 \\
11 & 1 & 1 & 2322 & 0 & 255 & 2 & 8355 & 12 & 17 & 40 & 3 \\
12 & 2 & 7 & 3115 & 0 & 551 & 1 & 7214 & 21 & 25 & 70 & 2 \\
13 & 0 & 6 & 2886 & 0 & 671 & 1 & 7319 & 17 & 20 & 87 & 1 \\
14 & 0 & 2 & 2205 & 0 & 835 & 1 & 7583 & 13 & 23 & 345 & 1 \\
15 & 1 & 2 & 2534 & 0 & 1016 & 3 & 7183 & 16 & 19 & 231 & 3 \\
16 & 0 & 2 & 2133 & 0 & 1082 & 3 & 7360 & 25 & 29 & 371 & 3 \\
17 & 1 & 3 & 1907 & 0 & 1326 & 6 & 7183 & 27 & 49 & 499 & 7 \\
18 & 0 & 0 & 1211 & 0 & 1182 & 2 & 8214 & 13 & 37 & 343 & 6 \\
19 & 0 & 0 & 706 & 0 & 603 & 8 & 9457 & 16 & 33 & 181 & 4 \\
20 & 0 & 0 & 273 & 0 & 410 & 8 & 10108 & 13 & 20 & 172 & 4 \\
21 & 0 & 0 & 71 & 0 & 256 & 9 & 10502 & 14 & 29 & 121 & 6 \\
22 & 0 & 0 & 18 & 0 & 221 & 9 & 10562 & 11 & 25 & 156 & 6 \\
23 & 0 & 0 & 7 & 0 & 130 & 2 & 10771 & 4 & 10 & 75 & 9 \\
24 & 0 & 0 & 8 & 0 & 80 & 2 & 10843 & 1 & 7 & 60 & 7 \\
25 & 0 & 0 & 14 & 0 & 29 & 2 & 10931 & 3 & 3 & 24 & 2 \\
26 & 0 & 0 & 1 & 0 & 22 & 2 & 10953 & 3 & 4 & 22 & 1 \\
27 & 0 & 0 & 2 & 0 & 13 & 0 & 10978 & 3 & 3 & 8 & 1 \\
28 & 0 & 0 & 1 & 0 & 43 & 0 & 10937 & 3 & 5 & 15 & 4 \\
29 & 0 & 0 & 3 & 0 & 136 & 2 & 10724 & 34 & 45 & 58 & 6 \\
30 & 0 & 0 & 2 & 0 & 230 & 10 & 10339 & 82 & 200 & 127 & 18 \\
31 & 0 & 0 & 5 & 0 & 516 & 15 & 10076 & 119 & 146 & 115 & 16 \\
\textbf{Total} & 415 & 156 & 30146 & 159 & 11018 & 288 & 305449 & 552 & 807 & 3153 & 113 \\
\end{tabular}
\end{table*}

\begin{figure*}
	\subfigure{\includegraphics[width=.4\textwidth]{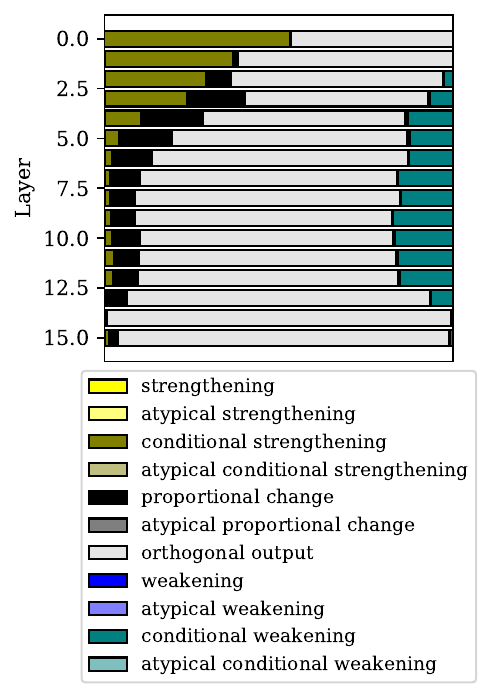}}
	\subfigure{\includegraphics[width=.4\textwidth]{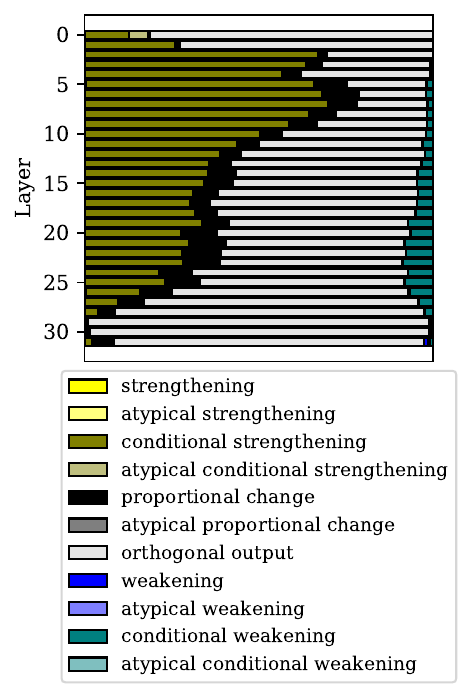}}
	\subfigure{\includegraphics[width=.4\textwidth]{plots/meta-llama/Llama-3.2-3B/refactored/coarse.pdf}}
	\subfigure{\includegraphics[width=.4\textwidth]{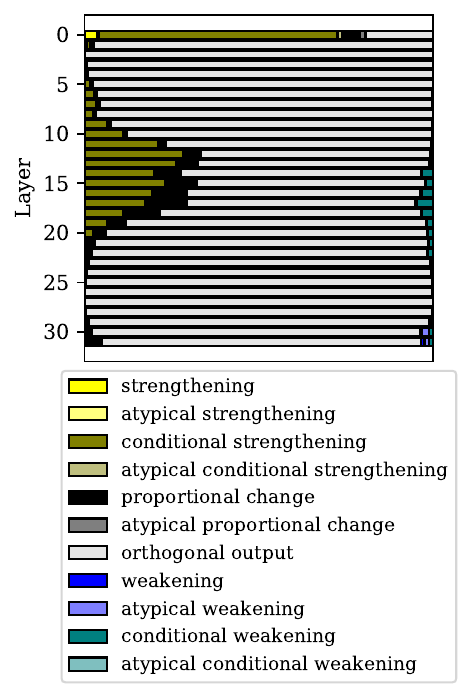}}
	\caption{
		Distribution of neurons by layer and category for a range of models.
		In reading order: OLMo-1B, OLMo-7B-0424, Llama-3.2-3B (copy of \cref{fig:bar} for convenience), Yi-6B.
		Exact numbers in \cref{tab:allenai/OLMo-1B-hf,tab:allenai/OLMo-7B-0424-hf,tab:meta-llama/Llama-3.2-3B,tab:yi-6b}.
		}
	\label{fig:coarse}
\end{figure*}

\begin{figure*}
	\centering
	\includegraphics[width=.67\textwidth]{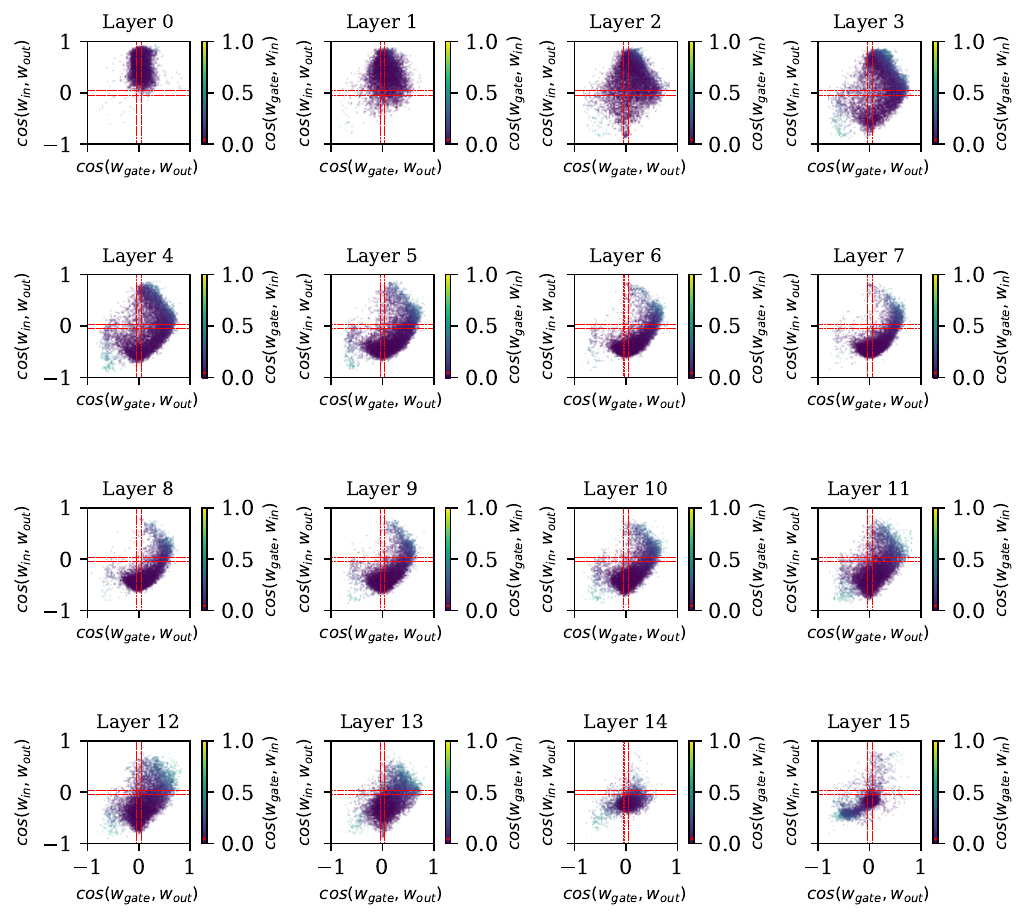}
	\caption{Equivalent of \cref{fig:wcos_selected}
		for OLMo-1B}
\end{figure*}
\clearpage
\begin{figure*}
	\centering
	\includegraphics[width=.67\textwidth]{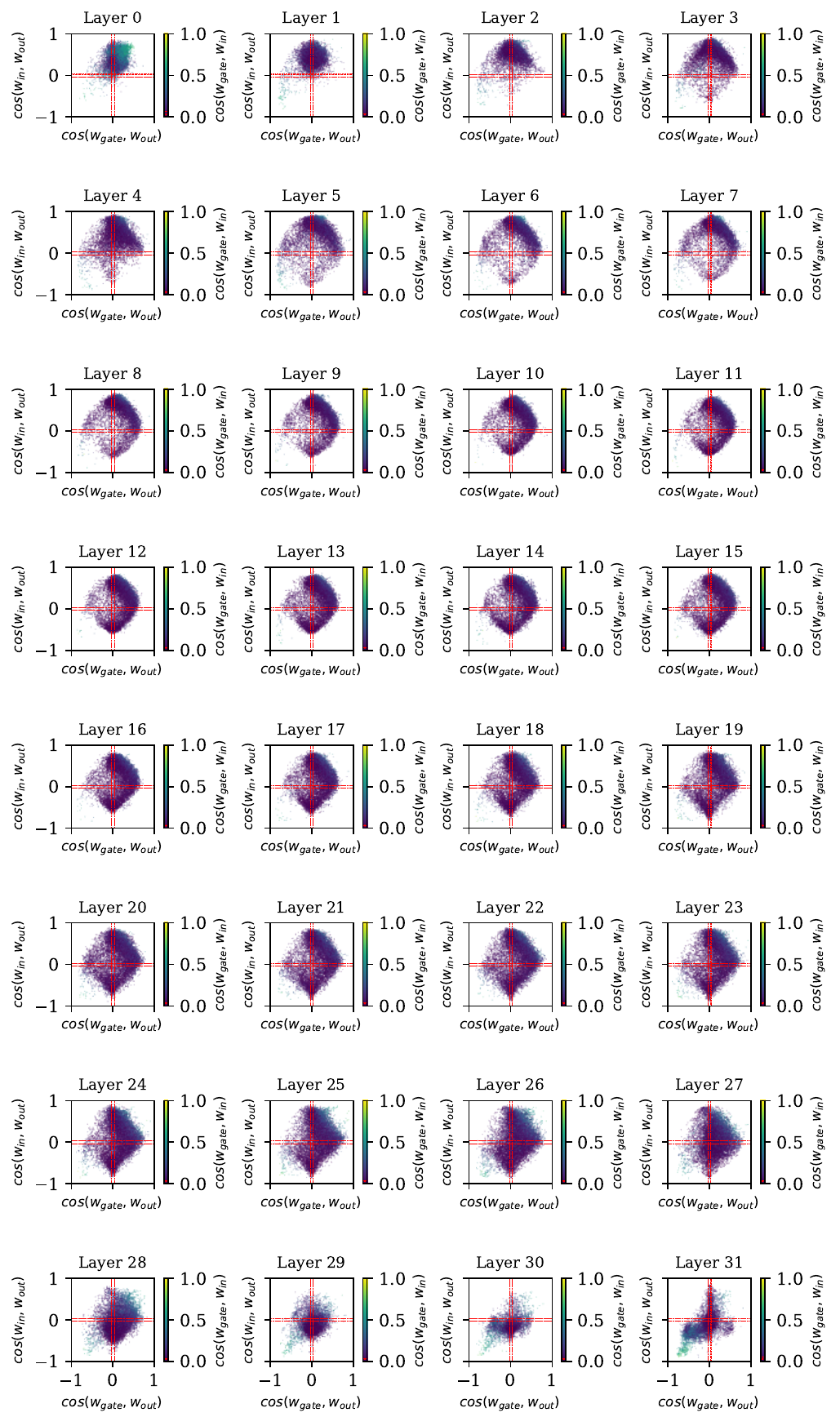}
	\caption{Equivalent of \cref{fig:wcos_selected}
		for OLMo-7B-0424}
\end{figure*}
\clearpage

\begin{figure*}
	\centering
	\includegraphics[width=.67\textwidth]{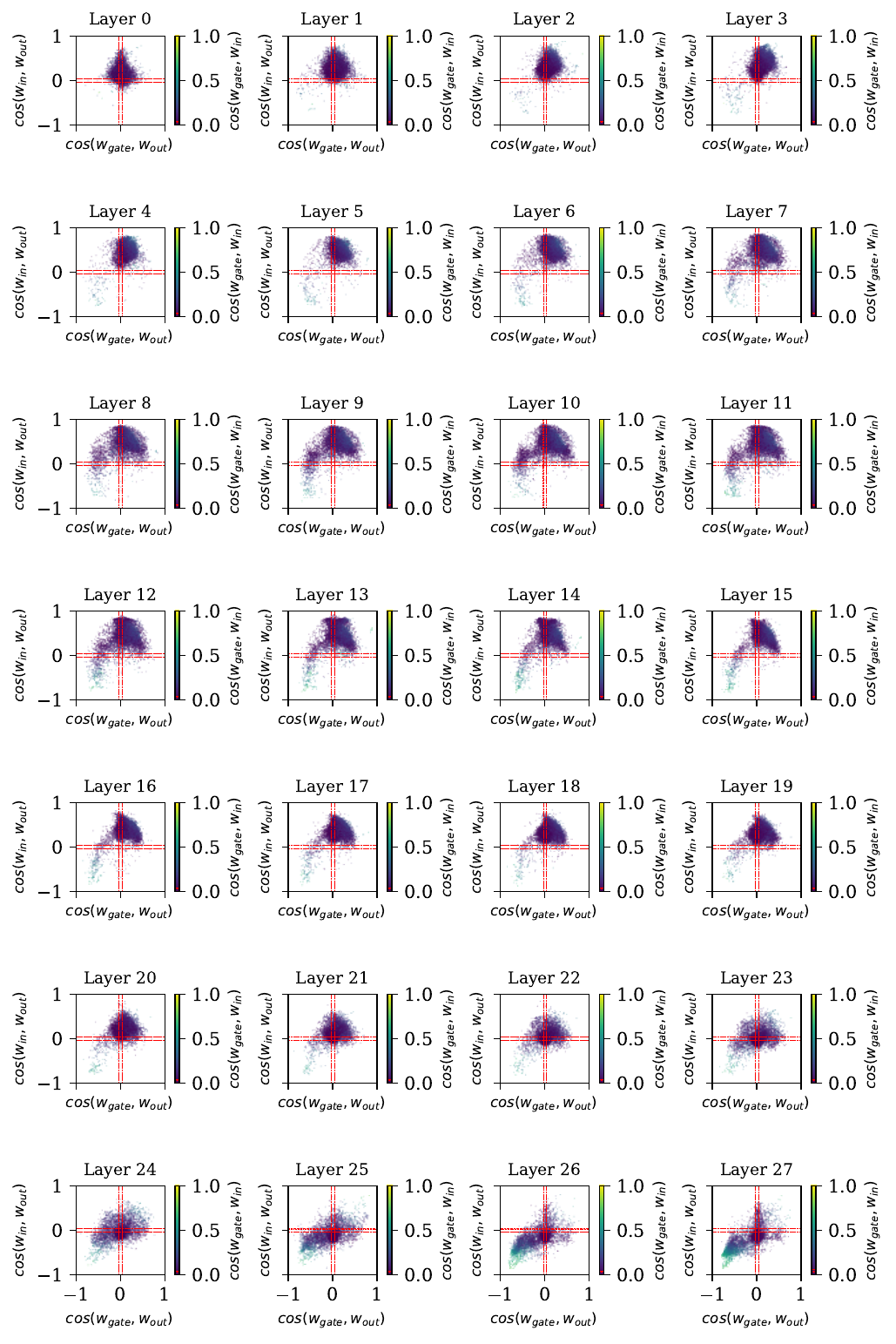}
	\caption{Equivalent of \cref{fig:wcos_selected}
		for Llama-3.2-3B (same model but all layers)}
	\label{fig:wcos}
\end{figure*}
\clearpage

\begin{figure*}
	\centering
	\includegraphics[width=.67\textwidth]{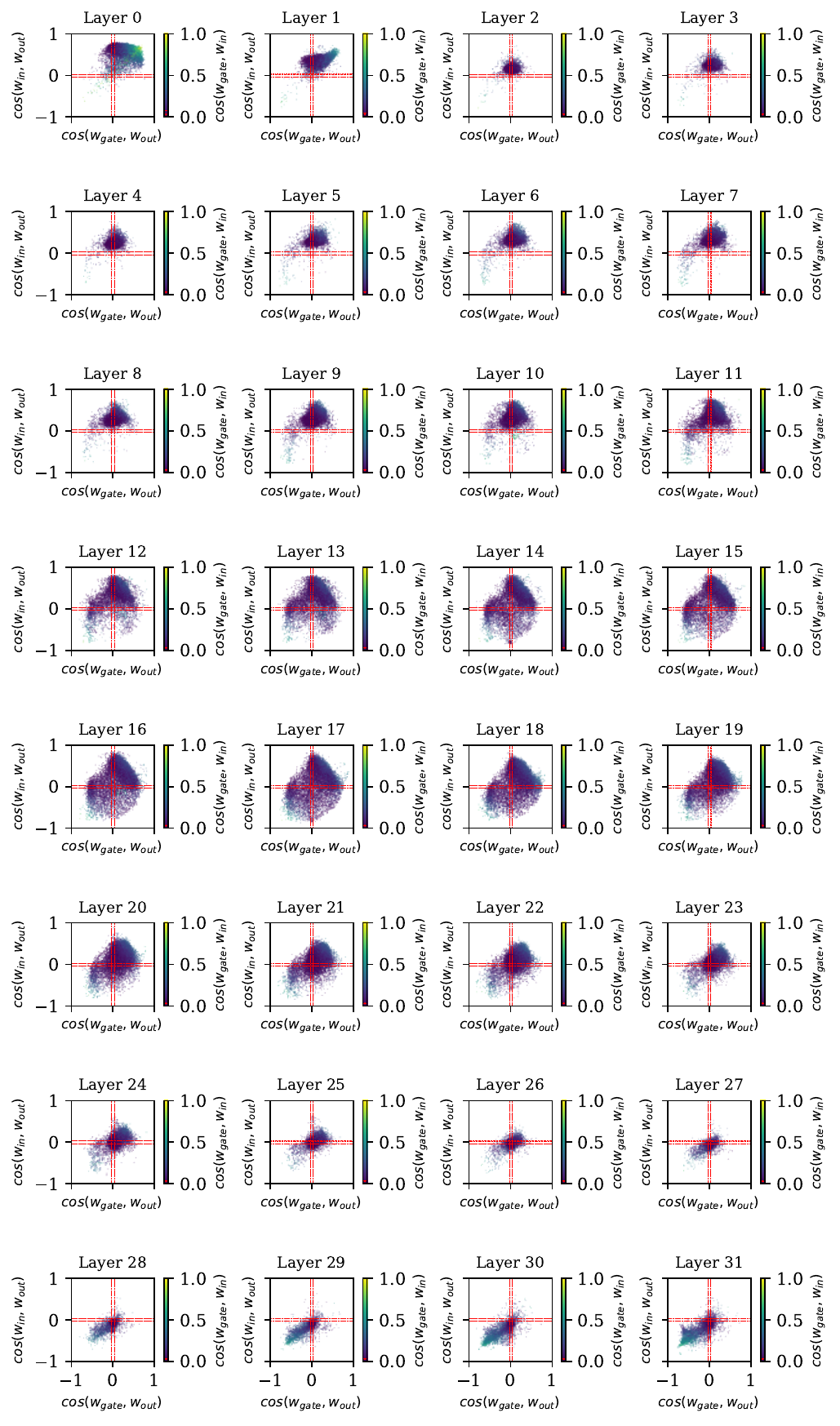}
	\caption{Equivalent of \cref{fig:wcos_selected}
		for Yi-6B}
\end{figure*}

\end{document}